%% file: main_arXiv.tex
\pdfoutput=1
\PassOptionsToPackage{table}{xcolor}
\documentclass[11pt]{article}

\usepackage[]{ACL}

\usepackage{times}
\usepackage{latexsym}

\usepackage[T1]{fontenc}

\usepackage[utf8]{inputenc}

\usepackage{microtype}

\usepackage{inconsolata}

 \usepackage{longtable}
\usepackage{amsmath}
\usepackage{amsfonts}
\usepackage{graphicx}
\usepackage{subcaption}
\usepackage{tikz}
\usepackage{multirow}
\usepackage{float}
\usepackage{booktabs}
\usepackage{tcolorbox}
\usepackage{enumitem}
\usepackage{multicol}
\usepackage{algorithm}
\usepackage{algpseudocode}
\usepackage{soul}
\usepackage{array}
\usepackage{arydshln}
\usepackage{bbm}
\usetikzlibrary{bayesnet}
\usepackage[table]{xcolor}
\usepackage{colortbl}
\usepackage[edges]{forest}
\usepackage{pifont}
\usepackage{orcidlink}
\usepackage{booktabs}
\usepackage{multirow}
\usepackage{makecell}
\usepackage{pifont}
\usepackage{graphicx}       
\usepackage{diagbox}
\usepackage{tabularx}
\usepackage{amssymb}

\usepackage{hyperref}
\usepackage{xcolor}
\usepackage{fontawesome5}
\usepackage{graphicx}

\newcommand{\hflogo}{\raisebox{-0.3ex}{\includegraphics[height=1.1em]{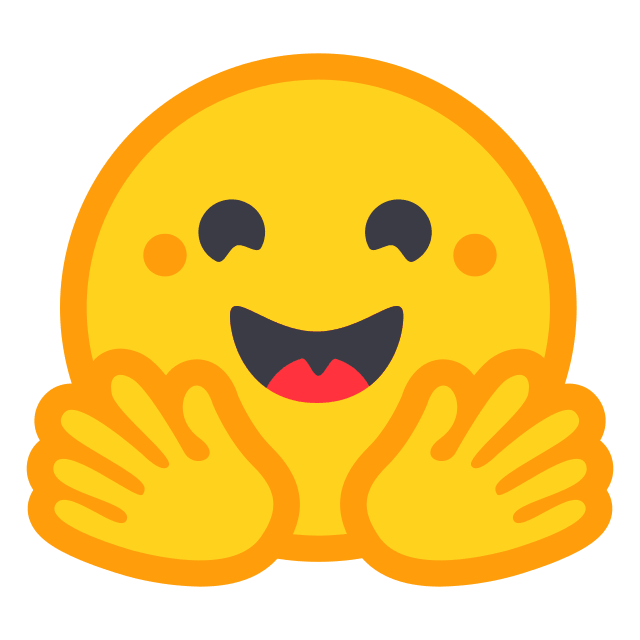}}}
\newcommand{\mslogo}{\raisebox{-0.3ex}{\includegraphics[height=1.1em]{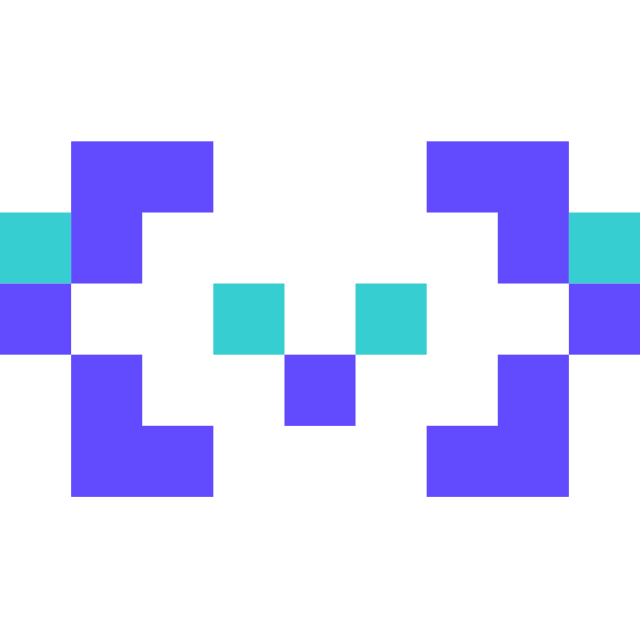}}}

\hypersetup{
    colorlinks=true,
    linkcolor=blue,
    urlcolor=blue,
    citecolor=blue
}

\title{EPIC-Bench: A Perception-Centric Benchmark for Fine-Grained Embodied Visual Grounding in Vision-Language Models}

\author{
 \textbf{Haozhe Shan\textsuperscript{1,2,}\thanks{Core contributors.}},
 \textbf{Xiancong Ren\textsuperscript{1,}\footnotemark[1]\thanks{Project leader.}},
 \textbf{Han Dong\textsuperscript{1,7,}\footnotemark[1]},
 \textbf{Yingji Zhang\textsuperscript{3}},
 \textbf{Haoyuan Shi\textsuperscript{1,4}},\\
 \textbf{Jiayu Hu\textsuperscript{1}},
 \textbf{Qinfan Zhang\textsuperscript{8}},
 \textbf{Yi Zhang\textsuperscript{1}},
 \textbf{Yong Dai\textsuperscript{1}},
 \textbf{Bin Shen\textsuperscript{6}},\\
 \textbf{Lizhen Qu\textsuperscript{5}},
 \textbf{Zenglin Xu\textsuperscript{2}\footnotemark[3]},
 \textbf{Xiaozhu Ju\textsuperscript{1}\thanks{Corresponding authors.}}
\\[0.2em]
 \textsuperscript{1}X-Humanoid,
 \textsuperscript{2}Fudan University,
 \textsuperscript{3}University of Manchester,
 \textsuperscript{4}USTC,\\
 \textsuperscript{5}Monash University,
 \textsuperscript{6}Celonis AI,
 \textsuperscript{7}UNSW,
 \textsuperscript{8}BUAA
\\[0.1em]
\small
{
\hypersetup{urlcolor=blue}%
\href{https://epic-bench.github.io/EPIC-Bench/}{{\color[RGB]{30,144,255}\faGlobe}\ Project Page}
\quad
\href{https://github.com/rxc205/EPIC-Bench-Eval}{{\color{black}\faGithub}\ Evaluation Code}
\quad
\href{https://huggingface.co/datasets/rxc205/EPIC-Bench}{\hflogo\ HuggingFace}
\quad
\href{https://modelscope.cn/datasets/macarich/EPIC-Bench}{\mslogo\ ModelScope}
}}

\begin{document}
\newcommand{\cmark}{\ding{51}} 
\newcommand{\xmark}{\ding{55}} 

\maketitle


\begin{figure*}[t]
\centering
\includegraphics[width=\textwidth]{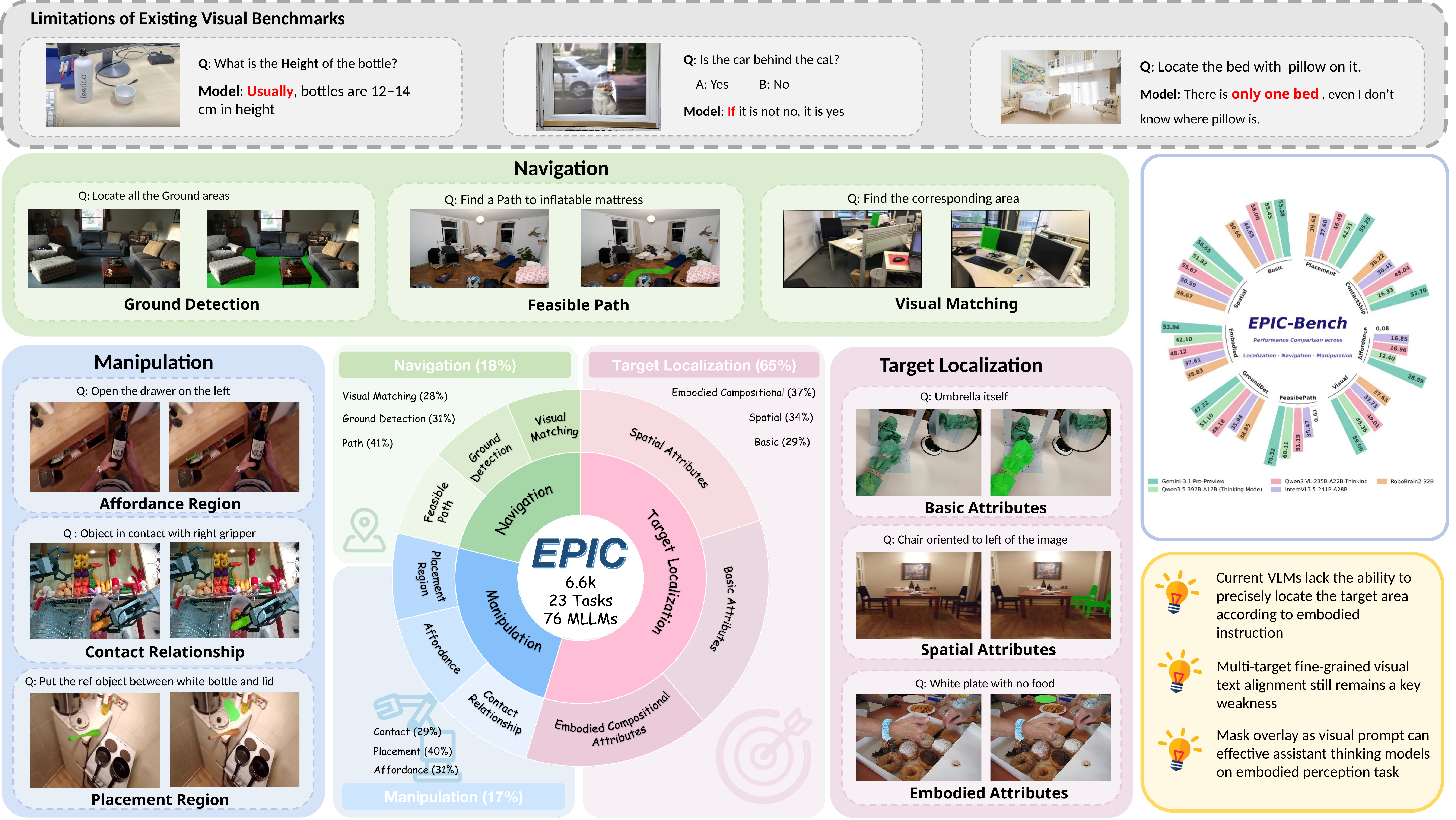}
\caption{Overview of EPIC-Bench. Existing embodied QA and MCQ style benchmarks suffer from language shortcuts and emphasize understanding rather than vision-language grounding. EPIC-Bench evaluates fine-grained embodied visual perception through mask-grounded tasks covering target localization, navigation, and manipulation. With 6,661 human-annotated samples across 23 tasks and multi-dimensional evaluation, EPIC-Bench supports a more diagnostic analysis.}
\label{fig:teaser}
\end{figure*}

\input{sections/abstract_new}
\input{sections/intro_new}

\input{sections/related_work_new}
\input{sections/vlm_grasp_exp}
\input{sections/epic-bench.tex}

\input{sections/evaluation_new}

\input{sections/exp_new}
\input{sections/conclusion.tex}

\input{sections/limitation}

\bibliography{main}

\appendix

\input{sections/appendix_new}

\end{document}

%% file: sections/abstract_new.tex
\begin{abstract}
While large vision-language models (VLMs) are increasingly adopted as the perceptual backbone for embodied agents, 
their fundamental vision-language grounding (VLG) capability in embodied scenarios remains unexplored. We present \textbf{EPIC-Bench}, \textbf{E}mbodied \textbf{P}ercept\textbf{I}on Ben\textbf{C}hmark, a fine-grained grounding benchmark designed to systematically evaluate the VLG capabilities of VLMs in real-world embodied environments. EPIC-Bench comprises 6.6k meticulously human-annotated tuples (Image, Text, Mask) spanning three core stages of the embodied interaction pipeline: Target Localization, Navigation, and Manipulation. With 23 fine-grained tasks, multi-dimensional settings and evaluation over 80 leading VLMs, EPIC-Bench provides a diagnostic analysis.

\end{abstract}

%% file: sections/intro_new.tex
\section{Introduction}

Vision-Language Models (VLMs) have emerged as a promising foundation for embodied intelligence due to their strong multimodal perception and language understanding capabilities. Recent embodied AI systems employ VLMs in different roles, including end-to-end Vision-Language-Action (VLA) policy learning~\cite{zitkovich2023rt,kim2024openvla,intelligence2025pi_05}, high-level planning and reasoning~\cite{ERQA,vlatrace}, and perception modules for grounding language instructions in embodied scenes before downstream execution~\cite{duan2024manipulate,deng2025graspvla,zhong2026dexgraspvla}. This modular perception-action paradigm demonstrates strong generalization potential.



However, these studies generally assume that VLMs possess sufficient vision-language priors and grounding capabilities for action execution. To examine this assumption, we conduct VLM-based grasping experiments and identify two key observations. First, VLMs struggle with compositional semantic following: replacing simple category descriptions (e.g., \textit{cookie box}) with compositional ones (e.g., \textit{the red object pointed to by the scissors}) leads to a substantial decline in localization accuracy. Second, coarse localization degrades action execution: compared with bounding-box grounding, fine-grained mask-level localization significantly improves grasp success in cluttered scenes and for objects with irregular geometries. These findings indicate that the limited grounding capability of VLMs remains a bottleneck for reliable embodied task execution, motivating our focus on vision-language grounding in embodied settings.
 
Existing embodied perception benchmarks mainly adopt question-answering (QA) or multiple-choice (MCQ) formats, which are vulnerable to language shortcuts and emphasize recognition over localization \cite{chen2024mmstar}. Meanwhile, traditional visual grounding benchmarks are not designed for embodied settings and offer limited diagnostic granularity \cite{tong2024eyes}. For example, when a model fails to ground an instruction such as \textit{place the largest red circular object to the right of the table}, existing benchmarks rarely identify whether the failure arises from deficiencies in color discrimination, shape understanding, spatial reasoning, size comparison, or other embodied-specific grounding capabilities.

To address this gap, we introduce \textbf{EPIC-Bench}, a fine-grained grounding benchmark for 2D embodied perception. Specifically,
EPIC-Bench is designed around grounding requirements across the embodied interaction pipeline, from target localization, to navigation, and task-oriented manipulation. 
It contains 6.6k meticulously human-annotated (Image, Text, Mask) tuples spanning 23 tasks for fine-grained diagnostic analysis. We adopt grounding as the primary evaluation protocol and complement it with three additional metrics that account for target multiplicity, ambiguity, and manipulation feasibility, enabling multi-dimensional evaluation. Based on EPIC-Bench, we evaluate over 80 representative VLMs and derive three key findings:

\paragraph{VLMs struggle with complex and fine-grained embodied grounding.} Most models remain substantially below human performance under complex target settings and show notable weaknesses in affordance regions, contact relationships, part-whole reasoning, material discrimination, and orientation-related spatial grounding tasks.

\paragraph{Visual prompting improves fine-grained embodied grounding.} Providing reference object overlays as visual prompts substantially improves fine-grained localization. However, gains vary across models and prompt types, with strong reasoning models benefiting more from mask overlays.

\paragraph{Reasoning improves performance, but inconsistently.} While the strongest-performing models are generally reasoning-enabled, the effectiveness of reasoning varies across model scales and embodied grounding tasks. In certain cases, explicit reasoning even leads to performance degradation, suggesting that reasoning strategies must be carefully designed for embodied applications.

Overall, this work introduces EPIC-Bench, a systematic benchmark for evaluating the embodied vision-language grounding capabilities of VLMs. Our findings provide actionable insights into current limitations and offer guidance for future research on embodied downstream improvement.

%% file: sections/related_work_new.tex
\section{Related Work}

In this section, we review related work from two perspectives: embodied benchmarks and vision-language grounding benchmarks. Table~\ref{tab:related_benchmark_comparison} summarizes representative benchmarks and highlights the differences between existing approaches and EPIC-Bench. We further discuss the importance of vision-language grounding for embodied applications, see Section~\ref{app:vlg_for_embodied_app}.

\input{tabs/related_benchmark_comparison_table}

\paragraph{Embodied Benchmarks.}
Several benchmarks have been proposed to evaluate visual perception in embodied scenarios. OpenEQA~\cite{OpenEQA} and ERQA~\cite{ERQA} focus on embodied understanding and reasoning, while VSI-Bench~\cite{yang2025vsi}, OmniSpatial~\cite{jia2026omnispatial}, and EmbSpatial~\cite{EmbSpatial-Bench} evaluate spatial reasoning capabilities. However, these benchmarks mainly adopt visual question answering or multiple-choice protocols, emphasizing scene understanding and reasoning rather than vision-language grounding. Similarly, BLINK~\cite{fu2024blink} evaluates visual perception but still relies on MCQ-based evaluation. RoboSpatial~\cite{RoboSpatial} and RefSpatial~\cite{zhou2026roborefer} introduce point-selection tasks for spatial reasoning and free-space perception, yet grounding remains only a limited component of their evaluation.

\paragraph{Vision-Language Grounding Benchmarks.}
Vision-language grounding is a fundamental task for evaluating vision-language alignment. RefCOCO~\cite{refcoco} is the most widely used grounding benchmark, while RefCOCOm~\cite{wang2024refcocom} introduces part-level annotations for finer-grained understanding. Subsequent works such as FineCops-Ref~\cite{liu2024finecops} and REF-ADV~\cite{dong2026refadv} further increase grounding difficulty through more detailed and challenging language descriptions. However, these benchmarks focus on generic visual scenes and do not explicitly evaluate grounding under embodied interaction constraints. EPIC-Bench addresses this gap by introducing a fine-grained task taxonomy and evaluation protocol to systematically evaluate and diagnose vision-language grounding capabilities in embodied scenarios.

%% file: tabs/related_benchmark_comparison_table.tex
\begin{table*}[t]
\centering
\scriptsize
\setlength{\tabcolsep}{3pt}
\begin{tabular}{l l |ccc ccc ccc| c |c| c l}
\toprule
\multirow{2}{*}{\textbf{Benchmark / Dataset}} &
\multirow{2}{*}{\textbf{GT Type}} &
\multicolumn{3}{c}{\textbf{Target Localization}} &
\multicolumn{3}{c}{\textbf{Navigation}} &
\multicolumn{3}{c|}{\textbf{Manipulation}} &
\multirow{2}{*}{\textbf{Data Domain}} &
\multirow{2}{*}{\textbf{Manual}} &
\multirow{2}{*}{\textbf{\makecell{Multi\\View}}} &
\multirow{2}{*}{\textbf{Size}} \\

\cmidrule(lr){3-5}
\cmidrule(lr){6-8}
\cmidrule(lr){9-11}
& & \textbf{BA} & \textbf{SRA} & \textbf{ECA} & \textbf{GD} & \textbf{FP} & \textbf{VM} & \textbf{AR} & \textbf{CR} & \textbf{PR} & & & & \\
\midrule

EmbSpatial-Bench\cite{EmbSpatial-Bench} & MCQ & \cmark & \cmark &   &   &   &   &   &   &   & Ind. &  &  & 3.6k \\
RoboSpatial-Home\cite{RoboSpatial} & Point+Binary & \cmark & \cmark &   &   &   &   &   &   & \cmark & Ind. & \cmark &  & 350 \\
RefSpatial-Bench\cite{zhou2026roborefer} & Point/Mask & \cmark & \cmark &   &   &   &   &   &   & \cmark & Gen./Ind. & \cmark &  & 200 \\
OmniSpatial\cite{jia2026omnispatial} & MCQ & \cmark & \cmark & \ding{51}$^{\dagger}$ &   &   &   & \cmark &   &   & Gen./Ind. & \cmark &  & 8.4k \\

OpenEQA \cite{OpenEQA}& QA & \cmark & \cmark & \cmark &   &   &   & \cmark &   &   & Ind./Ego. & \cmark & \cmark & 1.6k \\
ERQA\cite{ERQA} & MCQ & \cmark & \cmark & \cmark &   &   & \cmark & \cmark & \cmark & \cmark & Ind./Ego./Robo & \cmark & \cmark & 400 \\



\midrule

RefCOCO\cite{refcoco} & BBox & \cmark & \cmark &   &   &   &   &   &   &   & Gen. & \ding{51}$^{\ddagger}$ &  & 7.6k \\

RefCOCOm\cite{wang2024refcocom} & Mask & \cmark & \cmark &  \ding{51}$^{\dagger}$  &   &   &   &   &   &   & Gen. & \ding{51} &  & 70k \\

FineCops-Ref\cite{liu2024finecops} & BBox & \cmark & \cmark &   &   &   &   &   &   &   & Gen. & \ding{51}$^{\ddagger}$ &  & 9.6k \\
Ref-Adv\cite{dong2026refadv} & BBox & \cmark & \cmark &   &   &   &   &   &   &   & Gen. & \ding{51}$^{\ddagger}$ &  & 5k \\

\midrule
\rowcolor{blue!8}
\textbf{EPIC-Bench} & \textbf{Mask+Count} & \textbf{\cmark} & \textbf{\cmark} & \textbf{\cmark} & \textbf{\cmark} & \textbf{\cmark} & \textbf{\cmark} & \textbf{\cmark} & \textbf{\cmark} & \textbf{\cmark} & \textbf{Gen./Ind./Robo./Ego.} &\textbf{\cmark} &\textbf{\cmark} & \textbf{6.6k} \\

\bottomrule
\end{tabular}
\caption{
Comparison of perception benchmarks/datasets. $^{\dagger}$ denotes partial support; $^{\ddagger}$ denotes datasets that are automatically generated with manual selection. EPIC-Bench is a manually annotated benchmark with mask-level grounding annotations.
}
\label{tab:related_benchmark_comparison}
\end{table*}

%% file: sections/vlm_grasp_exp.tex
\section{Grounding Failures in VLM-based Grasping}
\label{sec:vlm_grasp}

\input{tabs/vlm_grasp_exp_v2}

First, we conduct a pilot study to reveal the limitations of current VLMs in fine-grained and complex grounding for embodied tasks, thereby motivating the development of EPIC-Bench. 
\paragraph{Experiment Setup.} Specifically, we construct a grasping pipeline in PyBullet by integrating VLMs~\cite{gemini3,gpt52} with an off-the-shelf grasping policy~\cite{kumra2020grconvnet}. Given visual observations and language instructions, the VLM first localizes the target object, after which the predicted localization is used to guide grasp execution. Grounding quality is measured by IoU between predicted localization and oracle masks. Localization is considered successful when IoU exceeds 0.25; in multi-object settings, the predicted target count must additionally match the ground truth. We also report the overall grasp success rate and conditional grasp success rate under correct localization. Note that when localization fails (e.g., low IoU or incorrect target count), grasp execution is still performed to analyze error propagation.

\paragraph{Instruction complexity.}
First, we vary the complexity of language instructions to evaluate grounding robustness in grasping tasks. Specifically, we replace simple category descriptions (e.g., \textit{cookie box}) with compositional expressions (e.g., \textit{the red object pointed to by the scissors}) while keeping the target configuration unchanged. As shown in Table~\ref{tab:grasp_results}, both localization accuracy and overall grasp success rate decline substantially, whereas the oracle baseline remains stable. These results indicate that current VLMs still struggle with embodied grounding, and that grounding errors directly propagate to downstream manipulation performance.

\paragraph{Localization granularity.} Next, we investigate whether more accurate localization improves downstream execution by comparing predicted boxes with refined mask-level localization. As shown in Table~\ref{tab:grasp_results}, refined localization consistently improves both overall and conditional grasp performance, suggesting that precise grounding reduces irrelevant visual interference during grasp generation.

Overall, these results suggest that embodied agents require more than world knowledge. Reliable interaction may also depend on accurately grounding language instructions to the corresponding physical entities in the environment. This motivates a systematic evaluation of vision-language grounding capabilities in embodied scenarios.





%% file: tabs/vlm_grasp_exp_v2.tex
\begin{table}[t]
\centering
\small
\setlength{\tabcolsep}{3pt}

\begin{tabular}{llcccccc}
\toprule
\multirow{2}{*}{Model}
&
\multirow{2}{*}{Setting}
&
\multirow{2}{*}{Loc.}
&
\multicolumn{3}{c}{GS}
&
\multicolumn{2}{c}{CGS}
\\

\cmidrule(lr){4-6}
\cmidrule(lr){7-8}

&
&
&
Box
&
Mask
&
Oracle
&
Box
&
Mask
\\
\midrule
\multirow{2}{*}{Gemini-3}
& Simp.
& 80.0
& 54.0
& 75.0
& 90.0
& 57.5
& 85.0 \\

& Comp.
& 45.5
& 40.0
& 44.0
& 90.0
& 63.0
& 95.0 \\

\midrule

\multirow{2}{*}{GPT-5.2}
& Simp.
& 36.0
& 28.0
& 34.0
& 90.0
& 47.4
& 70.6 \\

& Comp.
& 14.0
& 24.0
& 26.0
& 90.0
& 66.7
& 100.0 \\

\bottomrule
\end{tabular}

\caption{
Localization and grasping performance under different settings.
\textit{Loc.}, \textit{GS}, and \textit{CGS} denote localization success rate,
overall grasp success rate, and grasp success rate conditioned on successful
localization, respectively. Results are averaged over 10 tasks per setting,
with 10 trials per task.
}
\label{tab:grasp_results}
\end{table}

%% file: sections/epic-bench.tex
\begin{figure*}[tb]
    \centering
    \includegraphics[width=\textwidth]{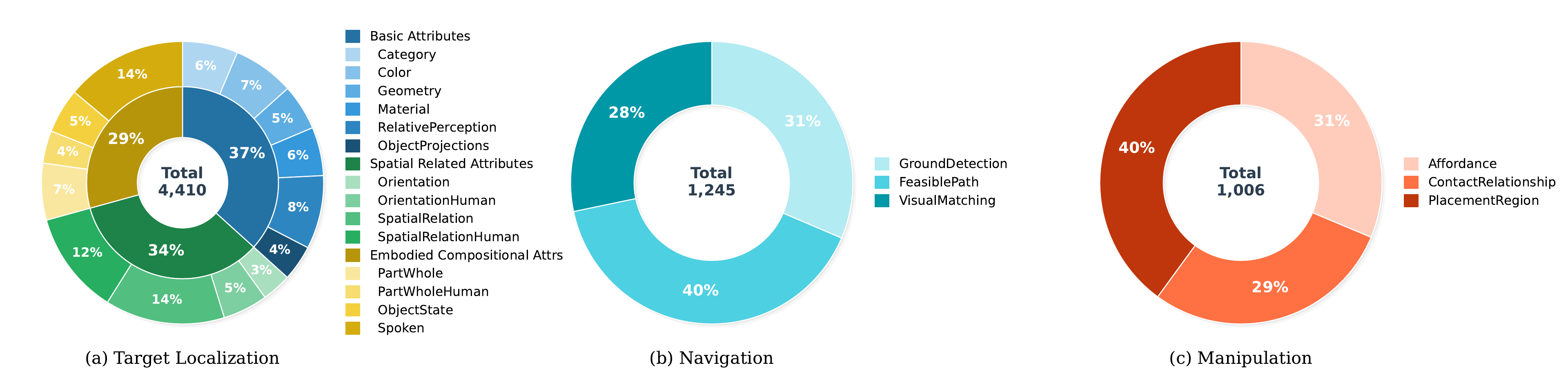}
    \caption{Distribution of EPIC-Bench across three primary task categories. }
    \label{fig:dataset_statistics}
\end{figure*}

\section{The EPIC-Bench}

\paragraph{Overview.} 
EPIC-Bench contains 6,661 manually curated annotations for mask-level embodied visual grounding, covering 3 categories, 9 subcategories, and 23 fine-grained tasks (Figure~\ref{fig:dataset_statistics}). It spans perception requirements from target localization to navigation and manipulation, and includes both single-image and multi-image tasks. Benchmark examples are provided in Appendix~\ref{app:epic_examples}.


\subsection{Task Taxonomy}
\label{sec:task_taxonomy}

\subsubsection{Target Localization (TL)}
Target Localization forms the foundation of embodied navigation and manipulation by evaluating whether a model can accurately ground targets from fine-grained language descriptions. Given an image and a textual instruction, the model must predict both the locations and the number of objects satisfying the description, where the number of valid targets may range from zero to multiple instances. To systematically analyze different sources of grounding difficulty in embodied settings, we organize the tasks into three groups according to the nature of the required grounding attributes:

\textbf{Basic Attributes (BA).}
This group evaluates grounding based on fundamental visual attributes, including \textbf{i.} object category, \textbf{ii.} color, \textbf{iii.} geometry, \textbf{iv.} material, \textbf{v.} relative attributes, and \textbf{vi.} projection. Task instances are designed to prevent target localization based solely on irrelevant attribute cues.

\textbf{Spatial-Related Attributes (SRA).}
This group evaluates grounding based on spatial attributes, including orientation-aware spatial reasoning. Tasks are divided into conventional and human-related spatial descriptions according to the reference subject, with instructions explicitly specifying the spatial reference frame (e.g., camera-centric or subject-centric) to reduce ambiguity.

\textbf{Embodied Compositional Attributes (ECA).}
This group evaluates attribute recognition abilities that are closely related to embodied tasks. It includes four tasks: \textbf{i.} part--whole relationships: identifying object components; \textbf{ii.} human-related part--whole relationships: localizing human anatomical regions for safe interaction; \textbf{iii.} target state differentiation: distinguishing objects based on their functional or physical states; and \textbf{iv.} spoken descriptions: grounding target objects referred to using natural everyday expressions.

\subsubsection{Navigation (NAV)}

This category evaluates visual perception capabilities required for navigation toward target locations. It consists of three subcategories:

\textbf{Ground Detection (GD).}
This task evaluates a model’s ability to identify traversable ground regions. Given an input image, the model is required to localize all ground surfaces for movement.

\textbf{Feasible Path (FP).}
This task evaluates a model’s ability to ground feasible navigation paths under two settings: \textbf{i.} egocentric and \textbf{ii.} exocentric. Given an image with an overlay indicating the target region, the model must generate a valid path either from the camera position to the target or between two specified targets. The predicted path must remain within traversable areas.

\textbf{Visual Matching (VM).}
This task evaluates an embodied agent’s ability to maintain visual correspondence across viewpoints during movement. Given multi-view images of the same environment and a target overlay in one image, the model is required to localize the same target in another view.

\subsubsection{Manipulation (MAN)}
This category evaluates visual perception capabilities required for embodied manipulation. It consists of three subcategories:

\textbf{Affordance Region (AR).}
This task evaluates the ability to identify task-specific operation and tool-use regions. Given an image, a task description, and annotated reference objects, the model must localize the target operational region.

\textbf{Contact Relationship (CR).}
This task evaluates understanding of object contact relationships, including gripper–object interactions during manipulation. Given an image and annotated reference objects, the model must localize targets under different contact conditions, including single-object contact, multi-object contact, and all objects involved in contact interactions.

\textbf{Placement Region (PR).}
This task evaluates perception and reasoning for placement after grasping. Given an image, an overlay and description of the manipulated object, and a textual description of the placement target, the model must localize the placement region and determine whether the placement is feasible under the given context.

\begin{figure*}[tb]
    \centering
    \includegraphics[width=0.9\textwidth]{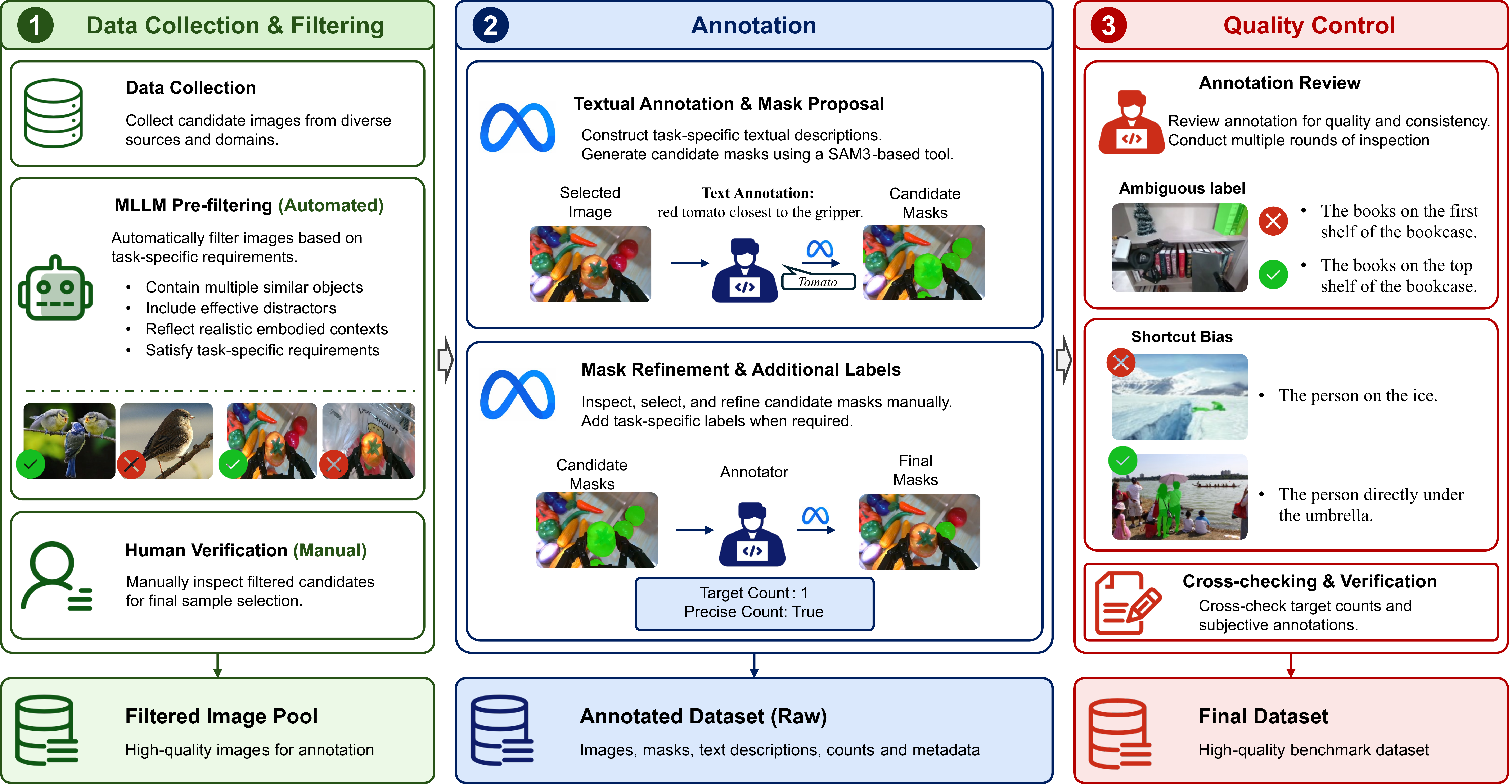}
    \caption{\textbf{Construction pipeline of EPIC-Bench.} The key steps include: (1) Filter images by task-specific requirements, (2) Annotate text and masks with SAM3-assisted tools and manual refinement, (3) Revise or remove low-quality samples through multi-round quality control.}
    \label{fig:data_collection_pipeline}
\end{figure*}


  
  
  
  

  



\subsection{Benchmark Construction}

We design a systematic data curation pipeline consisting of three stages: data filtering, annotation, and quality control, as illustrated in Figure~\ref{fig:data_collection_pipeline}.

\paragraph{Data Sources and Filtering.}
We collect candidate images from over 20 public datasets (Section~\ref{app:epic_data_sources}) covering diverse domains, including generic scenes, indoor environments, egocentric views, and robot-centric imagery, to ensure broad visual diversity and embodied relevance.

For the TL category, selected images must contain multiple objects with similar attributes to serve as distractors, preventing shortcut localization based solely on category cues. For NAV and MAN, we prioritize images that reflect realistic embodied contexts.
We first apply Qwen3-VL-32B for preliminary filtering, followed by manual verification to ensure compliance with task-specific requirements.

\paragraph{Annotation Pipeline.}
 For each subtask, annotators first select appropriate images according to the task requirements and manually construct the corresponding textual descriptions. They then use a SAM3-based annotation tool to generate candidate masks for the target regions. These candidate masks serve only as initial proposals and are manually inspected, selected, and refined by the annotators. When SAM3~\cite{carion2025sam} fails to produce a suitable proposal, annotators directly create the mask annotations manually. For certain task categories, we further collect task-specific annotations, including target counts and binary labels such as precise-count indicators and placement feasibility judgments.

\paragraph{Quality Control.}
To ensure annotation quality and consistency, we conduct at least two rounds of inspection to revise ambiguous descriptions, inconsistent annotations, and shortcut-prone samples. Through this rigorous curation process, EPIC-Bench provides diverse, high-quality annotations for embodied perception tasks. More details can be found in Section~\ref{app:quality_control_details}.

%% file: sections/evaluation_new.tex
\section{Evaluation on EPIC-Bench}




\subsection{Evaluation Setup}
We comprehensively evaluate over 80 VLMs across diverse model families, parameter scales, and training recipes. All evaluations are conducted in zero-shot settings, with results averaged over at least three runs. Prompt templates tailored to individual sub-tasks and prompt robustness experiments are provided in Appendix~\ref{app:prompts}.

\input{tabs/main_evaluation_results}

\subsection{Evaluation Metrics}
Our evaluation framework adopts task-specific metrics. As summarized in Table~\ref{tab:Score_Composition}, the overall score consists of Localization, Counting, Path, and Feasibility scores, with each task using an appropriate subset. Sample scores are computed as weighted averages of selected metrics, and the final benchmark score is averaged over all samples.


\textbf{Localization Score.}
This metric evaluates a model’s grounding precision by computing the intersection-over-union (IoU) between its predicted region and the GT mask. Our main evaluation is based on the models’ original predictions, which are standardized into a unified mask representation for evaluation. We further analyze alternative post-processing strategies and their effects on the evaluation results in Section~\ref{app:post_process}.

\textbf{Counting Score.}
This metric evaluates vision-language alignment through target counting. Let $P$ and $G$ denote the predicted and ground-truth target counts, respectively. The scoring strategy is determined by a binary \textit{precise count} label. If exact counting is required (True), we use binary accuracy:
$
S_{\text{count}}=
\begin{cases}
1 & P=G\\
0 & \text{otherwise}
\end{cases}
$.

Otherwise (False), we apply a soft penalty based on counting error:
$
S_{\text{count}}
=
\max\left(
1-\frac{|P-G|}{G},
0
\right)
$

\textbf{Path Score.}
This metric evaluates a model’s ability to generate physically feasible navigation paths. Models are required to output at least three consecutive trajectory points constrained within traversable regions. Detailed scoring formulations are provided in Section~\ref{app:path_score}.

\textbf{Feasibility Score.}
This metric evaluates a model’s ability to judge the feasibility of embodied actions in a binary format. For example, in placement region tasks, the model determines whether a proposed action is feasible by reasoning over visual cues such as object shape, size, and stability.

\input{tabs/score_composition}




\subsection{Main Results}
Table~\ref{tab:table-performance-select} presents results for representative models, with the full results provided in Section~\ref{app:full_results}.

\paragraph{Reasoning Models Lead, but Differ in Strengths.}
Reasoning-enabled VLMs generally dominate the leaderboard, yet their advantages vary across categories. For example, Qwen3-VL-235B-Thinking performs better on conventional target localization groups, while Gemini-3-Pro achieves stronger performance on embodied grounding tasks. Segmentation-capable models perform strongly at comparable parameter scales and substantially outperform other models on the Ground Detection task. Existing embodied foundation models show no clear grounding advantage, and some exhibit weaker instruction-following ability, potentially due to optimization trade-offs that prioritize other objectives over fine-grained grounding.

%% file: tabs/main_evaluation_results.tex
\begin{table*}[htb]
    \centering
    \setlength{\tabcolsep}{3.5pt}
    \renewcommand{\arraystretch}{1} 
    \footnotesize
    \begin{tabular}{l c ccc ccc ccc}    \toprule
    \multirow{3}{*}{\textbf{Model}}
        & \multirow{3}{*}{\textbf{Average}}
        & \multicolumn{3}{c}{\textbf{Target Local.}}
        & \multicolumn{3}{c}{\textbf{Navigation}}
        & \multicolumn{3}{c}{\textbf{Manipulation}} \\
    \cmidrule(r){3-5} \cmidrule(r){6-8} \cmidrule(r){9-11}
    &  & BA & SRA & ECA & GD & FP & VM & AR & CR & PR. \\
    \midrule

 \rowcolor{blue!8}[0pt][0pt]
    \multicolumn{11}{@{}c@{}}{\textbf{Proprietary Models}} \\

    Gemini-3-Pro$^{\dagger}$ &\colorbox{green!10}{\textbf{54.81}}& \underline{56.81} & \underline{54.50} & \colorbox{green!10}{\textbf{53.90} }& \underline{48.82} & \colorbox{green!10}{\textbf{70.82} }& \colorbox{green!10}{\textbf{60.08}}
    & \colorbox{green!10}{\textbf{30.40}} & \colorbox{green!10}{\textbf{51.78}} & \colorbox{green!10}{\textbf{53.42} }\\

    Claude-Sonnet-4.6$^{\dagger}$         & 43.24 & 45.43 & 45.49 & 36.47 & 40.16 & \underline{68.46} & 42.77 & 9.79 & 46.21 & 43.81 \\


    GPT-5.5                       & 50.16 & \textbf{57.78}	&\textbf{55.37}	&45.19	&\textbf{50.97}	&38.57	& \underline{51.65}	& \underline{24.62}	& \underline{49.67}	& \underline{48.70} \\

    Doubao-Seed-1.8$^{\dagger}$           & \underline{46.32} & 55.29 & 49.68  & \underline{46.10} & 46.02 & 53.43 & 45.90 & 10.94 & 14.36 & 40.79 \\


    HunYuan-T1-Vision$^{\dagger}$         & 43.00 & 51.24 & 47.60  & 39.67 & 42.63 & 50.08 & 33.54 & 6.63  & 38.95 & 34.49 \\

    \midrule

    \rowcolor{blue!8}[0pt][0pt]
    \multicolumn{11}{@{}c@{}}{\textbf{Open-source Models}} \\


    Qwen3-VL-235B-A22B$^{\dagger}$      & \textbf{50.93} & \colorbox{green!10}{\textbf{58.00}} & 
\colorbox{green!10}{\textbf{55.67}}   & \textbf{48.12} & \textbf{48.18} & \textbf{51.19} & \textbf{49.01} & \textbf{16.96} & \textbf{48.04} & \textbf{46.49} \\


    InternVL3.5-241B-A28B & 40.75 & 44.65 & \underline{50.59} & 37.61 & 35.94 & 35.47 & 33.73 & \underline{16.85} & \underline{36.41} & 37.60 \\



    MiMo-VL-7B-RL-2508                            &34.65	&36.65&	42.31	&34.92	&18.31	&36.90	&\underline{37.45}	&6.51	&30.37&	32.86 \\
    Gemma-3-27B-IT                                & 27.17 & 26.93 & 33.66 & 26.68 & 11.62 & 32.17 & 26.33 & 1.65  & 31.26 & 32.18 \\
    GLM-4.6V$^{\dagger}$                & \underline{42.84} & \underline{50.19} & 47.71 & \underline{37.89} & \underline{42.46} & \underline{51.11} & 36.55 & 8.32  & 34.14 & \underline{39.79} \\
    Step3-VL-10B                                  & 32.40 & 42.22 & 40.10 & 31.88 & 17.96 & 24.03 & 17.77 & 5.94  & 27.22 & 27.56 \\
    LLaVA-NeXT-72B                                & 20.40 & 26.54 & 29.42 & 19.93 & 14.18 & 6.34  & 6.19  & 2.99  & 12.75 & 18.63 \\

    \midrule

    \rowcolor{blue!8}[0pt][0pt]
    \multicolumn{11}{@{}c@{}}{\textbf{Segmentation-capable Models}} \\
    GLaMM-7B  & 28.03 & 37.64 & 27.26 & 18.92 & \colorbox{green!10}{\textbf{84.23}} & \underline{37.88} & 0.01 & 1.27 & 3.40 & 17.67 \\

    UFO-8B    & \textbf{42.39} & \textbf{50.29} & \textbf{51.21} & \textbf{34.31} & \underline{78.51}  & \textbf{45.45} & \underline{3.52} & \textbf{8.00} & \textbf{35.97} & \underline{30.21} \\

    LISA++-7B & \underline{37.17} & \underline{37.96} & \underline{50.22} & \underline{28.94} & 76.34 & 23.56 & \textbf{26.51} & \underline{6.70} & \underline{26.73} & \textbf{31.37} \\

    \midrule

    \rowcolor{blue!8}[0pt][0pt]
    \multicolumn{11}{@{}c@{}}{\textbf{Embodied Foundation Models}} \\
    Pelican1.0-VL-72B                             & \underline{35.29} & \underline{44.08} & \underline{42.11} & \underline{34.79} & \textbf{44.91} & 0.00  & \textbf{38.13} & \textbf{7.32} & 22.68 & \underline{37.96} \\

    RoboBrain2-32B                                & \textbf{39.32} & \textbf{50.66} & \textbf{49.67} & \textbf{38.83} & \underline{38.85} & 0.61  & \underline{37.63} & 0.08 & \textbf{36.22} & \textbf{39.61} \\

    RoboBrain2.5-8B-MT                            & 29.83 & 31.20 & 37.41 & 26.76 & 33.52 & \textbf{30.89} & 27.84 & 2.65 & \underline{25.45} & 24.34 \\

    RynnBrain-Plan-8B                             & 23.59 & 23.53 & 32.53 & 26.08 & 2.12 & 19.57  & 21.41 & 0.11 & 18.79 & 31.02 \\

    VeBrain                                       & 29.27 &	33.63 &	35.98 & 27.69 & 29.56 &	\underline{20.04}  & 27.58 & 2.85&19.25 & 32.30 \\
    Cosmos-Reason1-7B                             & 23.25 & 29.74 & 26.64 & 19.67 & 24.68 & 3.03  & 25.44 & \underline{5.37} & 22.12 & 32.75 \\

    \bottomrule
    \end{tabular}
    \caption{
        Overall performance of representative VLMs on EPIC-Bench.
        \colorbox{green!10}{\textbf{Green bold}} indicates the overall best result across all models.
        \textbf{Bold} and \underline{underline} indicate the best and second-best results within each model category, respectively.
        $^{\dagger}$ denotes models evaluated with thinking mode enabled.
    }
    \label{tab:table-performance-select}
\end{table*}

%% file: tabs/score_composition.tex
\begin{table}[t]
\centering
\small
\resizebox{8cm}{!}{
\begin{tabular}{l l}
\toprule
\textbf{Task Type} & \textbf{Score Composition} \\
\midrule
TL-All &
$(1-\alpha)\,\text{Localization Score} + \alpha\,\text{Counting Score}$ \\
\midrule
NAV-Ground Detection &
Localization Score \\
NAV-Feasible Path &
Path Score \\
NAV-Visual Matching &
$(1-\alpha)\,\text{Localization Score} + \alpha\,\text{Counting Score}$ \\
\midrule
MAN-Affordance Region &
Localization Score \\
MAN-Contact Relationship &
$(1-\alpha)\,\text{Localization Score} + \alpha\,\text{Counting Score}$ \\
MAN-Placement Region &
$(1-\alpha)\,\text{Localization Score} + \alpha\,\text{Feasibility Score}$ \\
\bottomrule
\end{tabular}
}
\caption{\textbf{Task-specific score compositions.} The weighting factor $\alpha$ defaults to 0.5. For TL, $\alpha=0.3$ when \textit{Precise Count} is False and $\alpha=0.1$ when the instruction contains explicit numerical information. For MAN-CR, $\alpha=0.3$ when \textit{Precise Count} is False.}
\label{tab:Score_Composition}
\end{table}

%% file: sections/exp_new.tex
\subsection{Empirical Analysis}
Beyond the main benchmark results, we further analyze embodied vision-language grounding from three perspectives to understand its key limitations.


\subsubsection{What Limits Embodied VLG?}
\begin{figure}[ht!]
  \centering
  \includegraphics[width=0.5\textwidth]{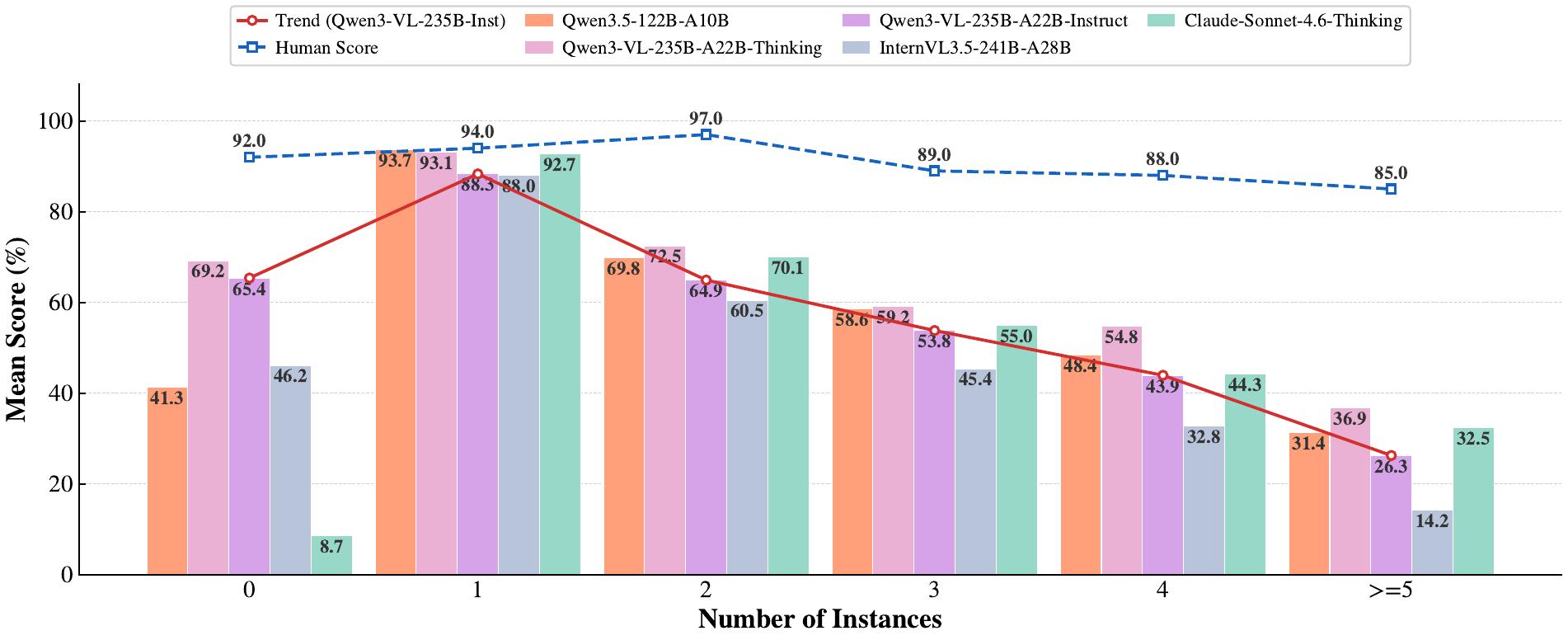}
\caption{
Counting accuracy across different target counts. Performance drops substantially in zero-target and multi-target settings compared with single-target cases.
}
  \label{fig:number_score}
\end{figure}

\paragraph{Target Number Generalization Remains Limited.} Preliminary analysis shows that models frequently struggle to predict the correct number of targets. We therefore evaluate counting performance using samples with Precise Count label. As shown in Figure~\ref{fig:number_score}, models achieve near-human accuracy for single-target cases, but performance drops substantially in zero-target and multi-target settings. One possible explanation is the bias of existing grounding datasets, where most samples assume one-to-one correspondence between text and targets, encouraging hallucinated or incomplete predictions. We hypothesize that this limitation contributes to the lower performance on Target Localization tasks, which jointly evaluate target counting and grounding accuracy.

\begin{figure}[ht!]
  \centering
\includegraphics[width=0.5\textwidth]{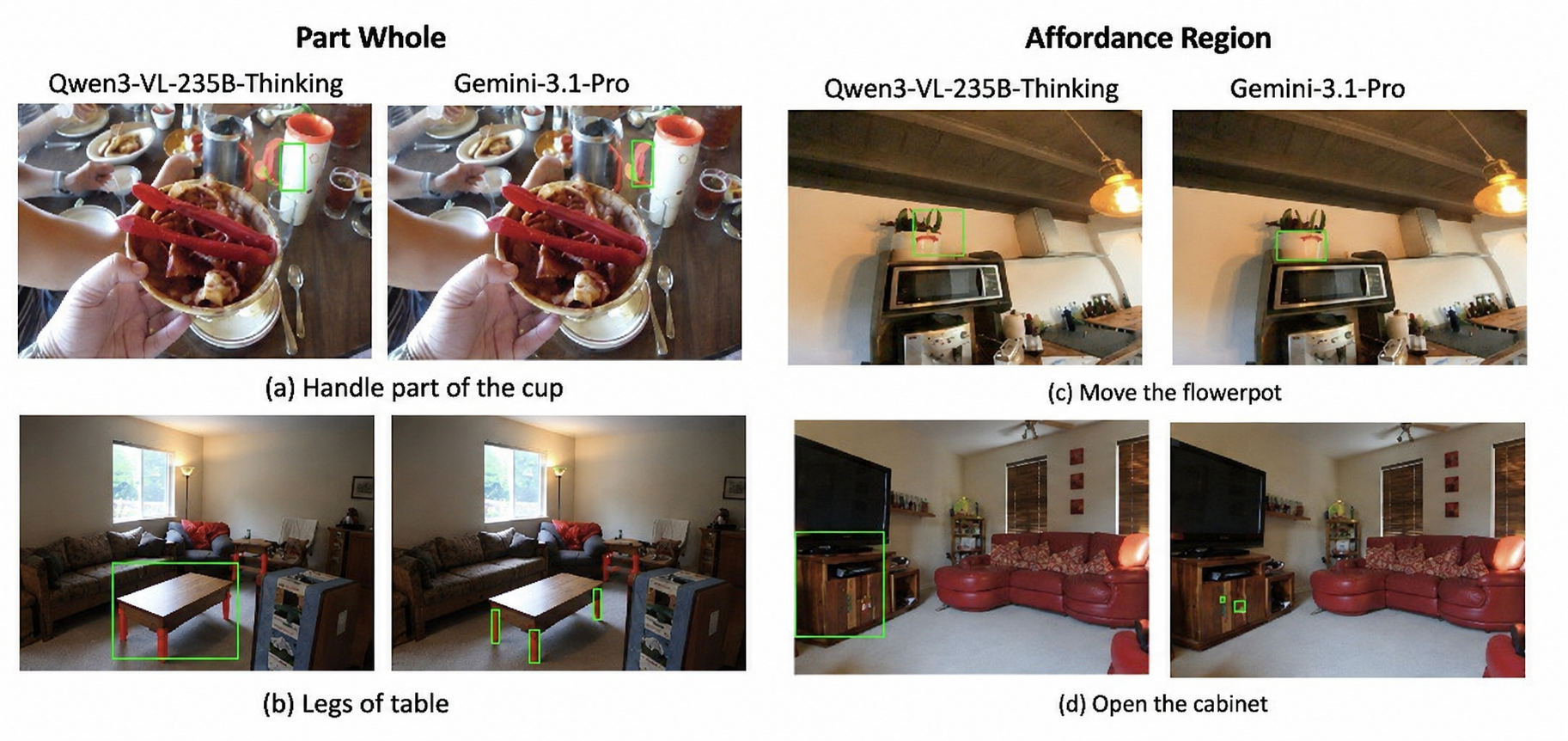}
\caption{
Case studies on Part--Whole and Affordance Region tasks. GT masks are shown in red and model predictions as green boxes. Gemini-3.1-Pro achieves better grounding performance than Qwen3-VL-235B, but still exhibits incomplete or inaccurate localization in challenging cases.
}
  \label{fig:affordance_case}
\end{figure}

\paragraph{Fine-Grained Grounding Remains Challenging.}
In addition, models consistently underperform on affordance region and part-whole tasks. The case study in Figure~\ref{fig:affordance_case} suggests that models tend to produce object-level localization, whereas these tasks require finer-grained, part-level grounding. Moreover, although models perform relatively well on standard spatial relations, their performance drops noticeably on orientation-sensitive grounding, indicating limited understanding of viewpoint-dependent spatial cues.

\subsubsection {Effect of Visual Prompting.}
Next, we conduct ablation studies to evaluate the effectiveness of overlay-based visual prompts for embodied grounding.

\begin{figure}[ht!]
  \centering
  \includegraphics[width=0.5\textwidth]{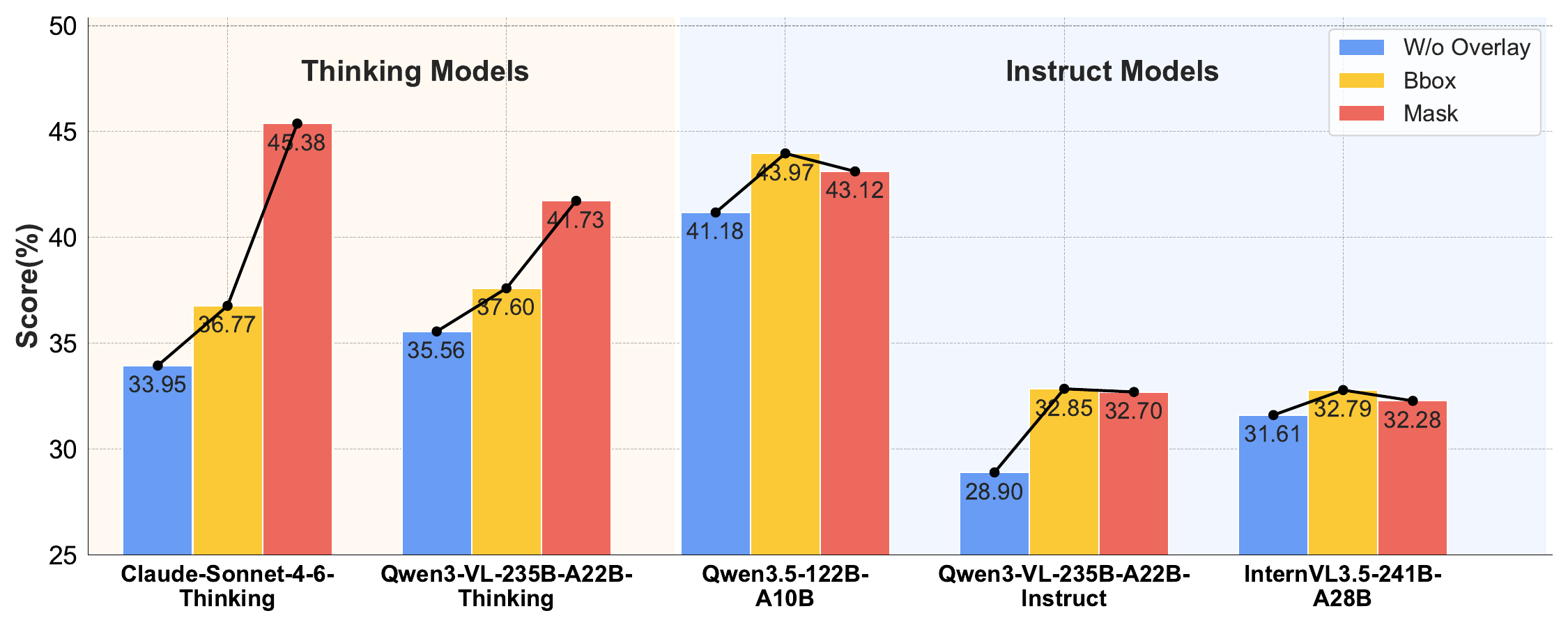}
  \caption{Effect of visual prompting on embodied grounding. Introducing visual overlays consistently improves performance, with stronger reasoning models benefiting more from mask overlays and smaller models preferring simpler bbox prompts.}
  \label{fig:visual_prompt_cmp}
\end{figure}

\paragraph{Coarse Visual Prompts Improve Fine-Grained Grounding.}
EPIC-Bench introduces mask overlays in selected navigation and manipulation tasks. As shown in Figure~\ref{fig:visual_prompt_cmp}, both bbox and mask overlays improve performance over the no-overlay setting, highlighting the effectiveness of visual prompting for embodied grounding. Reasoning models benefit more from mask overlays, while weaker models tend to perform better with simpler bbox overlays.

\subsubsection{Does Thinking Always Help?}
\label{sec:thinking helps}

\begin{figure}[ht!]
  \centering
  \begin{subfigure}{0.5\textwidth}
  \includegraphics[width=\linewidth]{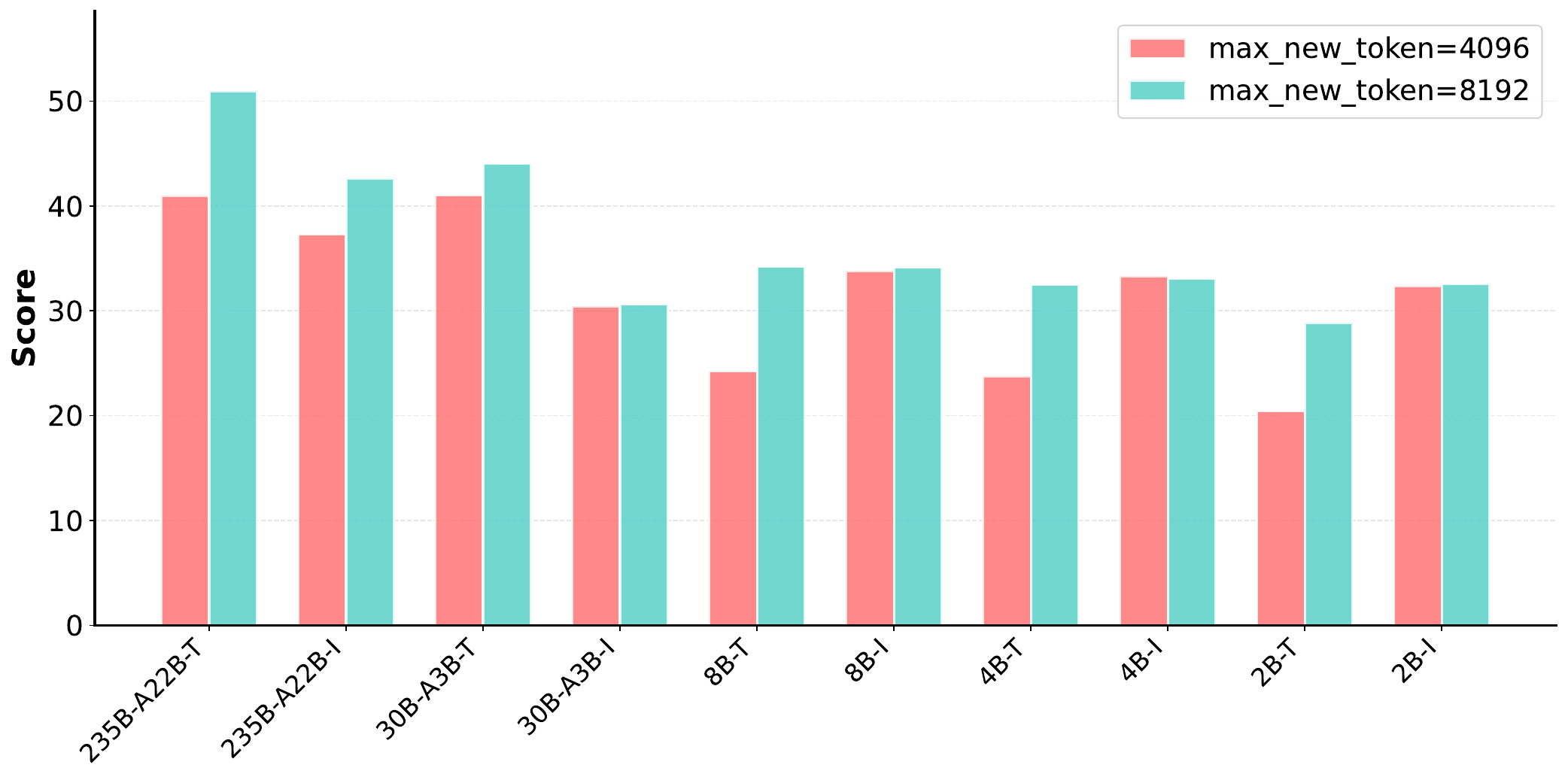}
    \caption{Impact of maximum generation length on model performance. T and I denote Thinking and Instruct variants, respectively.}
    \label{fig:max_length_cmp}
  \end{subfigure}
  \hfill
  \begin{subfigure}{0.5\textwidth}
  \includegraphics[width=\linewidth]{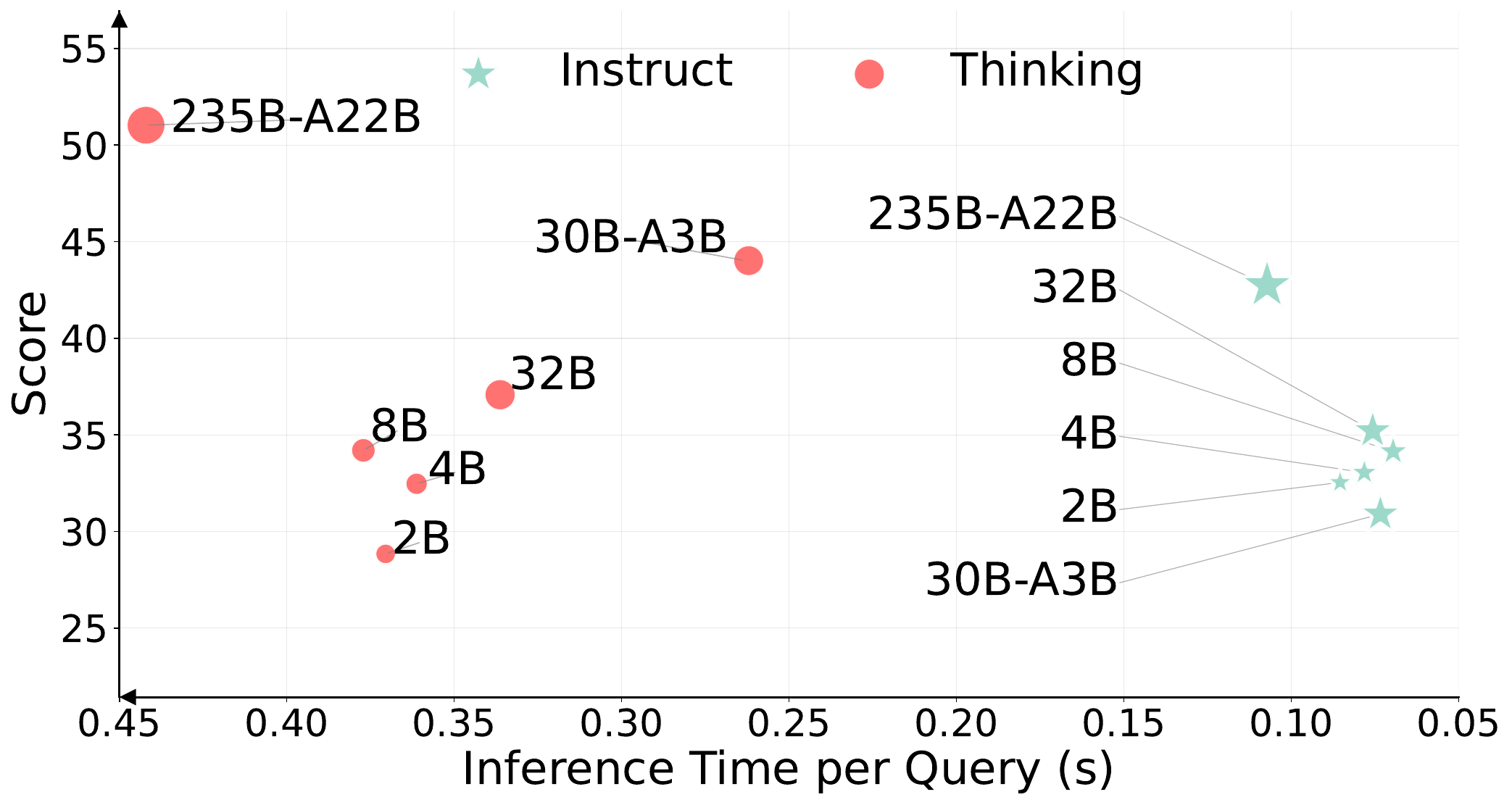}
    \caption{Performance--efficiency trade-off between benchmark score and average inference time.}
    \label{fig:efficiency}
  \end{subfigure}
  \caption{Generation budget and efficiency analysis of the Qwen3-VL model family}
  \label{fig:short}
\end{figure}

Although top-performing models are predominantly reasoning-enabled, our ablation study suggests that explicit reasoning is not universally beneficial.

\paragraph{Impact of Reasoning Length Constraints.} 
We analyze the Qwen3-VL series under different maximum answer lengths to study reasoning efficiency. As shown in Figure~\ref{fig:max_length_cmp}, reasoning models generally outperform or match non-reasoning counterparts when generation budgets are sufficient, with larger models benefiting the most. Under restricted output lengths, however, smaller reasoning models may underperform due to repetitive reasoning loops that exhaust the generation budget, whereas larger models remain more robust to truncation.

\paragraph{Performance-Efficiency Trade-off.}
Figure~\ref{fig:efficiency} illustrates the relationship between average inference time and benchmark performance (8×A100 GPUs), providing practical guidance for embodied system deployment. Although increasing model scale and extending reasoning length can improve performance, both introduce additional inference costs, particularly in latency. Consequently, larger reasoning models are not always the optimal choice in practice. Instead, selecting an appropriate balance between reasoning capability and computational efficiency may lead to more favorable deployment trade-offs.

%% file: sections/conclusion.tex
\section{Conclusion}

We introduce EPIC-Bench, a benchmark for systematically evaluating embodied vision-language grounding in VLMs beyond existing evaluation protocols. Through extensive evaluation across representative models, we identify key challenges in embodied grounding and provide insights for future model development and embodied applications.

%% file: sections/limitation.tex
\section{Limitations}


EPIC-Bench focuses on vision-language grounding from 2D RGB observations, which remain widely available and commonly used as a primary visual input modality in contemporary embodied systems and robot-learning models~\cite{brohan2022rt,black2024pi_0,intelligence2025pi_05,vlatrace,kim2024openvla}. Recent works have explored complementary settings with richer modalities or dynamic interaction~\cite{hong2026esi,lu2026wrbench}. Extending EPIC-Bench toward these settings is a natural direction for future work. Given the substantial challenges observed in the current setting, improving grounding from 2D visual observations remains an important and non-trivial problem.

%% file: sections/appendix_new.tex
\appendix
\clearpage

\section{Appendix Outline}

In these supplementary materials, we provide:
\begin{itemize}[noitemsep, topsep=0pt, leftmargin=*]
    \item Additional related work on vision-language grounding in embodied applications (Section~\ref{app:vlg_for_embodied_app});
    \item Benchmark examples, data sources, and quality-control details (Section~\ref{app:construction_details});
    \item Evaluation prompts, prompt robustness experiments, post-processing analysis, and Path Score details (Section~\ref{app:evaluation_details});
    \item Full evaluation results for all tested models (Section~\ref{app:full_results}).
\end{itemize}

\input{sections/Appendix/Section_B}

\input{sections/Appendix/Section_C}

\input{sections/Appendix/Section_D}
\input{sections/Appendix/Section_E}

%% file: sections/Appendix/Section_B.tex
\section{Vision-Language Grounding for Embodied Application}
\label{app:vlg_for_embodied_app}

Vision-language grounding (VLG) is increasingly important for embodied intelligence, as many recent systems rely on intermediate grounding representations to connect language instructions with executable physical targets.

A common paradigm separates perception from control. This decomposition can facilitate generalization by decoupling semantic interpretation from physical execution. In these systems, VLMs first localize instruction-relevant target regions, and downstream policies generate actions conditioned on the grounded observations. For example,~\cite{stone2023moo,duan2024manipulate} use VLMs to localize the target regions referred to by language instructions before planning or grasp generation.~\cite{zhong2026dexgraspvla} convert language instructions into object-level grasp specifications and conditions a DiT-based policy on localized object representations. Similarly,~\cite{tziafas2024OWG,liu2024moka,huang2024rekep} rely on target masks for keypoint estimation or manipulation planning.~\cite{deng2025graspvla} explicitly introduces grounding as an intermediate stage before action generation, while~\cite{huang2025roboground} predicts target and placement masks that subsequently condition downstream VLA execution.

Recent end-to-end VLA methods also increasingly strengthen grounding-aware representations during training. For instance,~\cite{zhong2026dexgraspvla} uses target-mask features as action conditions to suppress background interference and promote object-centric representations, while~\cite{song2026reconvla} introduces an auxiliary reconstruction objective over instruction-relevant regions to improve fine-grained perception and grounding.

Overall, these works suggest that VLG is becoming a key component of embodied systems, with direct relevance to generalization, manipulation, and reliable interaction.

%% file: sections/Appendix/Section_C.tex
\section{Benchmark Construction Details}
\label{app:construction_details}

\subsection{Benchmark Examples}
\label{app:epic_examples}
We present detailed examples from EPIC-Bench to illustrate the input–output content required by tasks across different categories. For certain tasks, we also provide optional auxiliary annotations. These additional labels enable more accurate evaluation by accounting for the specific characteristics of different tasks and data types.

\begin{figure*}[tb]
\centering
\includegraphics[width=0.495\textwidth]{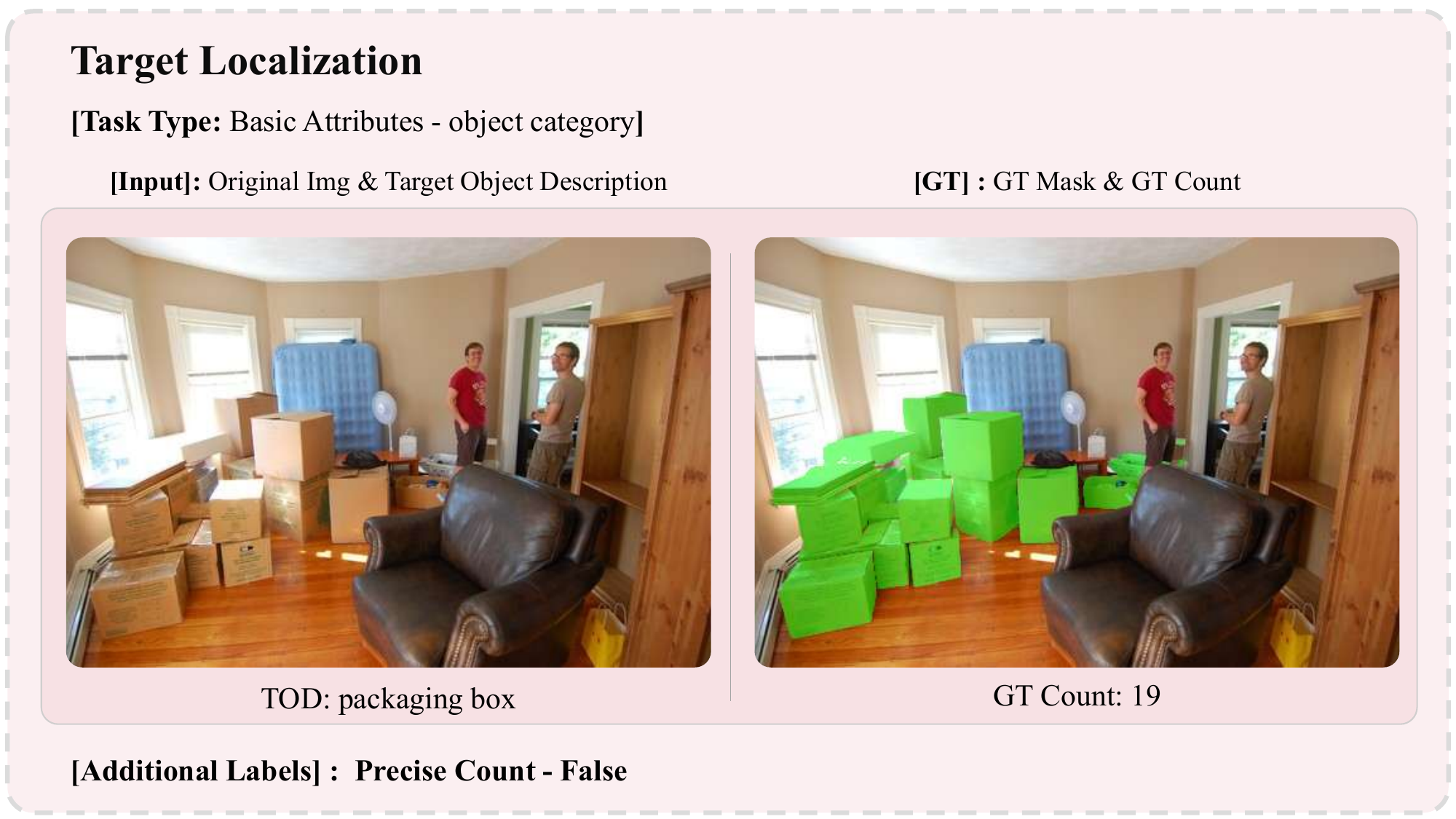}
\includegraphics[width=0.495\textwidth]{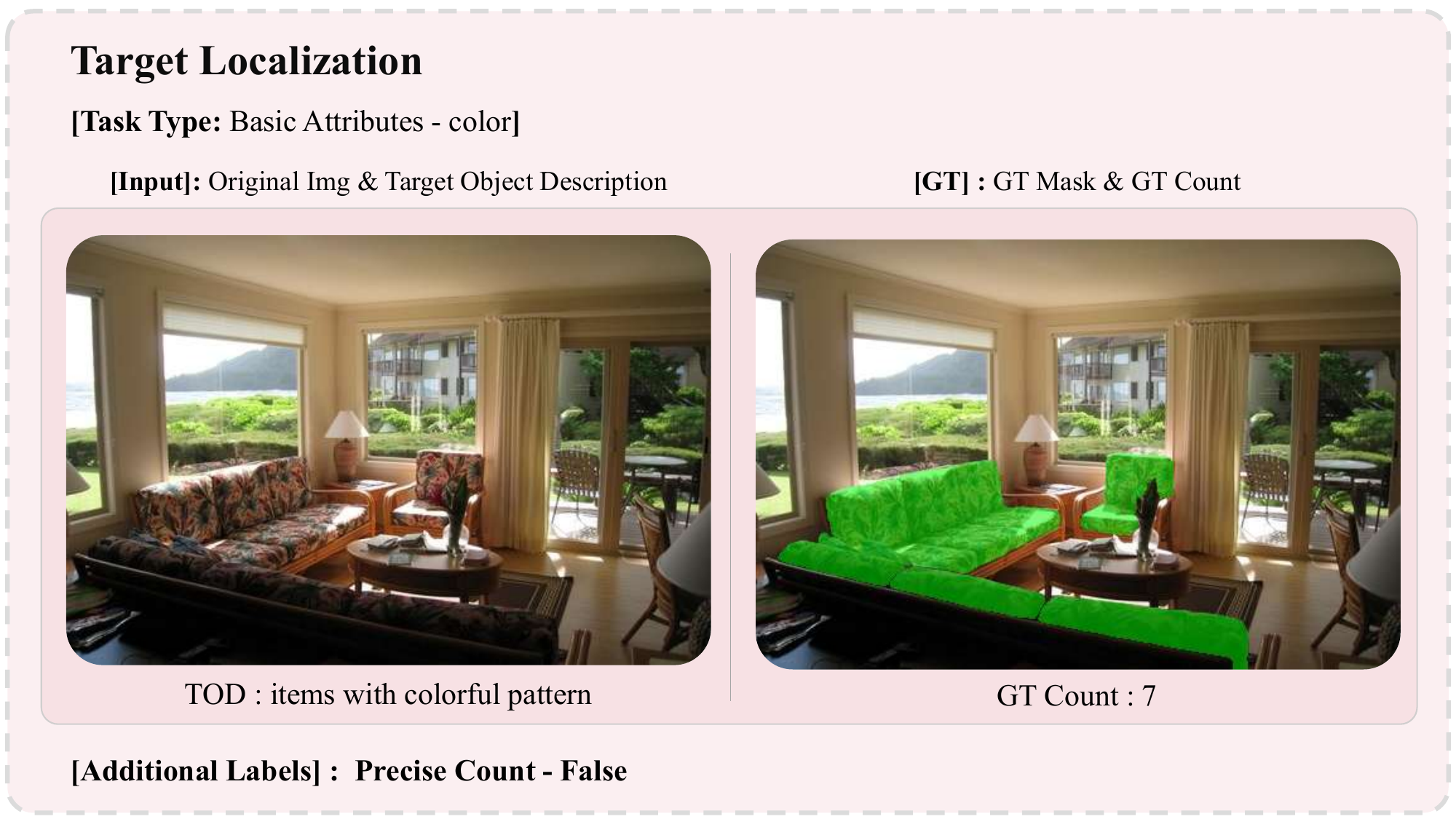}

\includegraphics[width=0.495\textwidth]{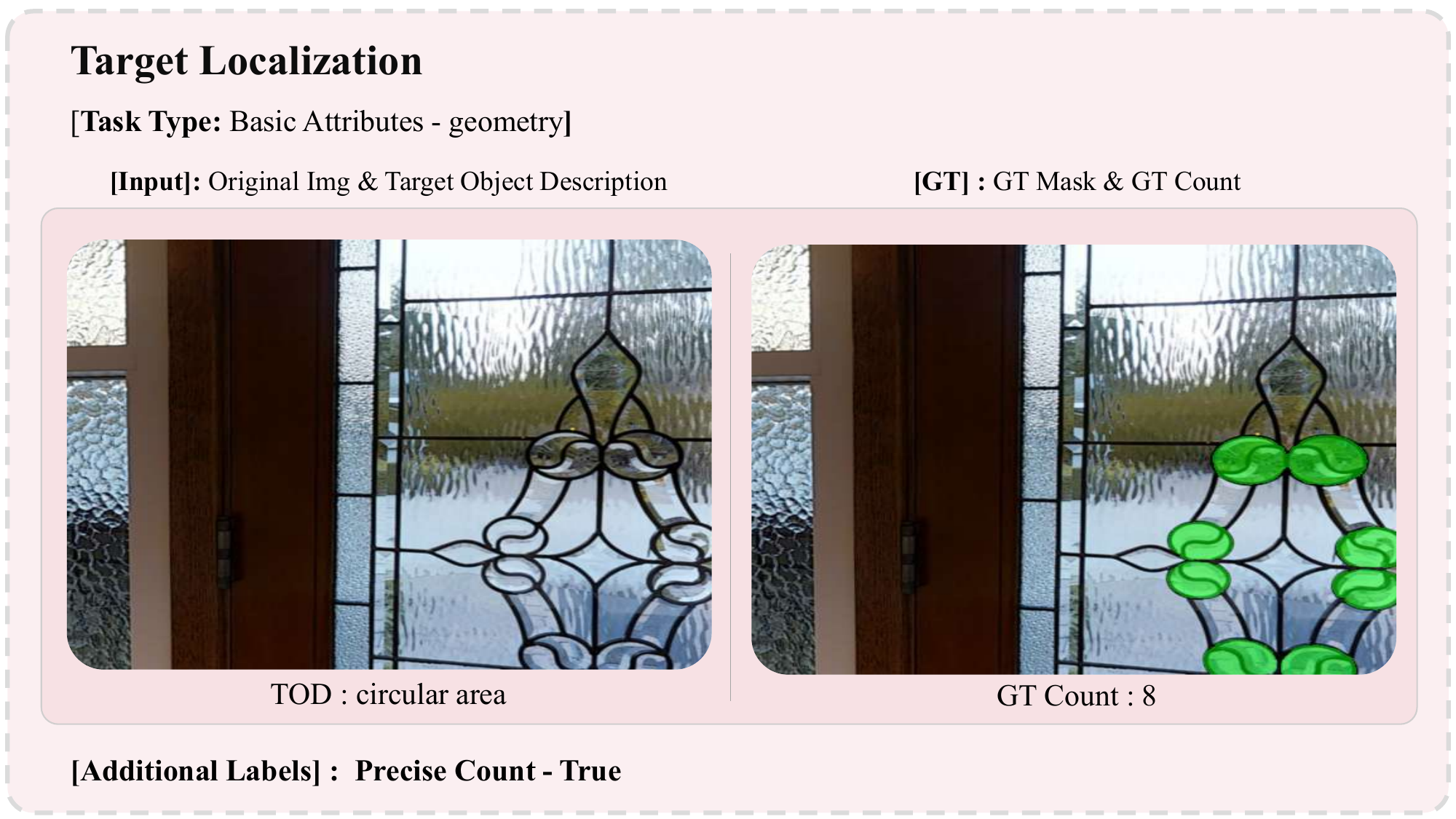}
\includegraphics[width=0.495\textwidth]{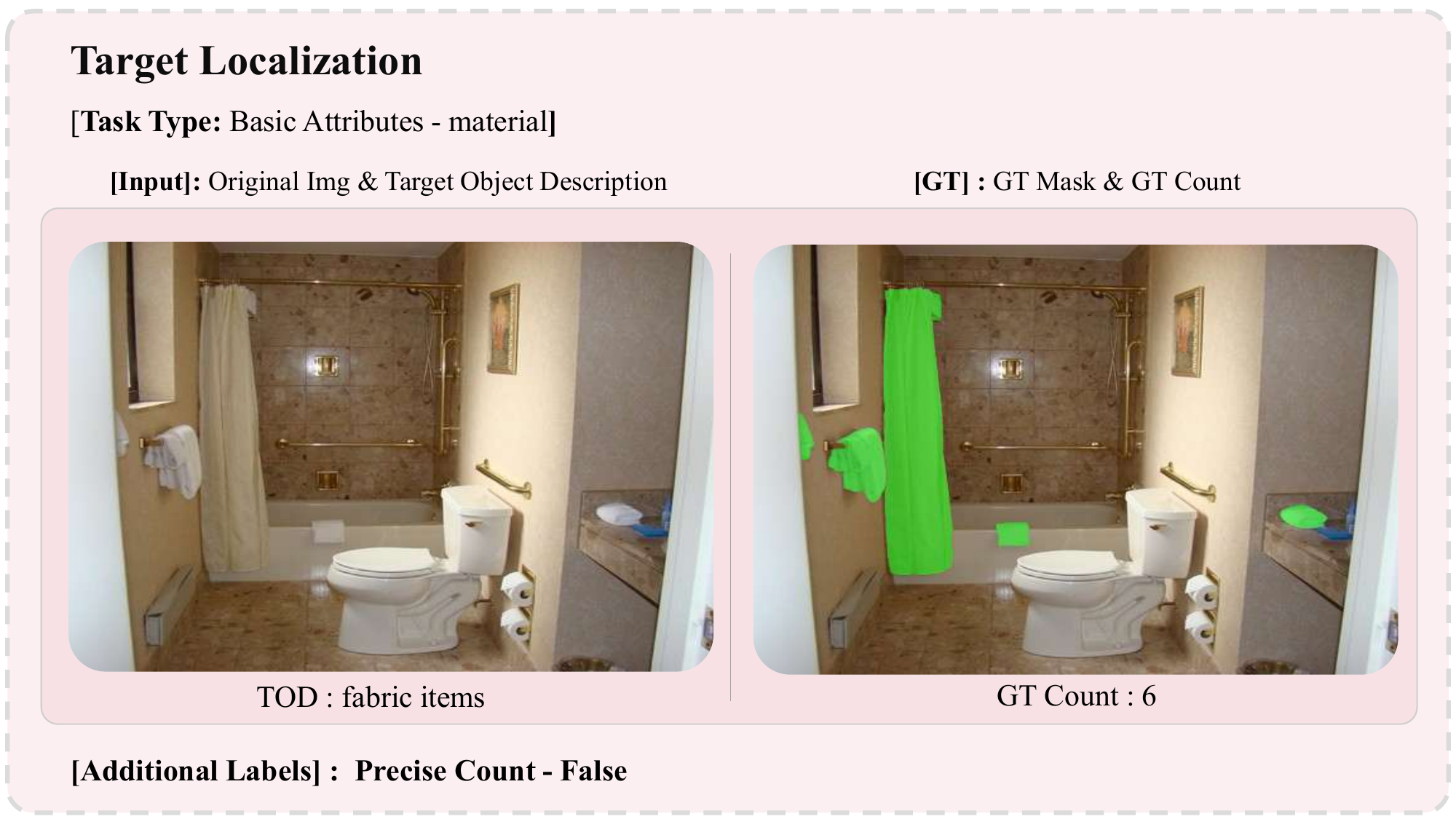}

\includegraphics[width=0.49511\textwidth]{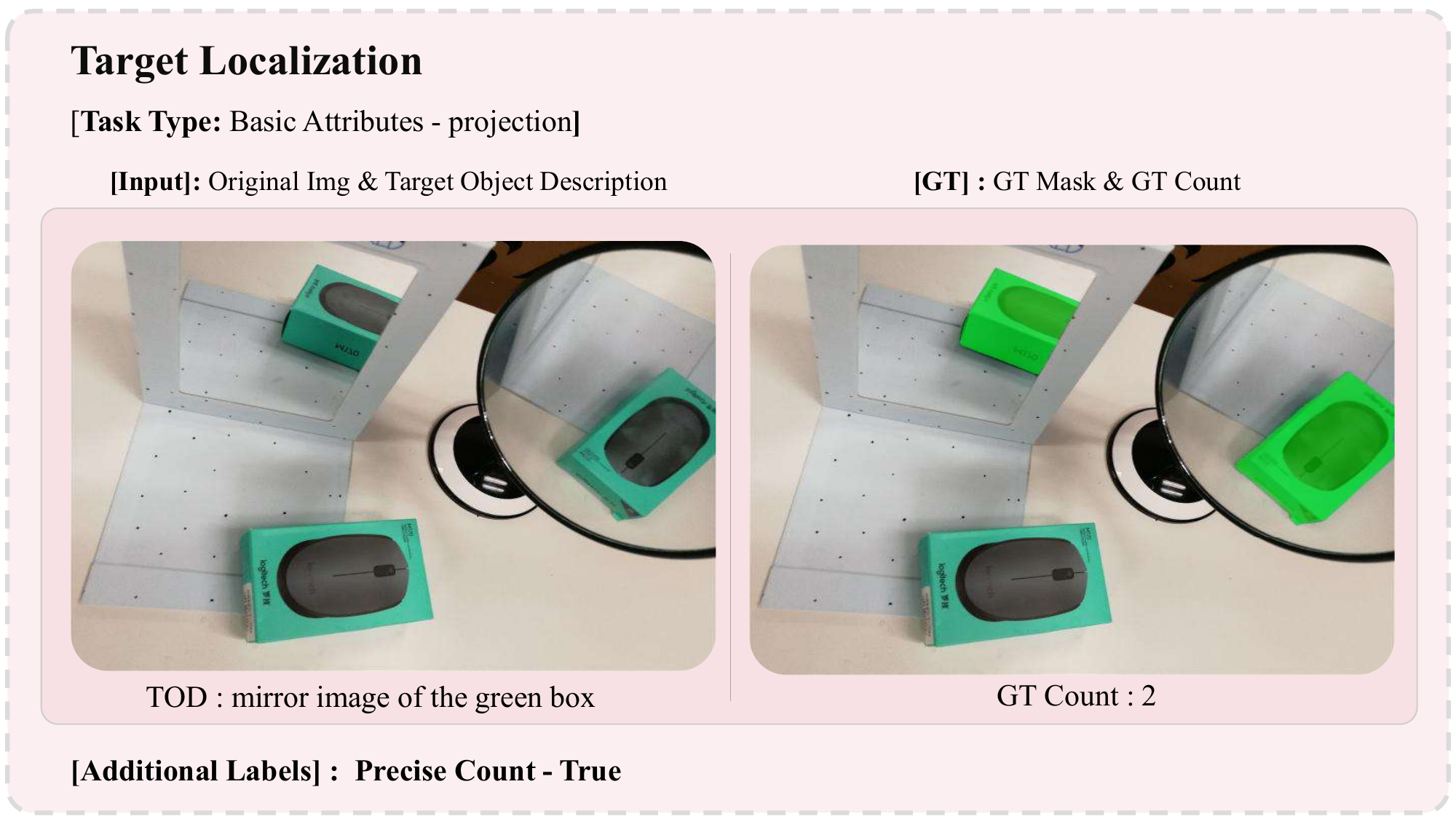}
\includegraphics[width=0.495\textwidth]{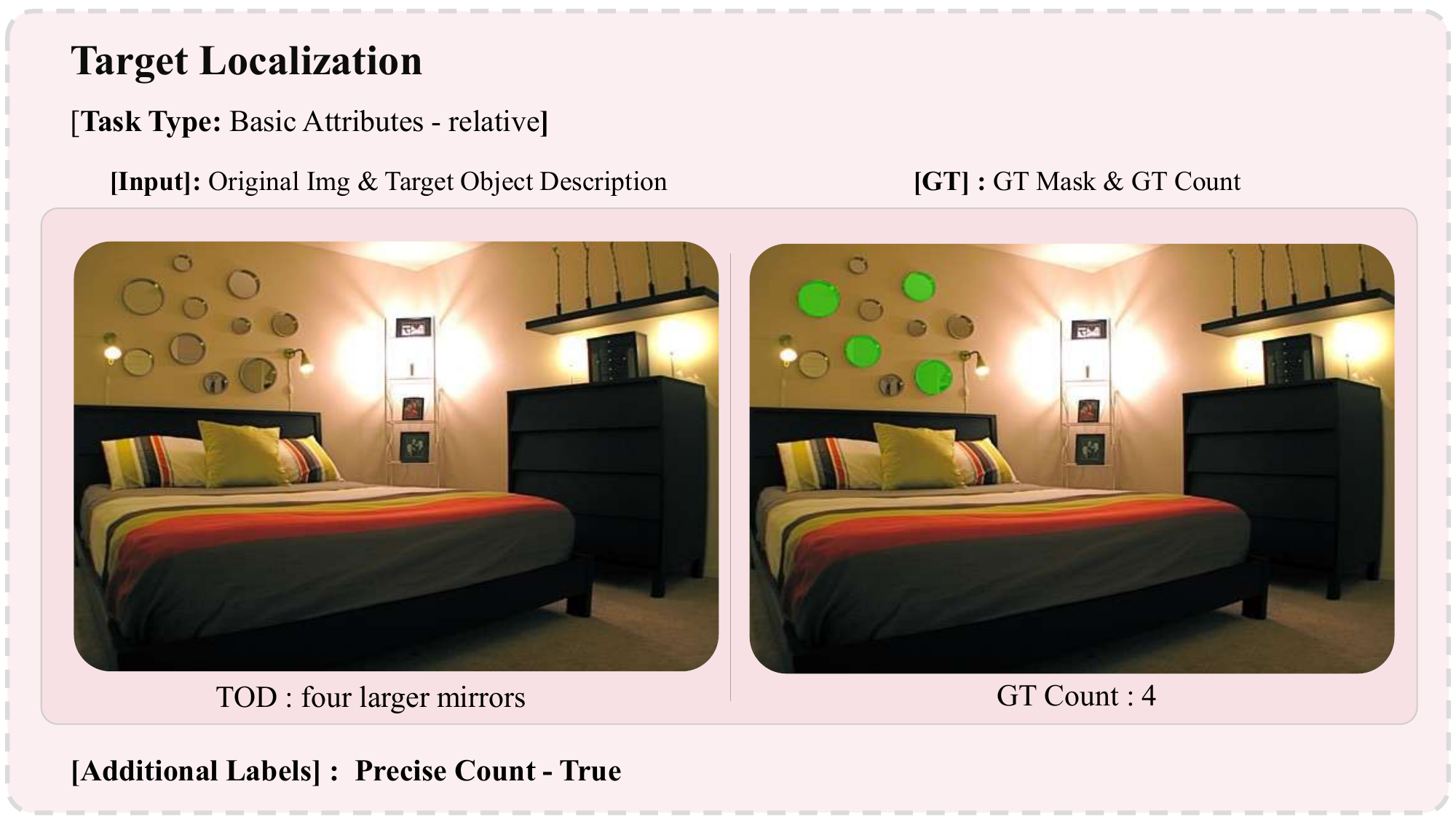}

\caption{Examples of Target Localization - Basic Attributes task.}
\label{fig:example_TL_BA}
\end{figure*}

\begin{figure*}[tb]
\centering
\includegraphics[width=0.495\textwidth]{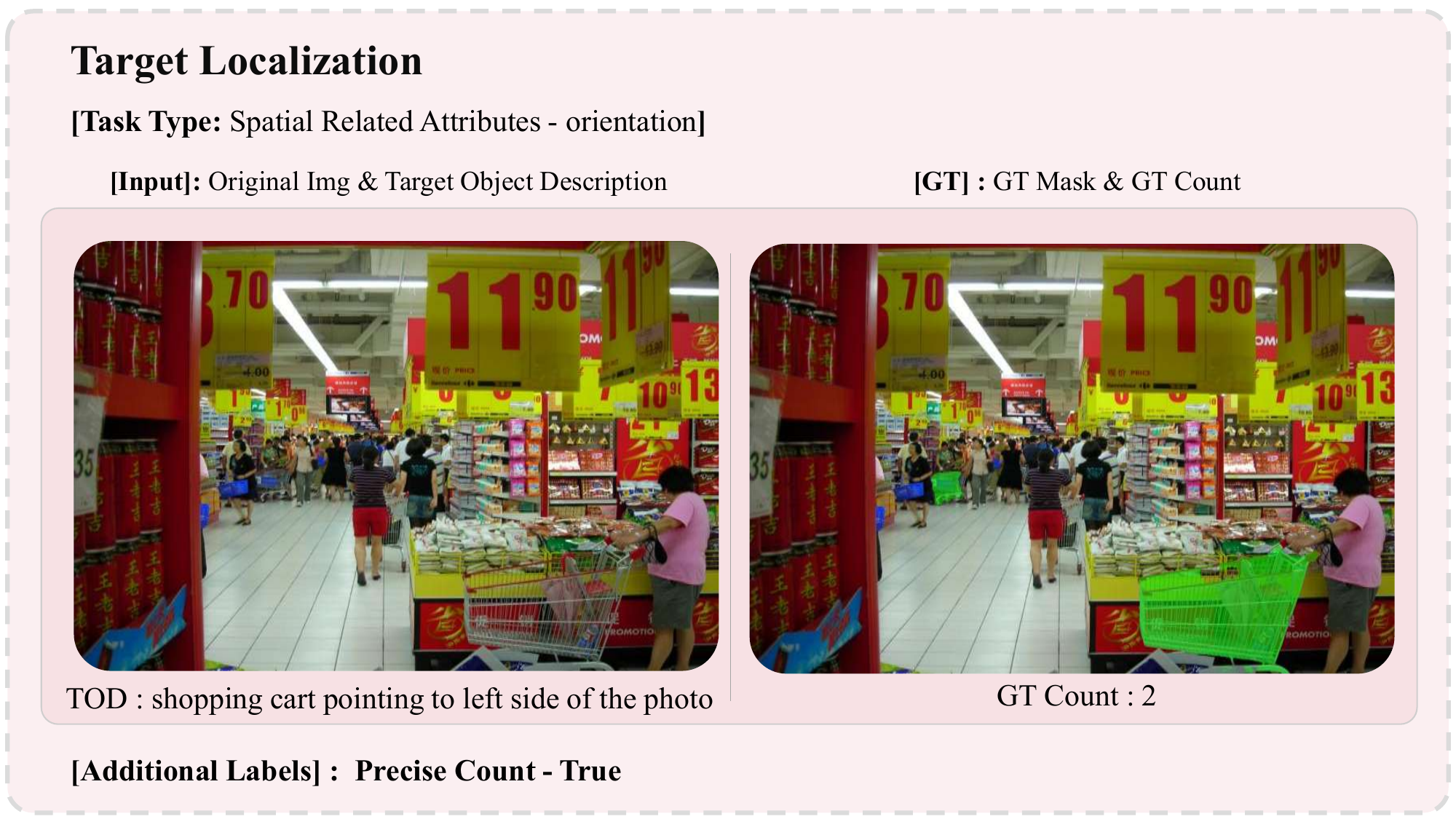}
\includegraphics[width=0.495\textwidth]{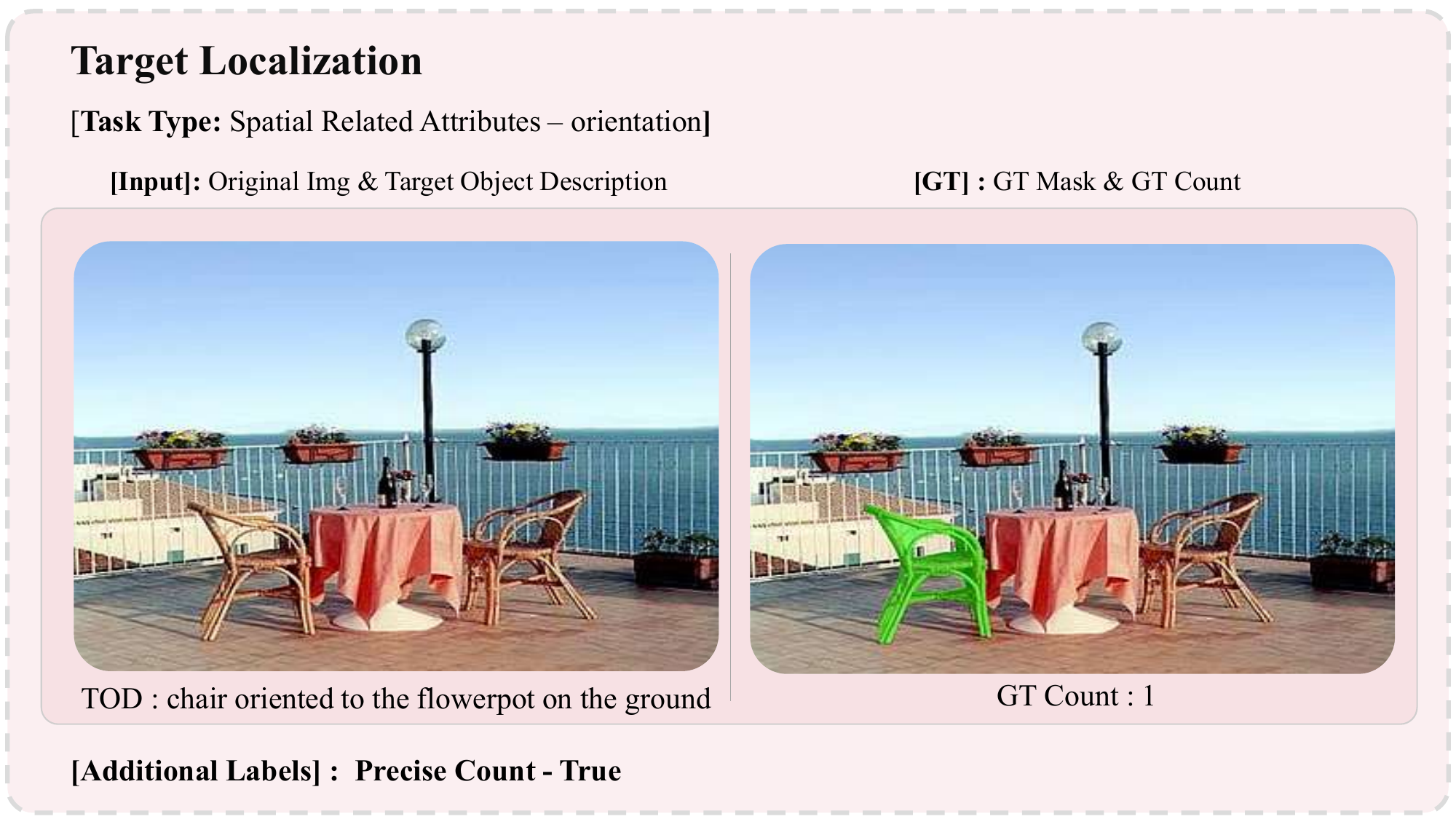}

\includegraphics[width=0.495\textwidth]{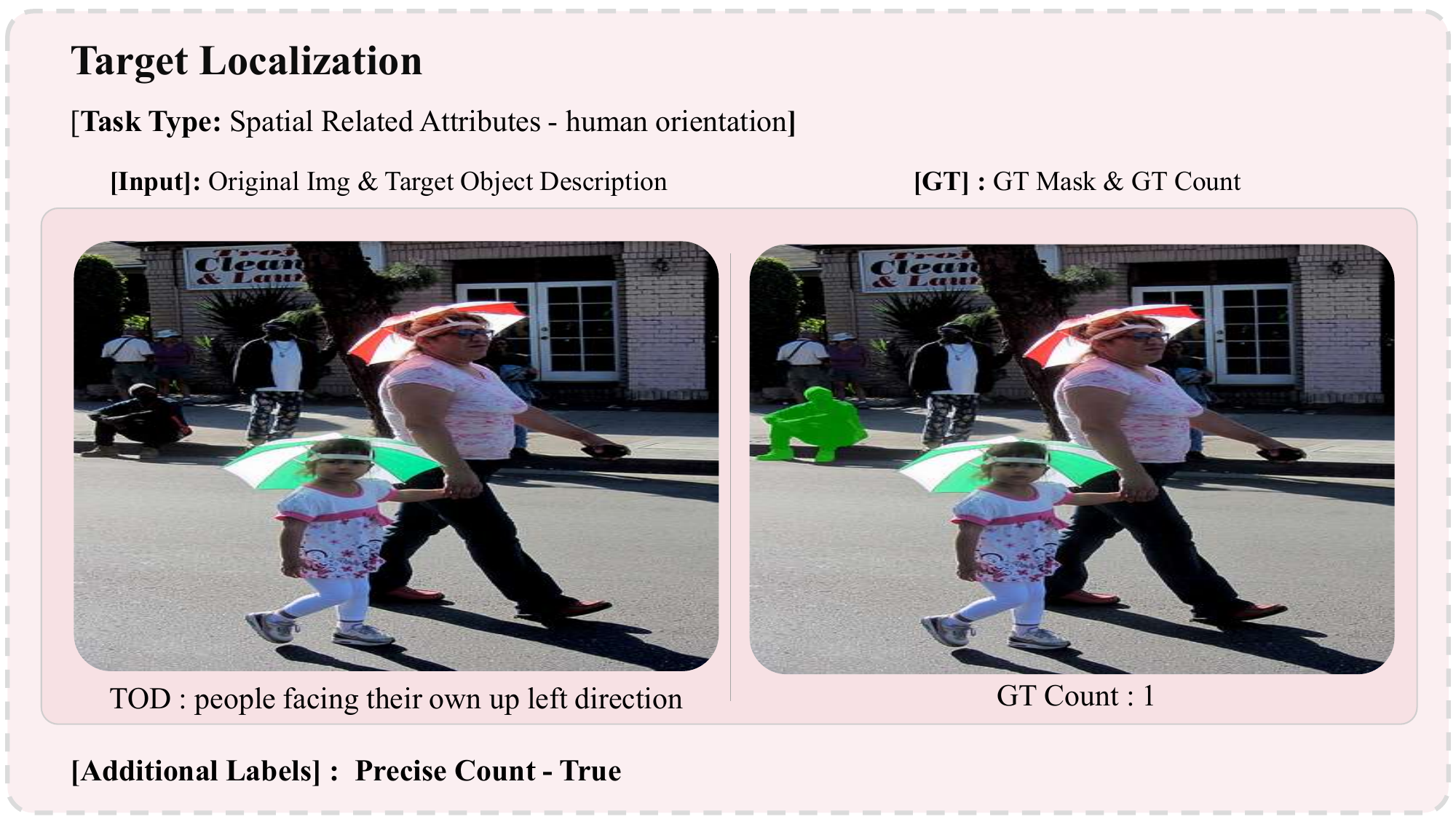}
\includegraphics[width=0.495\textwidth]{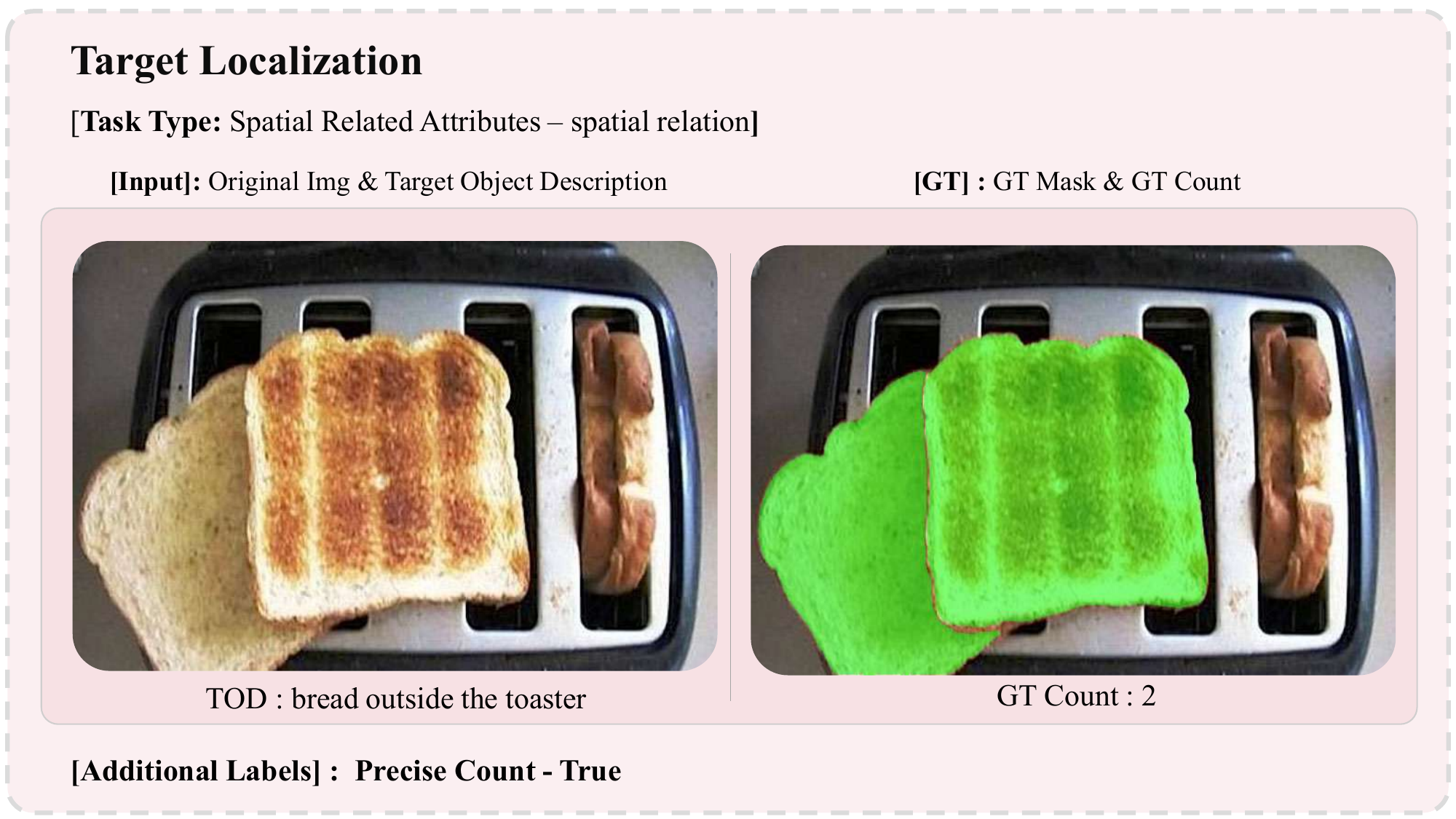}

\includegraphics[width=0.495\textwidth]{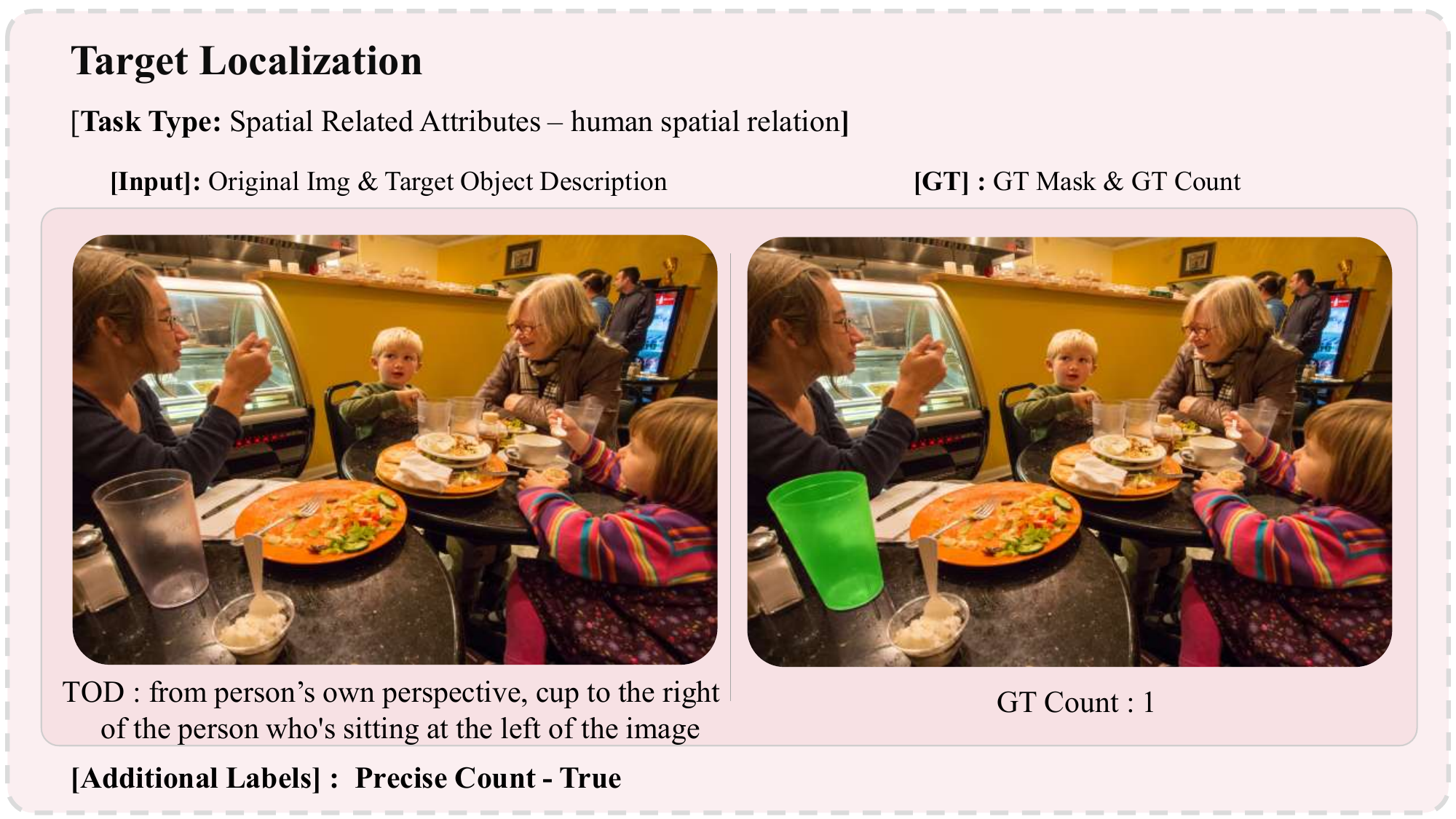}
\includegraphics[width=0.495\textwidth]{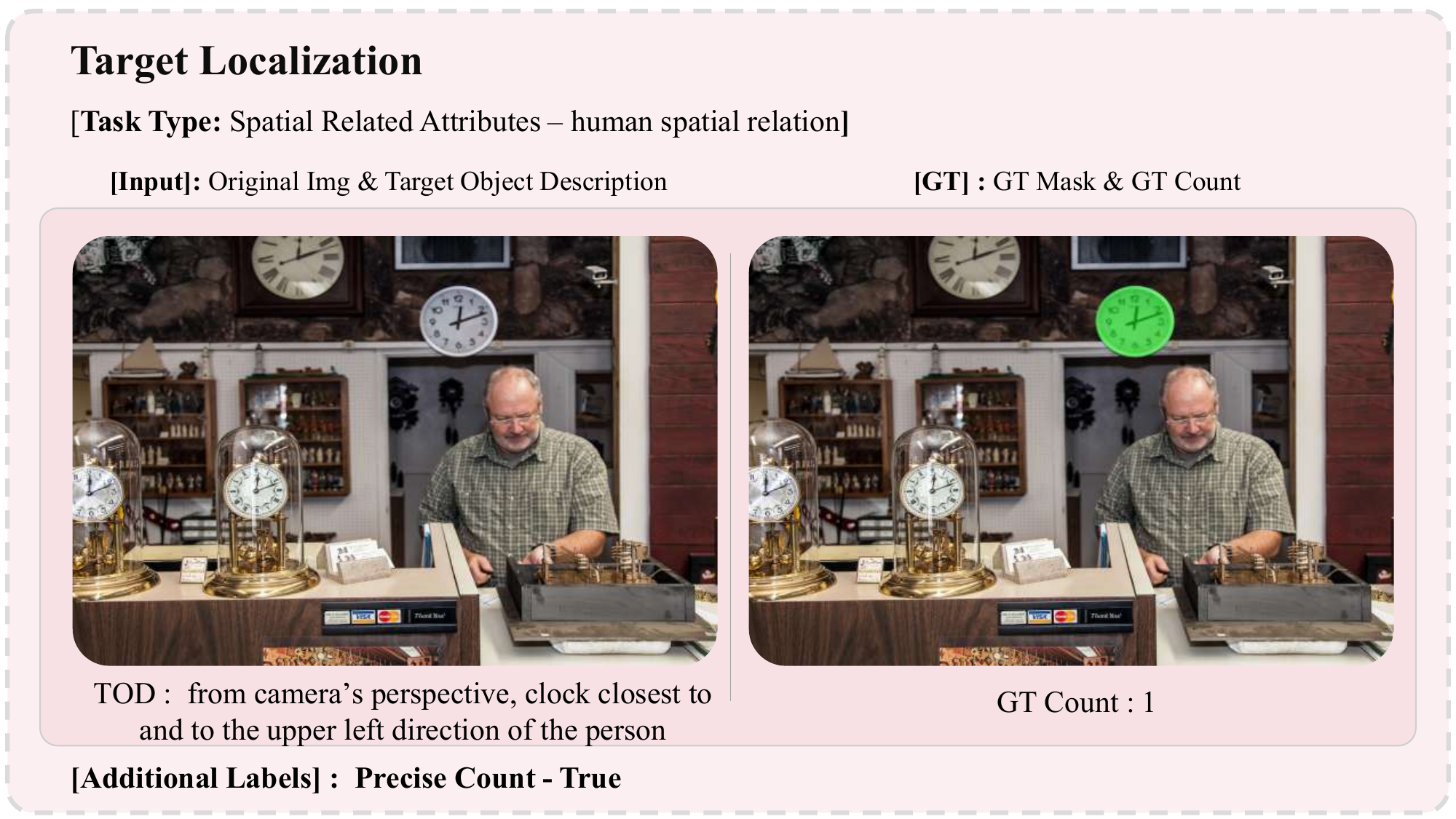}

\caption{Examples of Target Localization - Spatial Related Attributes task}
\label{fig:example_TL_SRA}
\end{figure*}

\begin{figure*}[tb]
\centering
\includegraphics[width=0.495\textwidth]{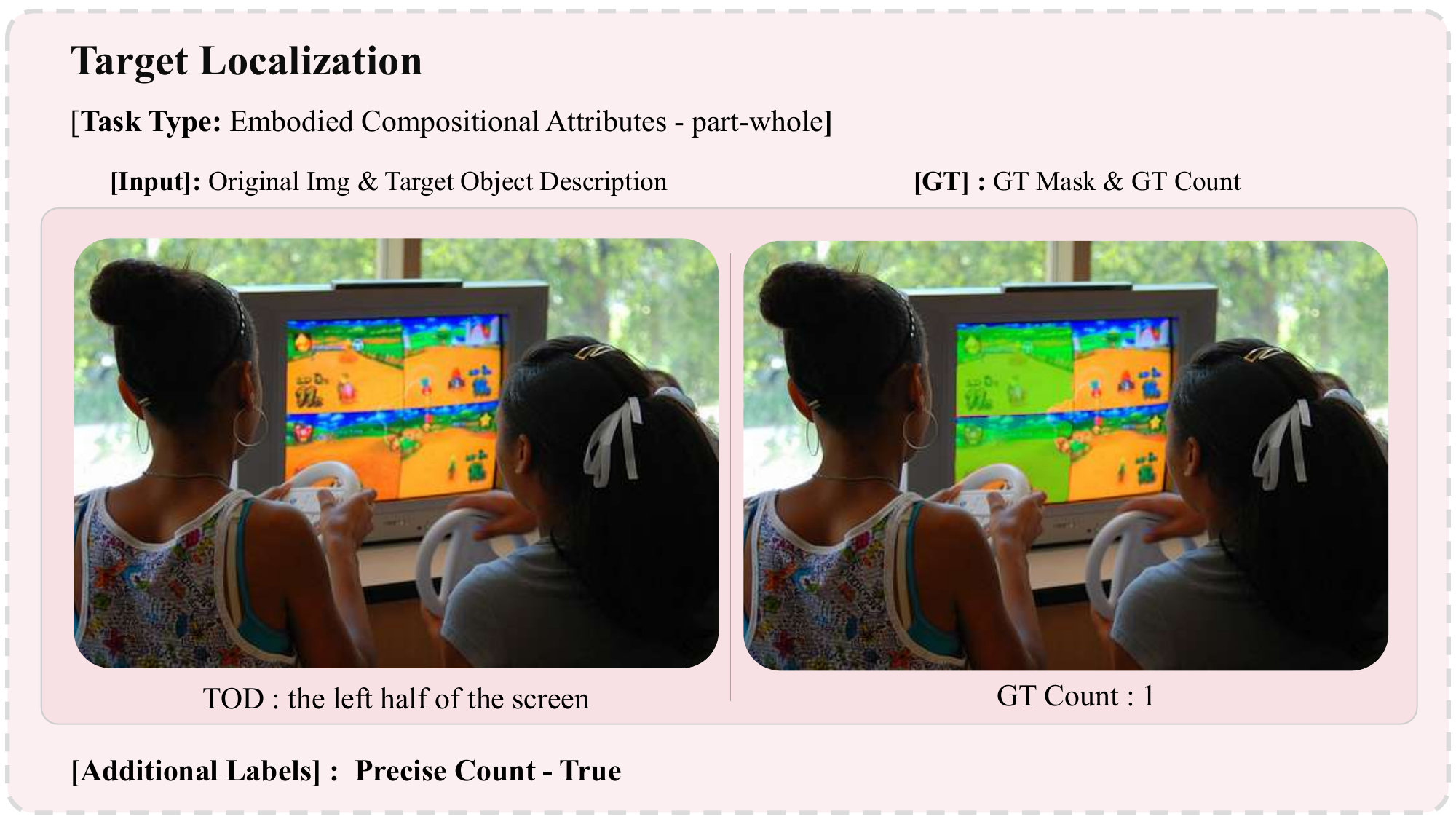}
\includegraphics[width=0.495\textwidth]{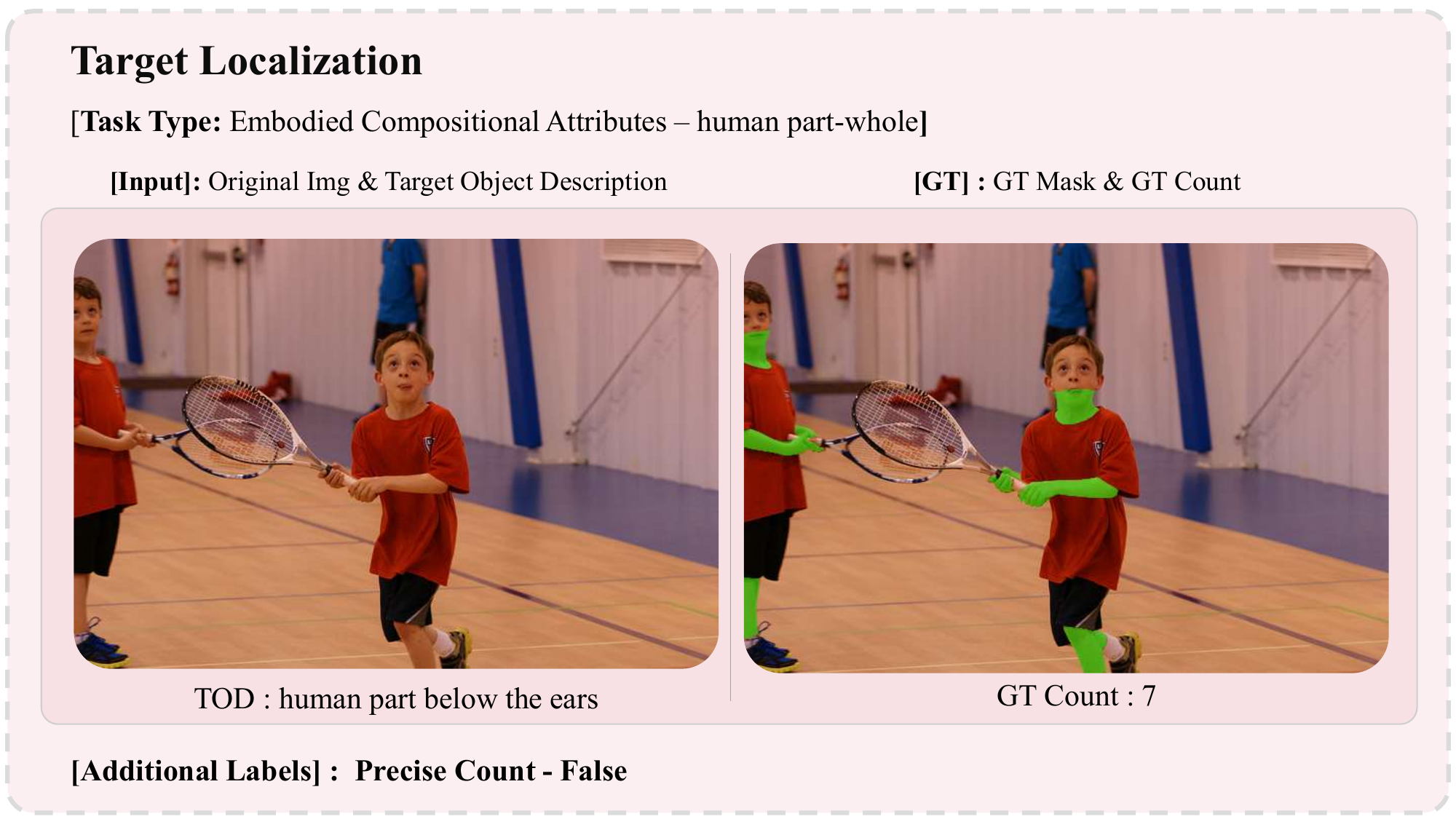}
\includegraphics[width=0.495\textwidth]{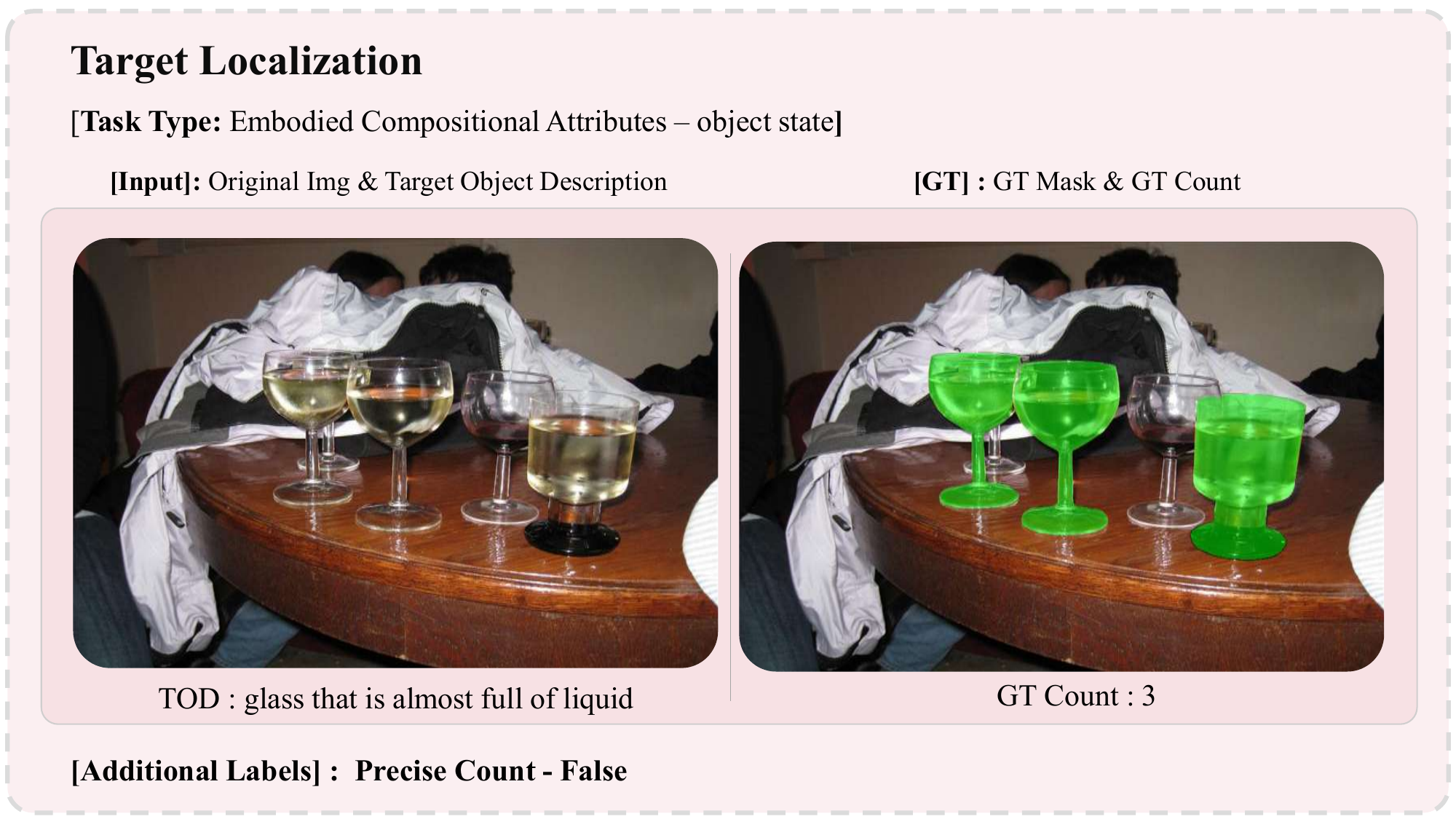}
\includegraphics[width=0.495\textwidth]{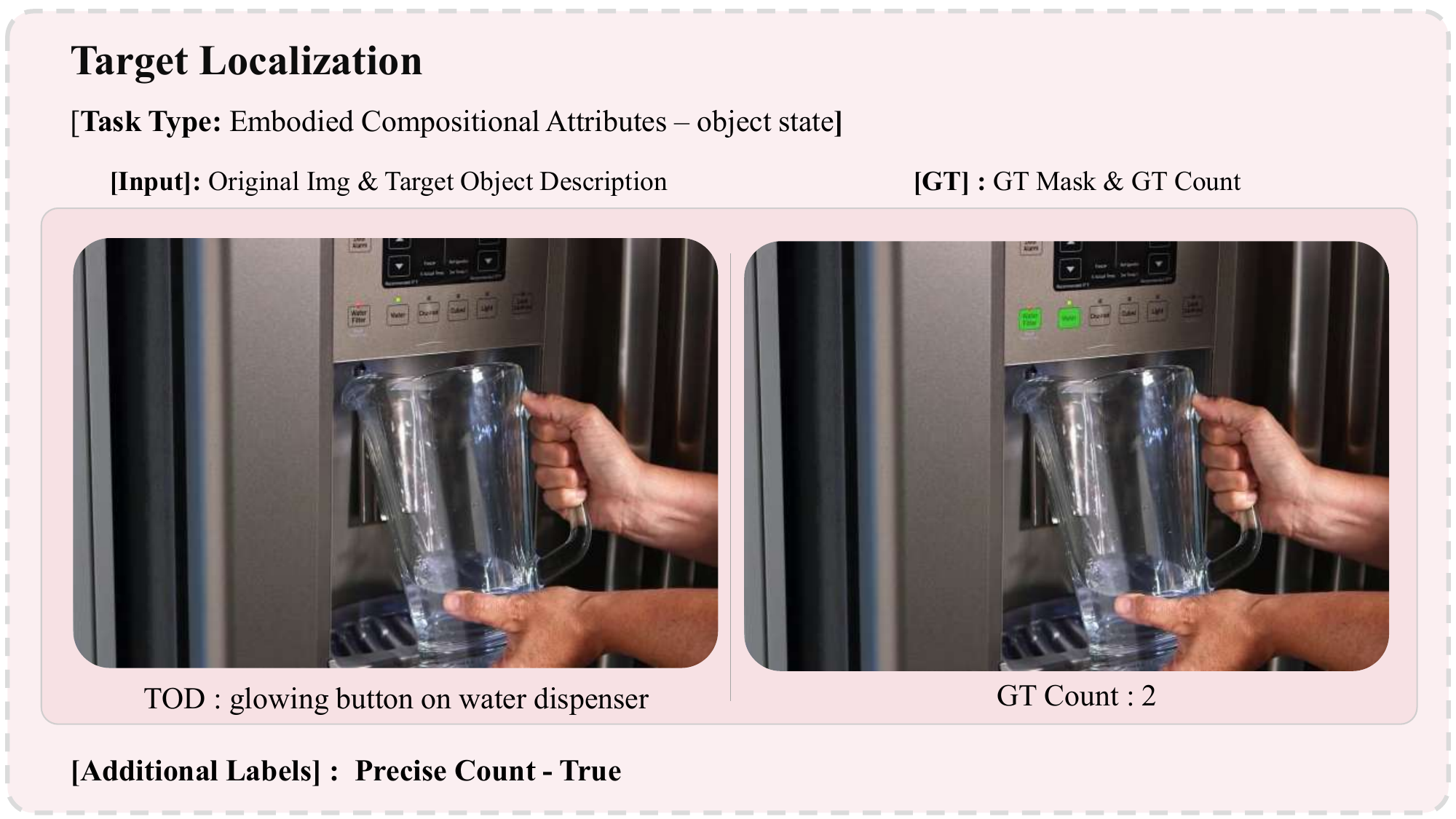}
\includegraphics[width=0.495\textwidth]{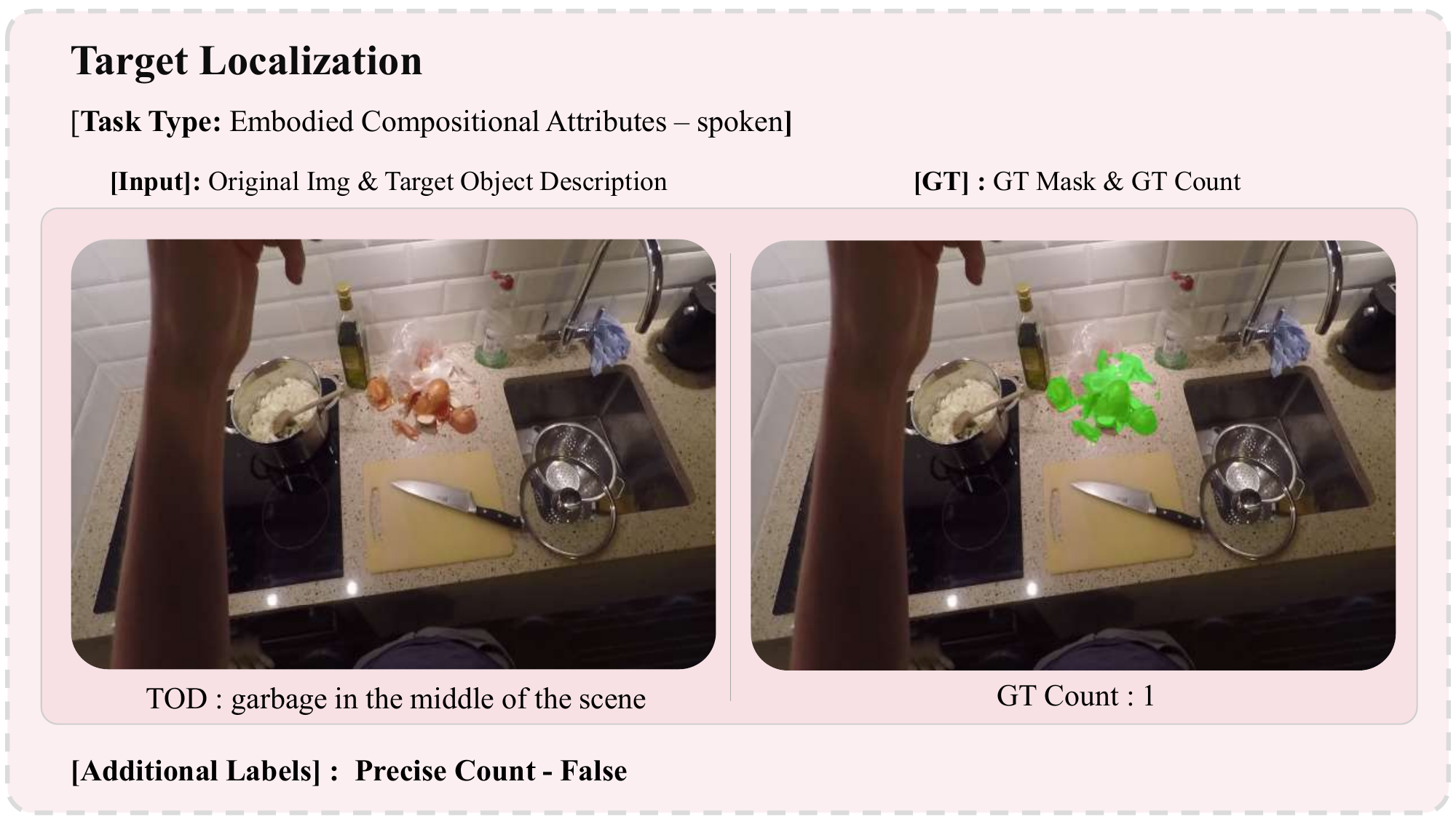}
\includegraphics[width=0.495\textwidth]{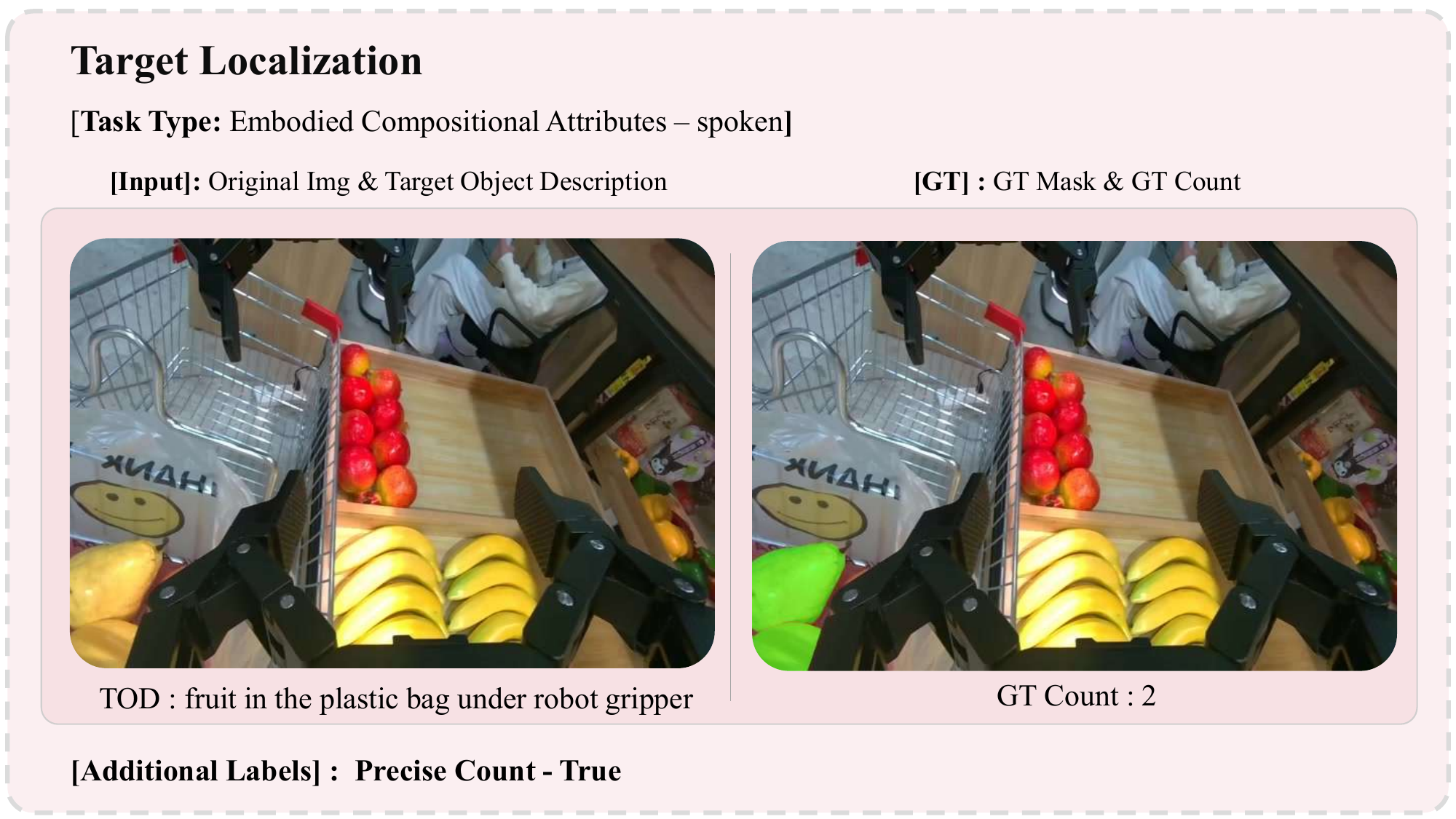}
\caption{Examples of Target Localization - Embodied Compositional Attributes task}
\label{fig:example_TL_ECA}
\end{figure*}

\paragraph{Target Localization} Figure~\ref{fig:example_TL_BA}, \ref{fig:example_TL_SRA}, and \ref{fig:example_TL_ECA} present examples from the Target Localization category, including the three task types of Basic Attributes, Spatial-Related Attributes, and Embodied Compositional Attributes.

\begin{figure*}[tb]
\centering
\includegraphics[width=0.495\textwidth]{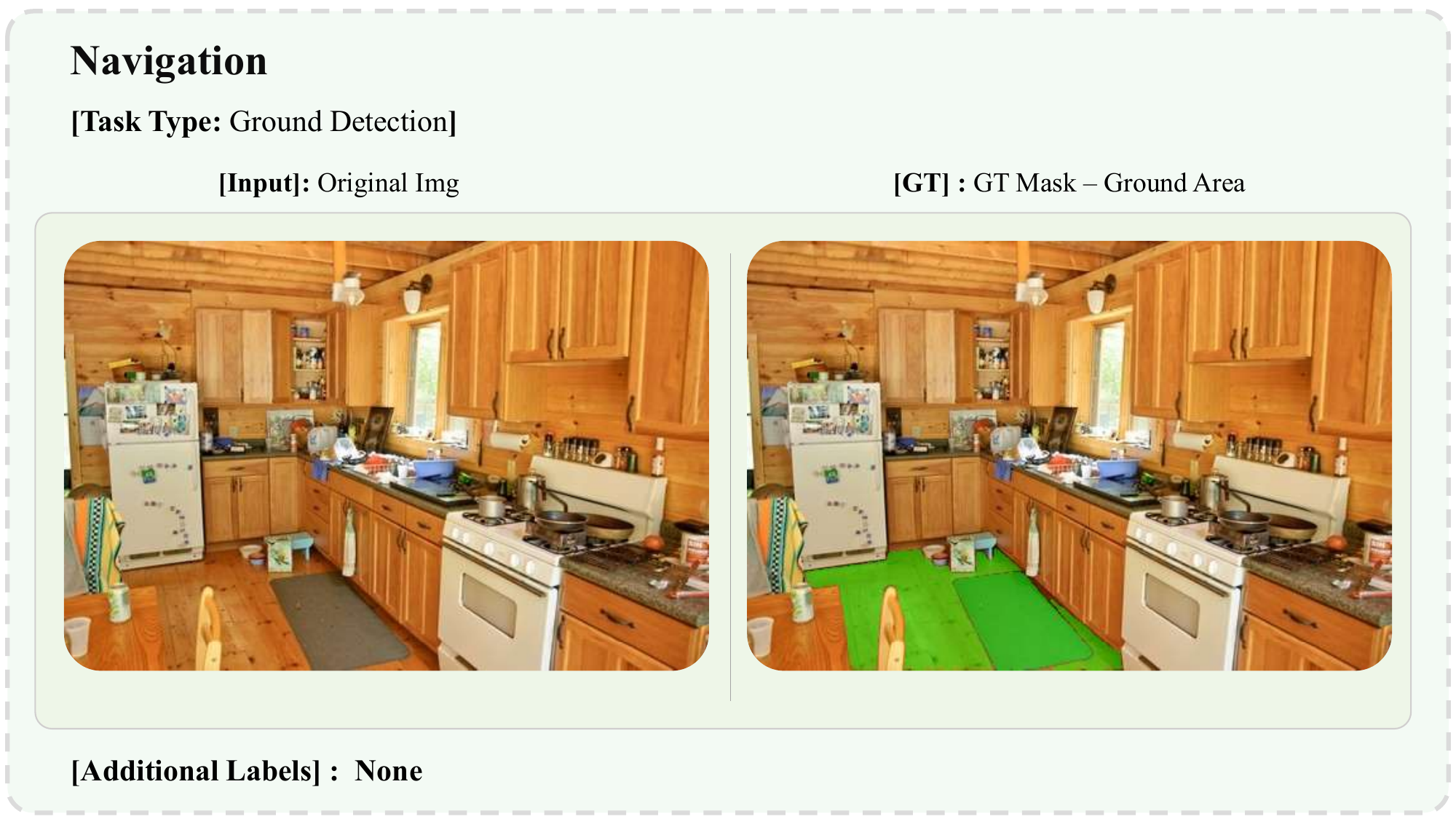}
\includegraphics[width=0.495\textwidth]{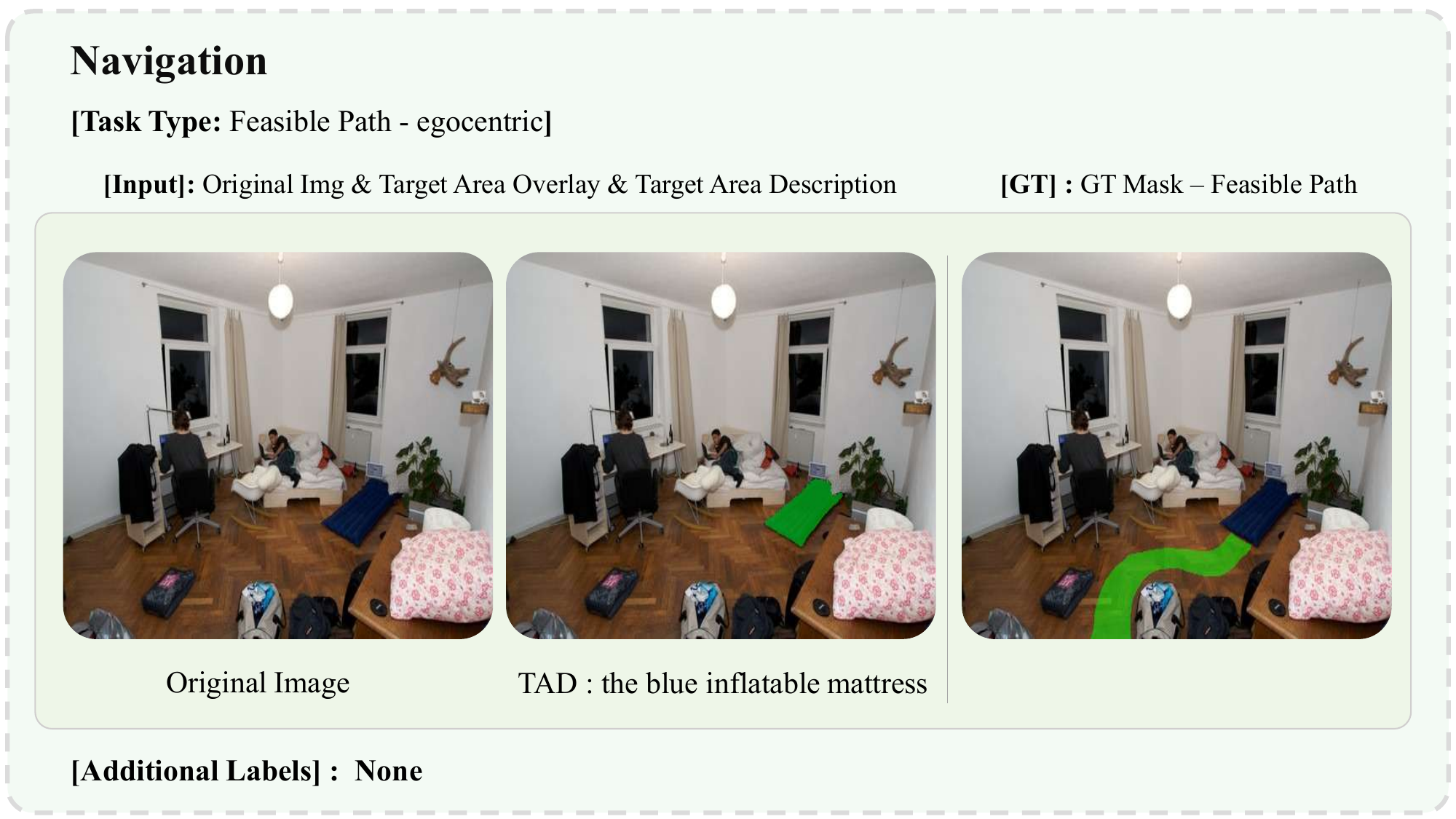}
\includegraphics[width=0.495\textwidth]{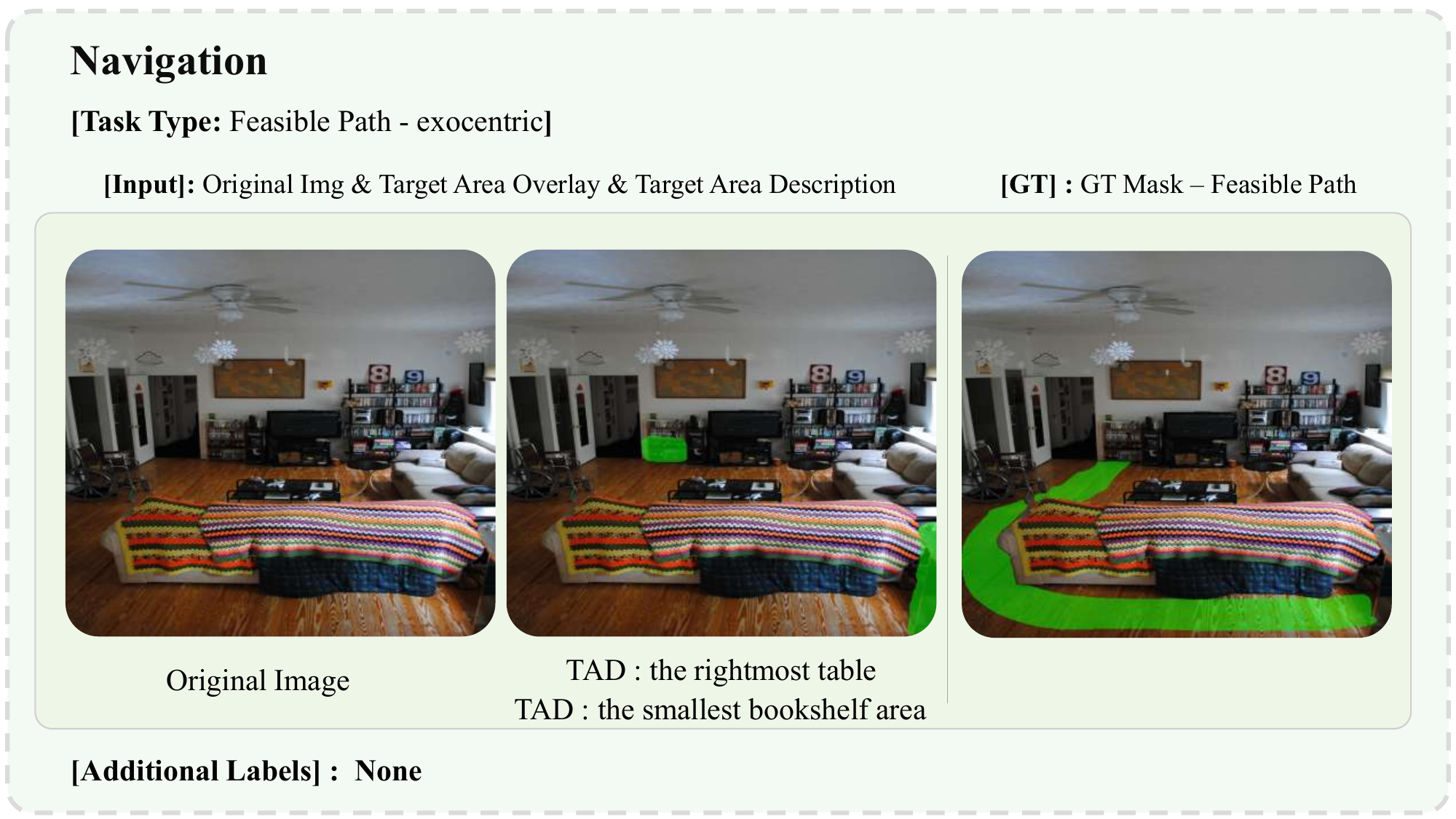}
\includegraphics[width=0.495\textwidth]{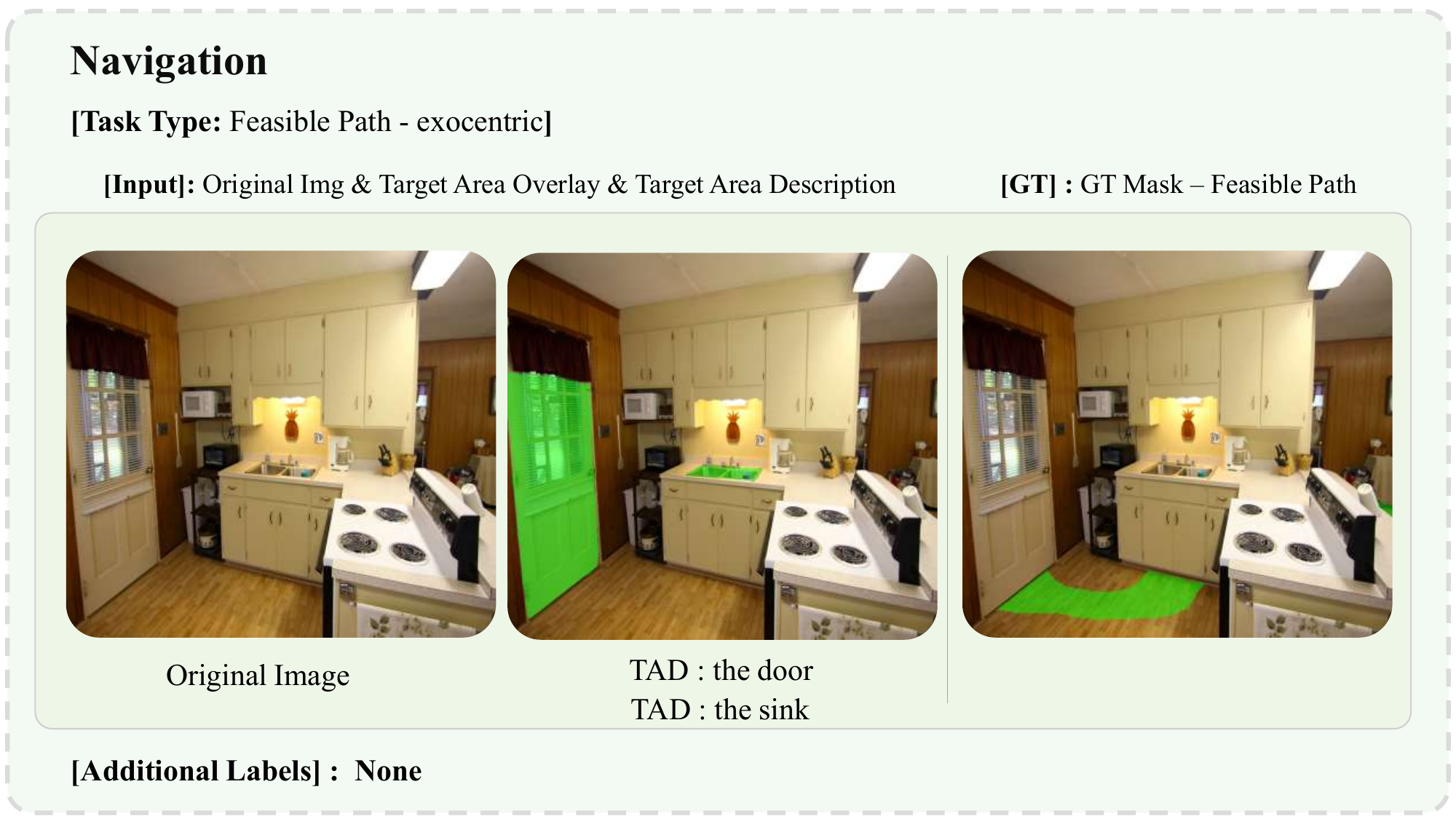}
\includegraphics[width=0.495\textwidth]{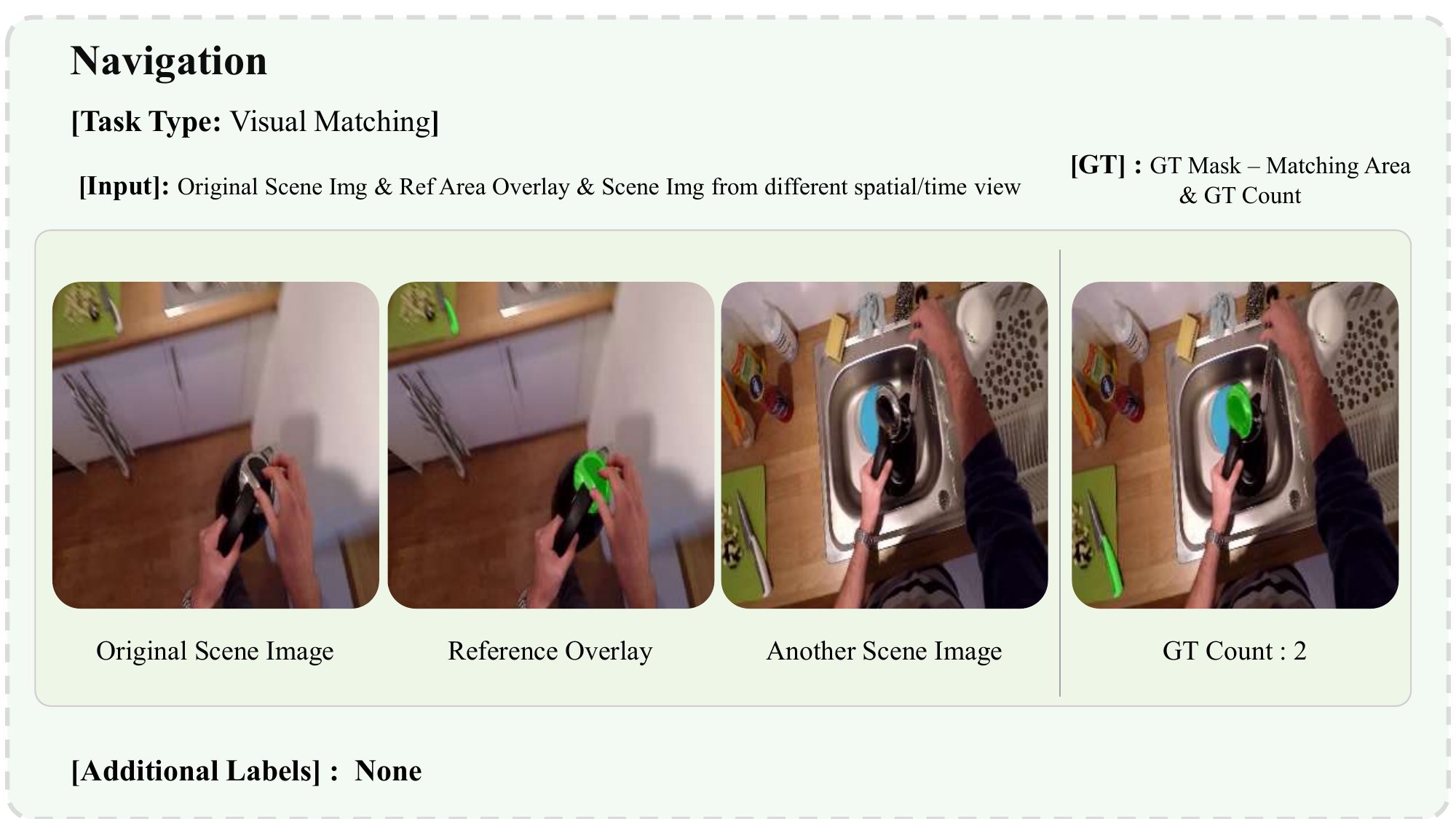}
\includegraphics[width=0.495\textwidth]{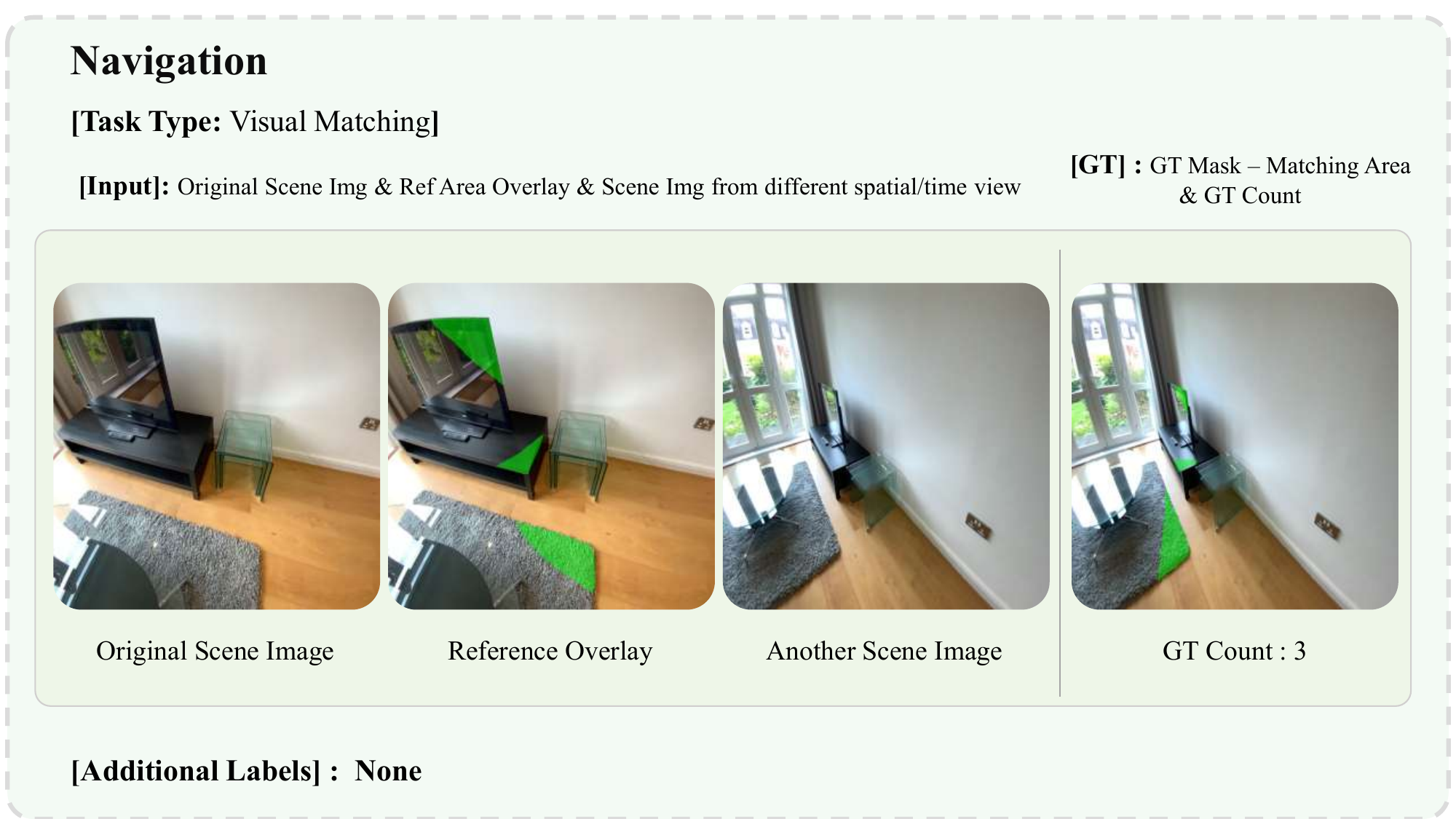}

\caption{Examples of Navigation tasks}
\label{fig:example_NAV}
\end{figure*}

\paragraph{Navigation} Figure \ref{fig:example_NAV} illustrates the Navigation category, which includes the tasks of Ground Detection, Feasible Path Recognition, and Visual Matching.

\begin{figure*}[tb]
\centering
\includegraphics[width=0.495\textwidth]{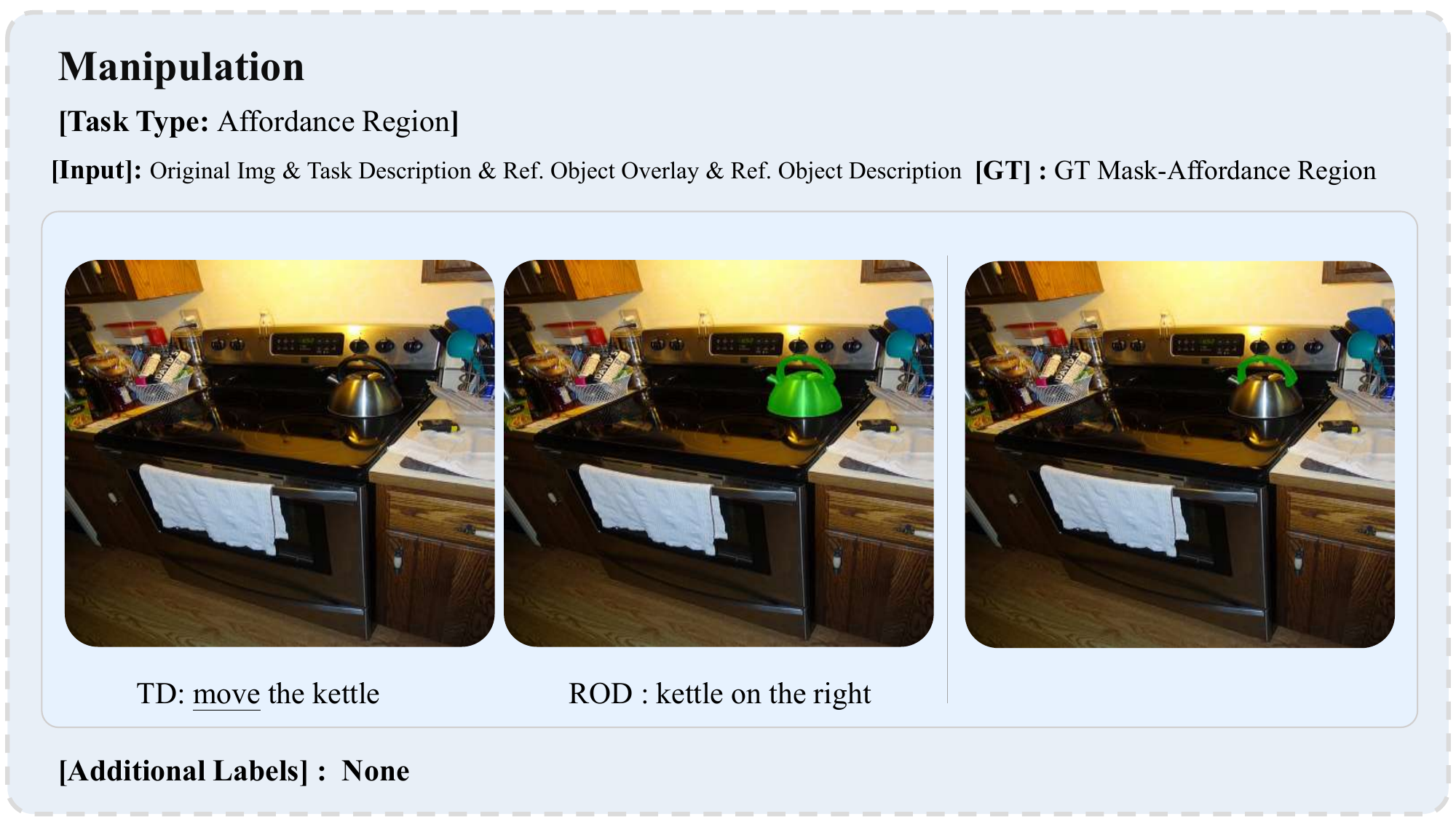}
\includegraphics[width=0.495\textwidth]{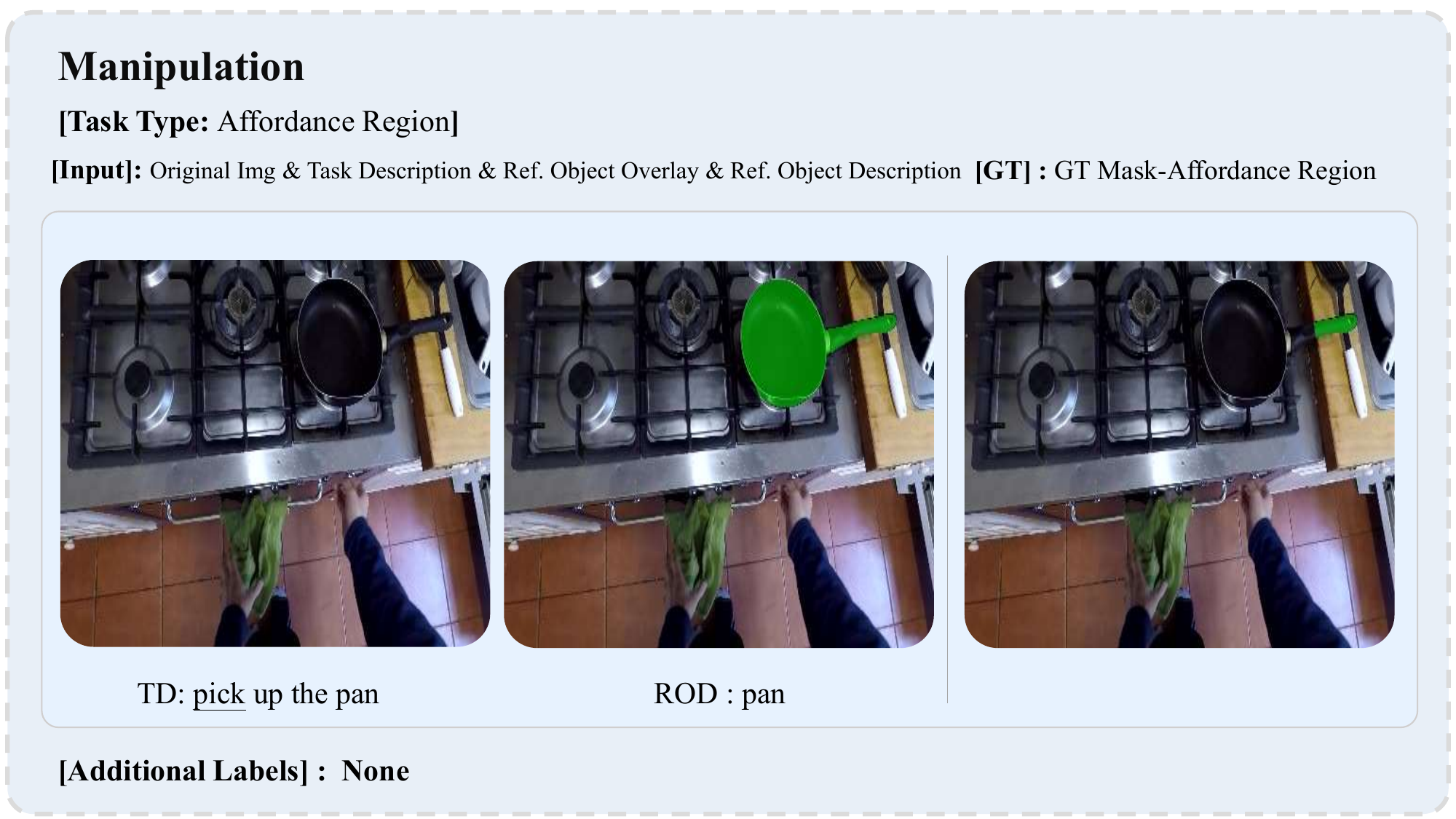}

\includegraphics[width=0.495\textwidth]{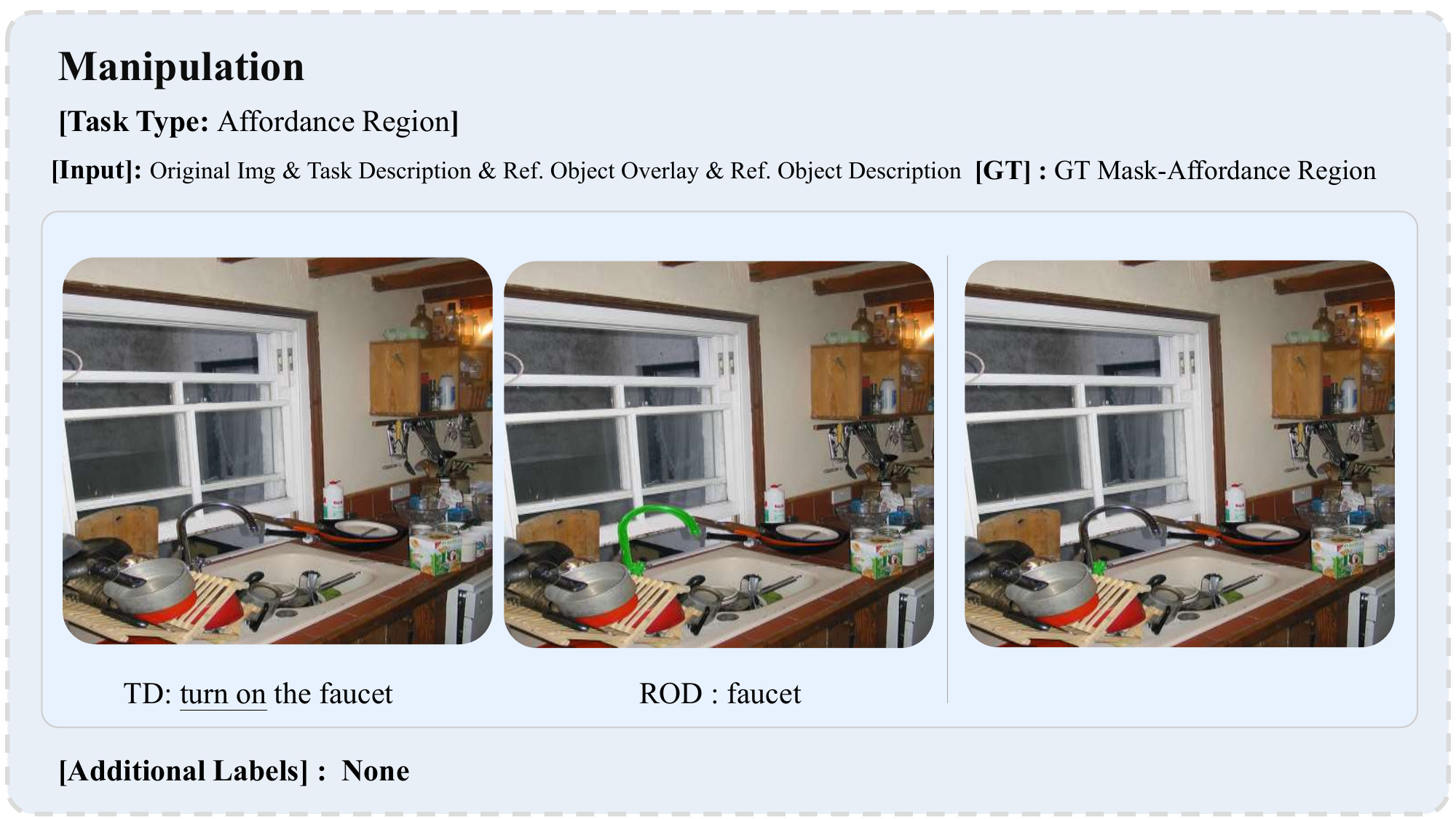}
\includegraphics[width=0.495\textwidth]{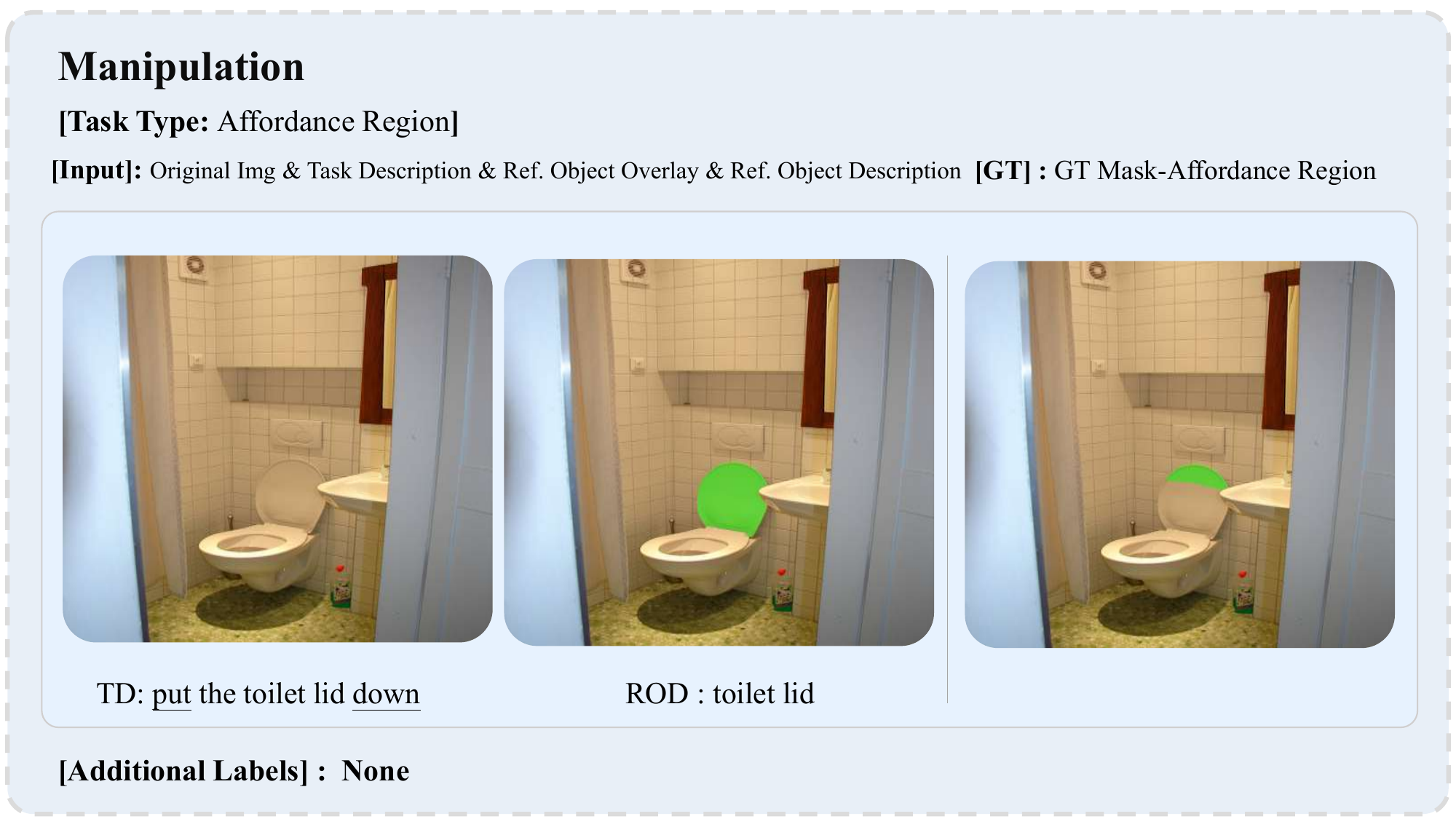}

\includegraphics[width=0.495\textwidth]{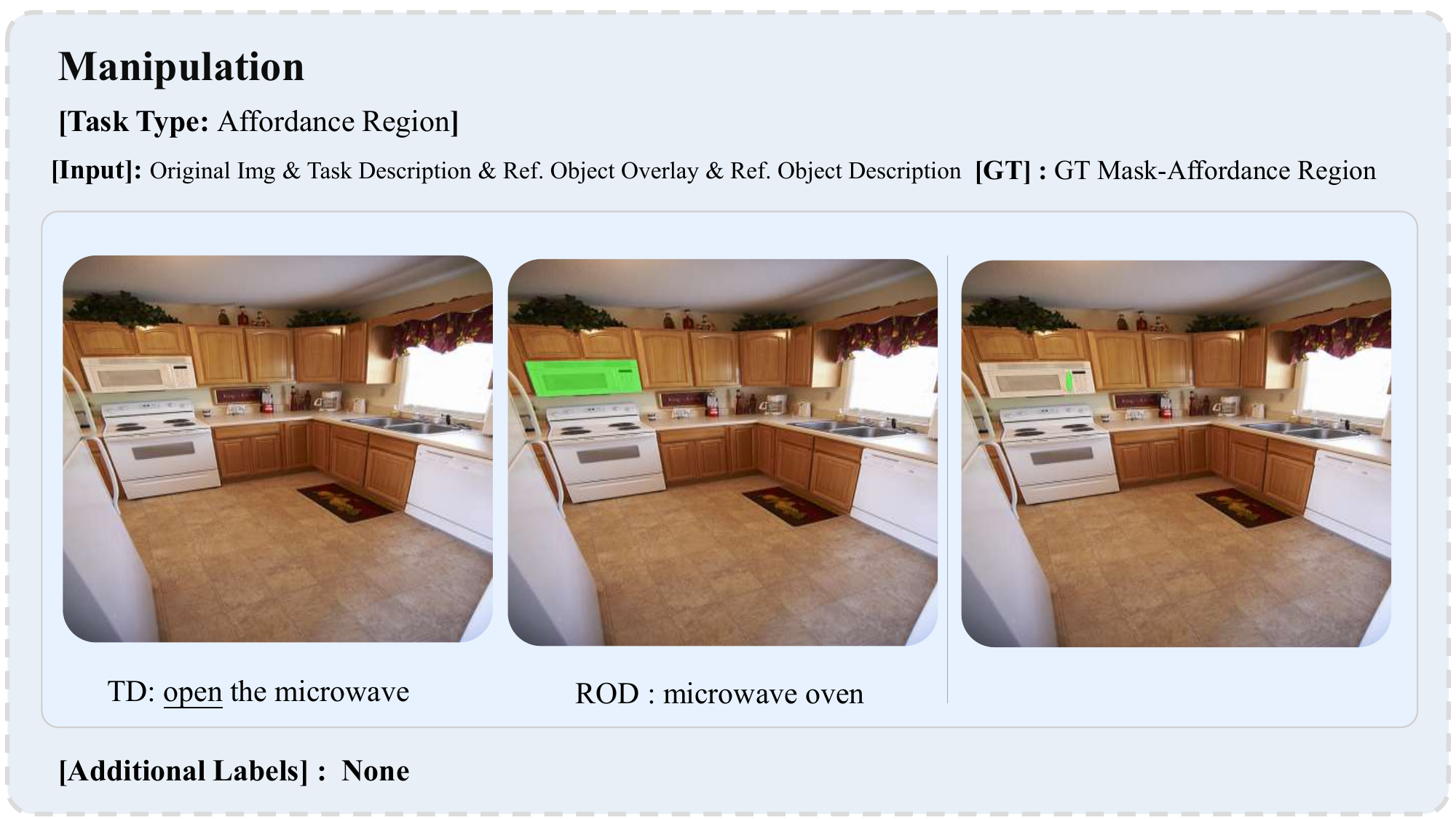}
\includegraphics[width=0.495\textwidth]{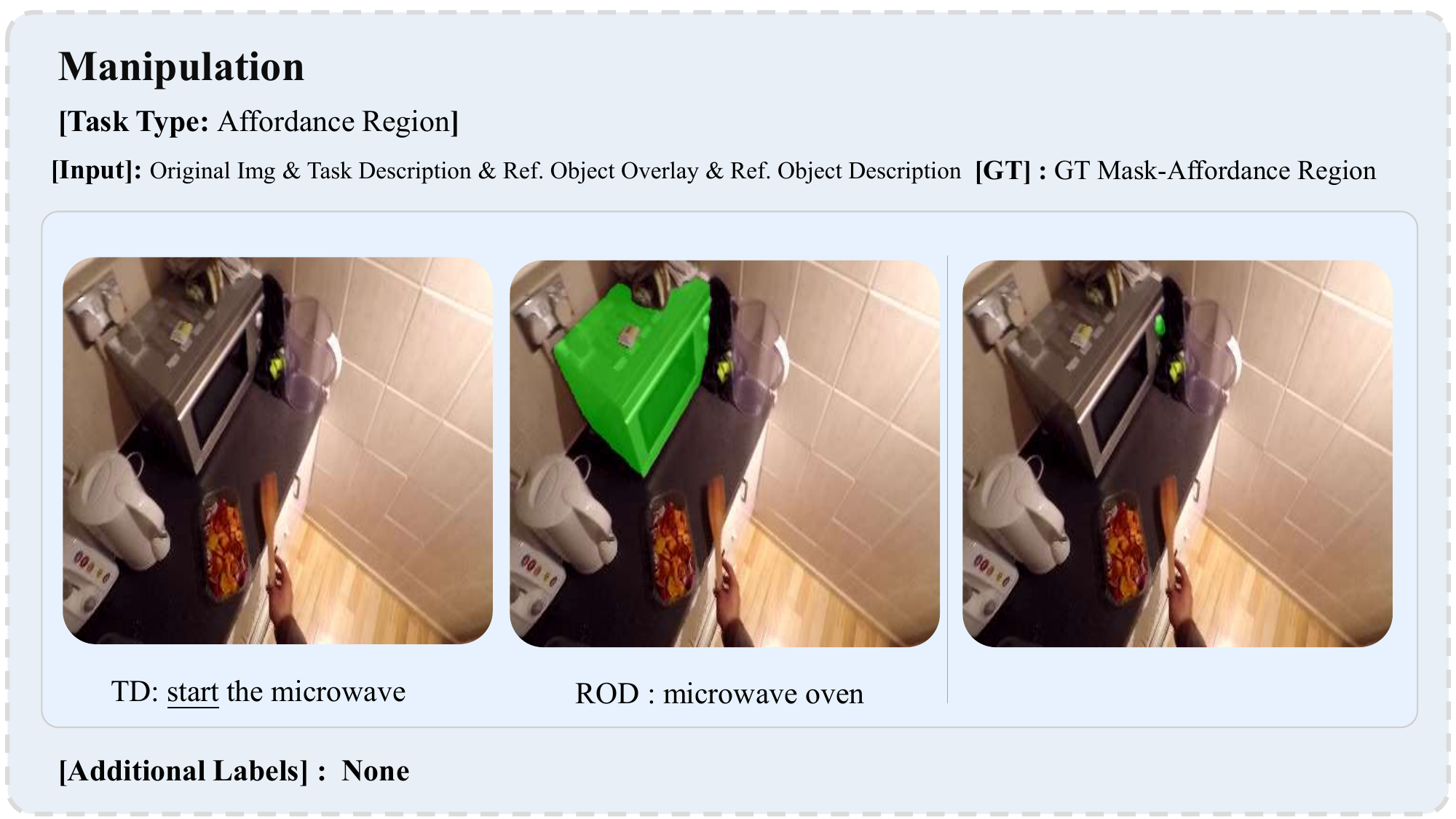}

\caption{Examples of Manipulation - Affordance Region task}
\label{fig:example_MAN_AR}
\end{figure*}

\begin{figure*}[tb]
\centering
\includegraphics[width=0.495\textwidth]{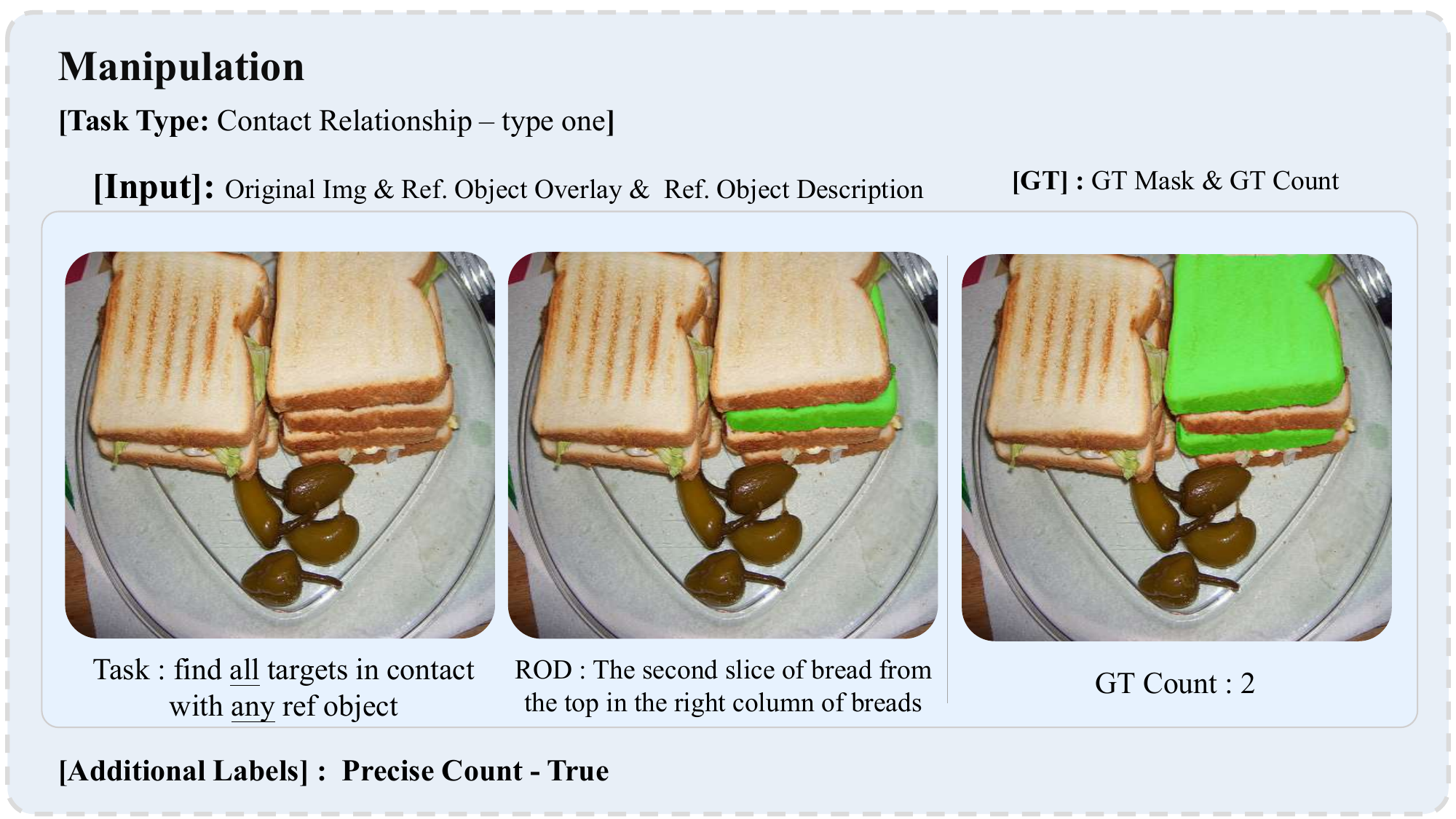}
\includegraphics[width=0.495\textwidth]{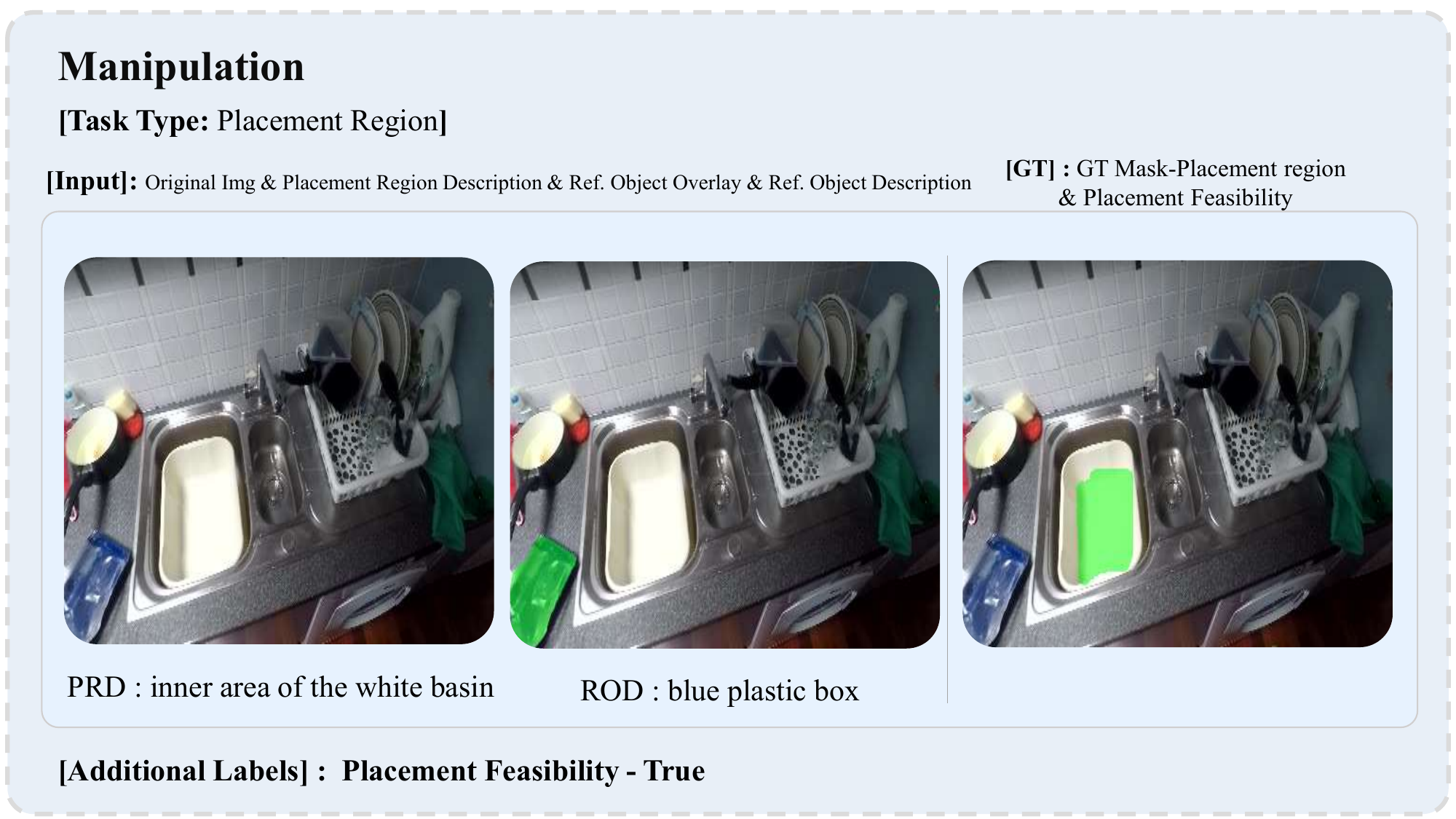}

\includegraphics[width=0.495\textwidth]{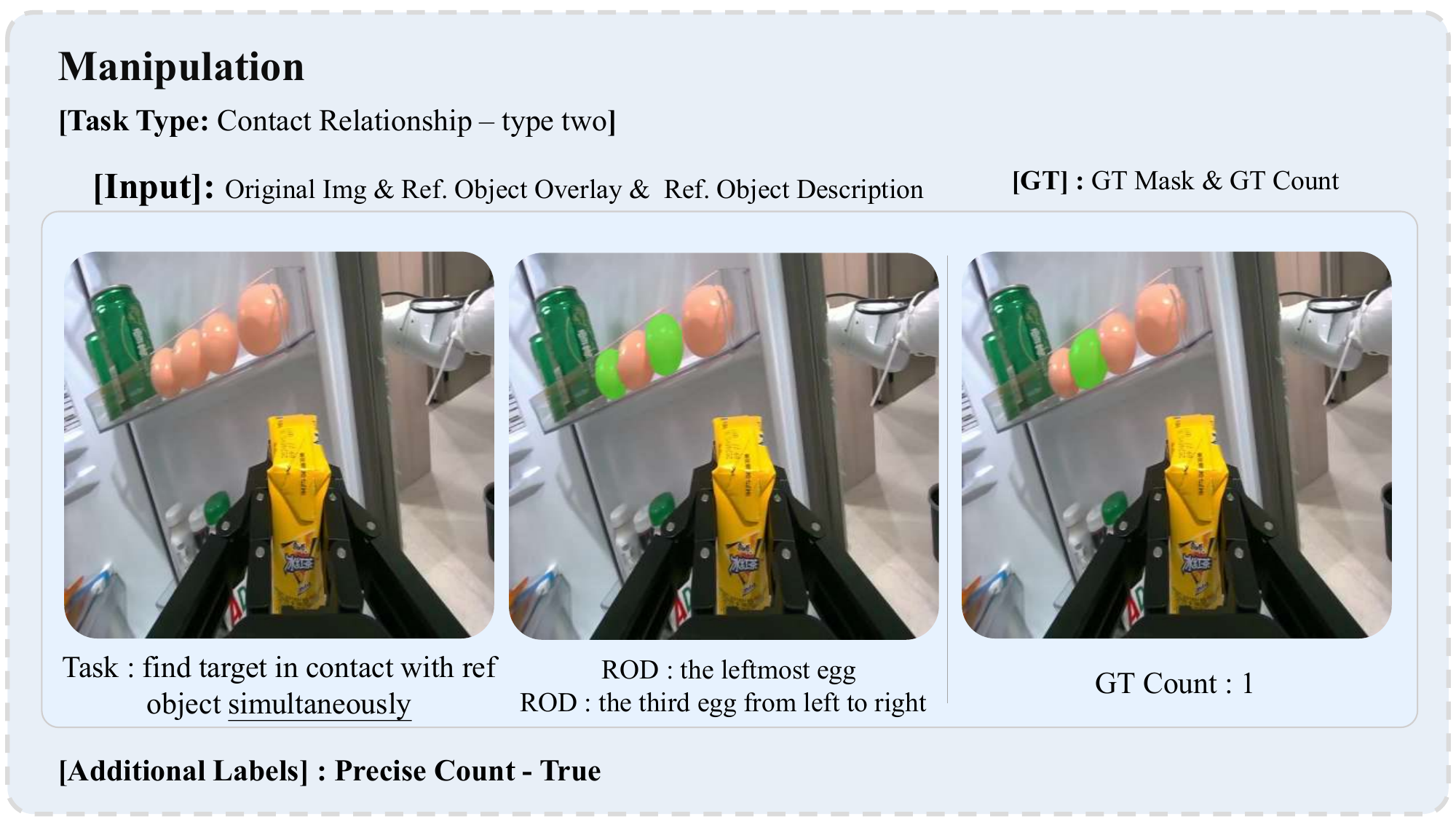}
\includegraphics[width=0.495\textwidth]{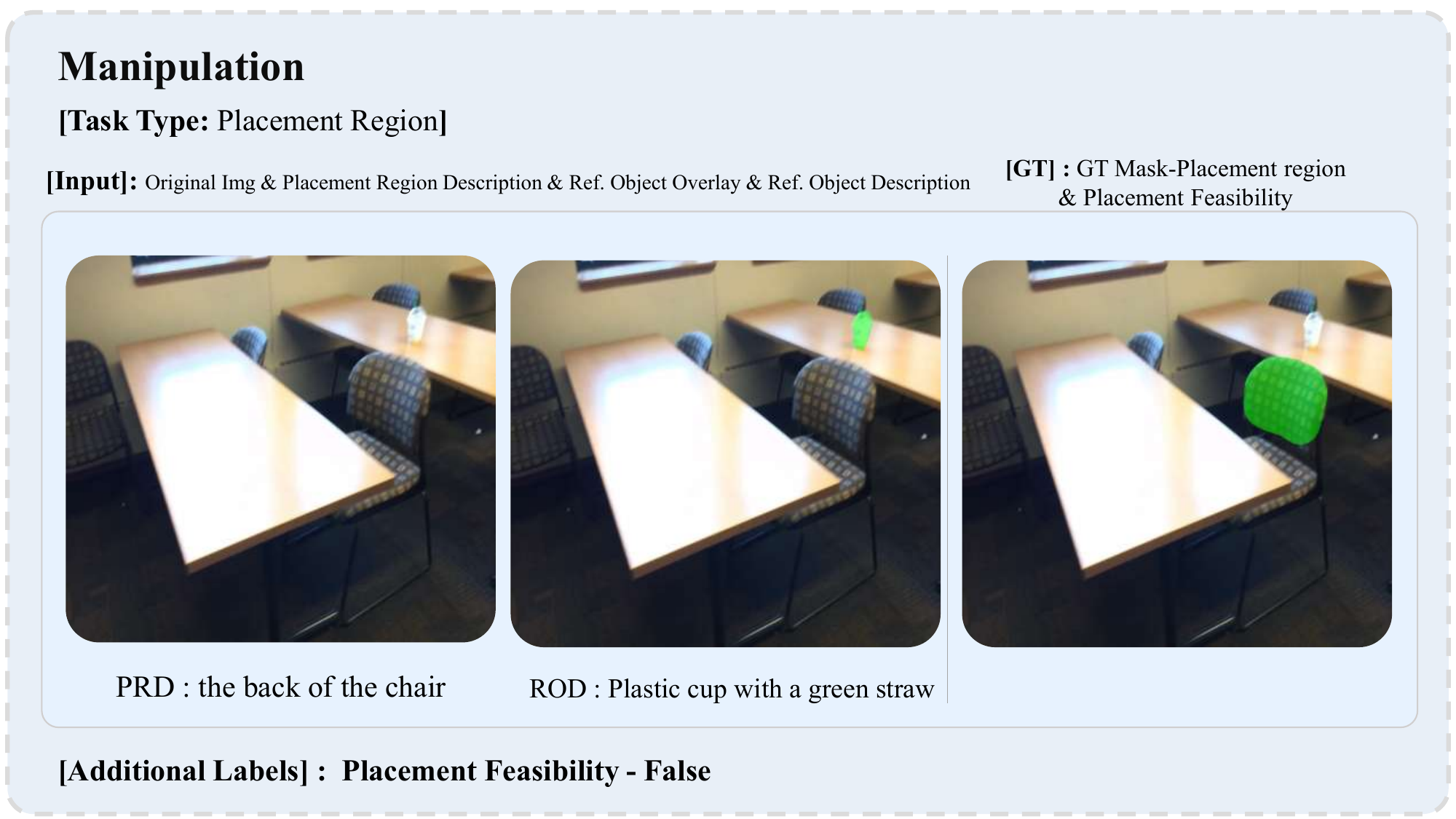}

\includegraphics[width=0.495\textwidth]{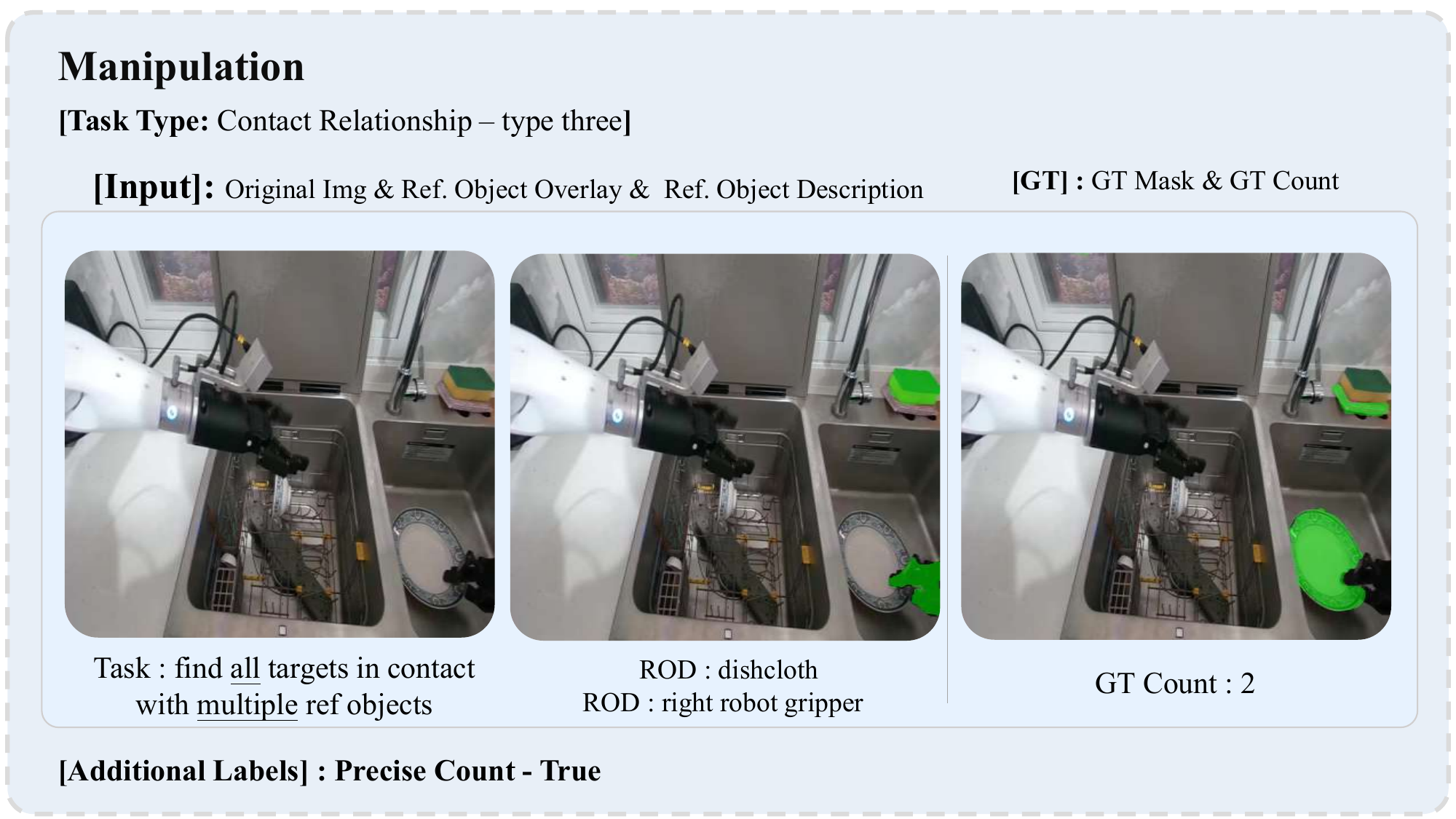}
\includegraphics[width=0.495\textwidth]{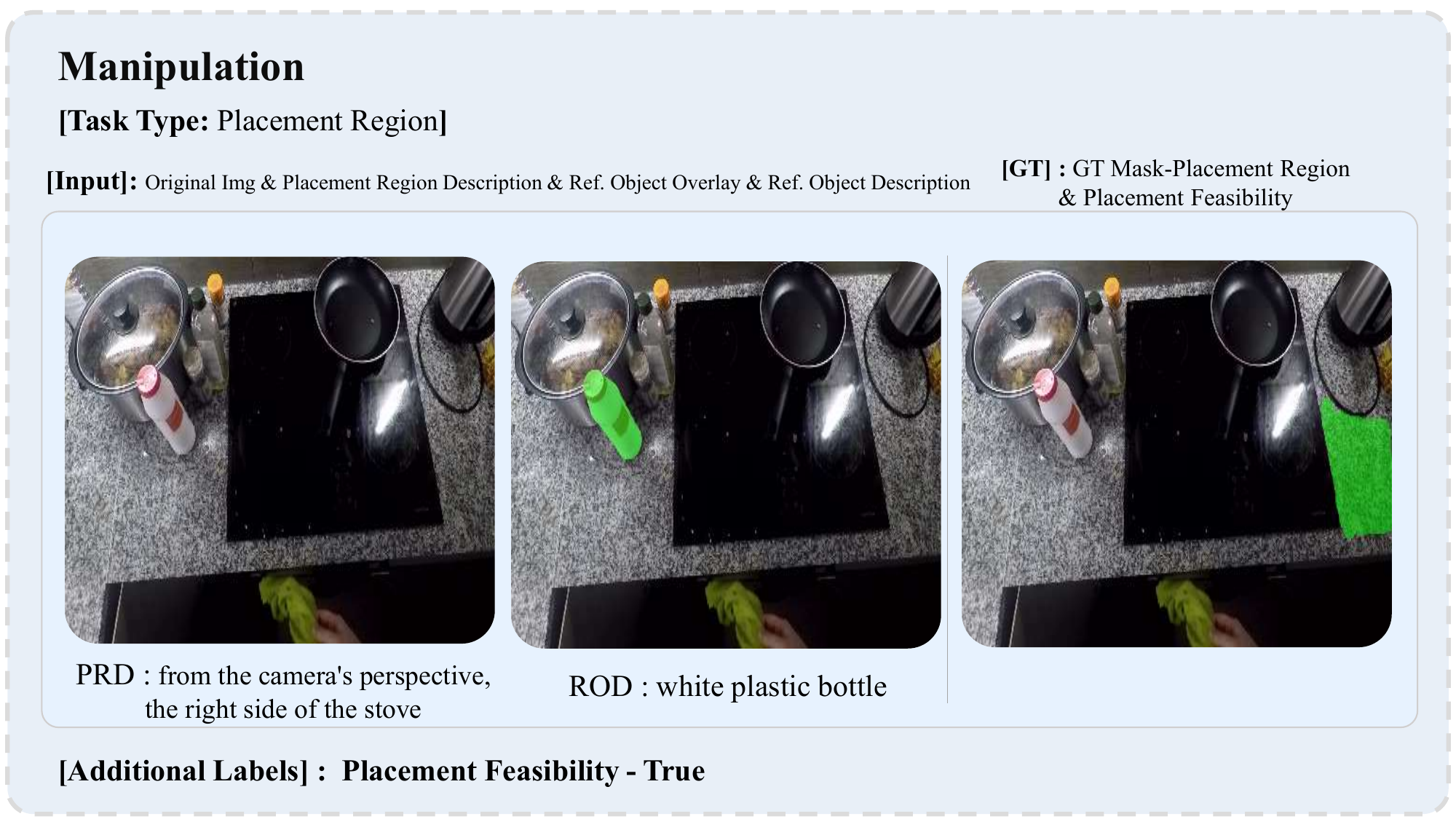}

\caption{Examples of Manipulation - Contact Relationship and Placement Region tasks}
\label{fig:example_MAN_CR_PR}
\end{figure*}

\paragraph{Manipulation} Figures \ref{fig:example_MAN_AR} and \ref{fig:example_MAN_CR_PR} present examples from the Manipulation category, covering the tasks of Affordance Region, Contact Relationship, and Placement Region.


\subsection{Data Sources}
\label{app:epic_data_sources}
We provide a detailed summary of the source datasets used in EPIC-Bench. As shown in Table~\ref{tab:data_source}, we curate images from 21 publicly available datasets spanning four domains: general scenes, indoor environments, robotics, and egocentric views, thereby ensuring the diversity of our dataset.

We obtain all existing datasets from their original sources released by the authors. We refer readers to these sources for the dataset licenses.

\begin{table*}[th!]
\centering
\begin{tabular}{l|p{13cm}}
\hline
\textbf{Task} & \multicolumn{1}{c}{\textbf{Benchmark Dataset Source}} \\
\hline
General & Visual Genome \cite{Visual_genome}, GQA \cite{GQA}, ADE20K \cite{ade20k}, MSD \cite{MSD} \\
& RefCOCO \cite{refcoco}, OmniSpatial\cite{jia2026omnispatial},  LVIS \cite{LVIS}, CLEVR \cite{clevr}\\
\hline
Indoor & OpenSurfaces \cite{os_v3}, ARKitScenes \cite{arkitscenes}, ScanNet \cite{scanNet}, MP3D\cite{MP3D}, OCID \cite{OCID} \\
& 3RScan \cite{3RScan} \\
\hline
Robotic & Agibot-World \cite{agibotworld}, RoboVQA \cite{robovqa2023arxiv}, BridgeData V2 \cite{BridgeData}, RoboNet \cite{robonet} \\
\hline
Egocentric & EPIC-KITCHENS \cite{epic-kitchens}, RoboVQA \cite{robovqa2023arxiv}\\
& HOVA-500K \cite{HOVA-500K}, EGTEA Gaze+ \cite{EGTEA} \\
\hline
\end{tabular}
\caption{Dataset source summary in EPIC-Bench.}
\label{tab:data_source}
\end{table*}


\subsection{Quality Control Details}

\label{app:quality_control_details}

\paragraph{Annotation and Cross-Checking.}
We employ 20 human annotators with undergraduate-level education, contributing over 4,800 person-hours across 30 working days. Annotators are initially assigned to different task categories, and all annotations undergo at least two rounds of cross-checking. The revision rates after the first and second rounds are 15\% and 6\%, respectively.

\paragraph{Ambiguity and Shortcut Filtering.}
We explicitly identify and revise ambiguous or shortcut-prone samples. For example, as shown in Fig.~\ref{fig:data_collection_pipeline}, the description \textit{the first shelf of the bookcase} is ambiguous because \textit{first} may refer to either the top or bottom shelf; we therefore revise it to \textit{the top shelf} according to the annotated target. Another sample with the description \textit{the person on the ice} is considered overly simple because both people in the image are on the ice, allowing the target to be identified from the category cue alone without understanding the spatial relation \textit{on the ice}. Such samples are revised or removed. We also avoid ambiguous expressions such as \textit{the bigger objects} when multiple same-category objects differ in size, and instead specify the target count, e.g., \textit{the two larger objects}.

\paragraph{Target-Count Verification.}
For samples with target-count labels, a second annotator independently verifies the number of targets using only the image and text. In cases of disagreement, we revise the count, set \texttt{precise\_count} to false, or remove the sample.

\paragraph{Subjective Annotation Validation.}
For tasks involving subjective spatial or interaction judgments, the ground truth is defined according to the region that a human would most likely traverse or interact with when executing the instruction. For example, for \textit{pick up the cup}, the annotated affordance region typically covers the handle or central body rather than the rim, although grasping the rim may occasionally be feasible. These annotations are further validated through cross-checking to improve consistency.

%% file: sections/Appendix/Section_D.tex
\section{Evaluation Details}
\label{app:evaluation_details}

\subsection{Prompt Templates and Prompt Robustness}
\label{app:prompts}

\subsubsection{Prompt Details}
\begin{figure*}[tb]
  \centering
  \includegraphics[width=\textwidth]{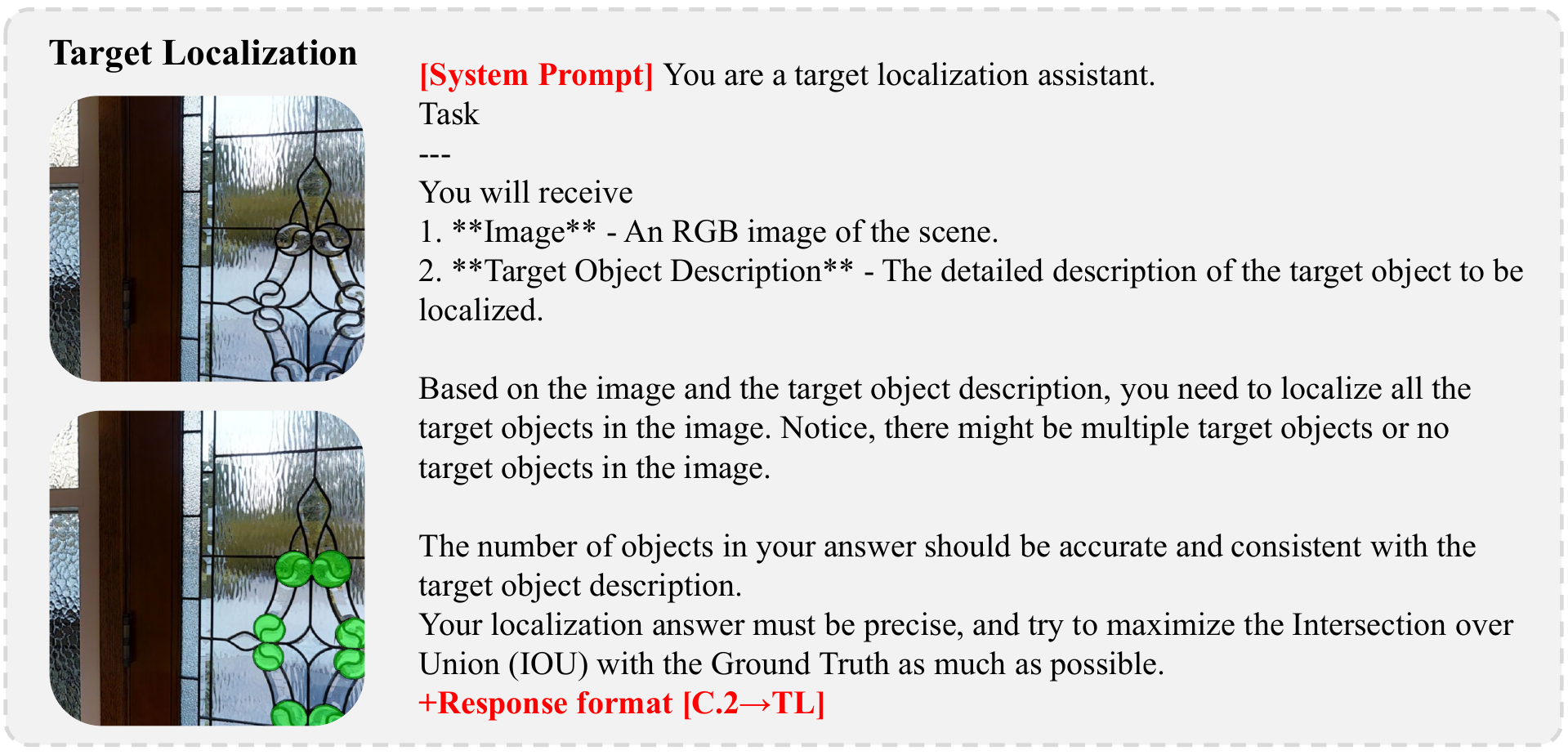}
    \caption{\textbf{Prompt for Target Localization.} Target Localization requires the model to identify all target objects matching the textual description.}
  \label{fig:TL}
\end{figure*}

\begin{figure*}[tb]
  \centering
  \includegraphics[width=0.9\textwidth]{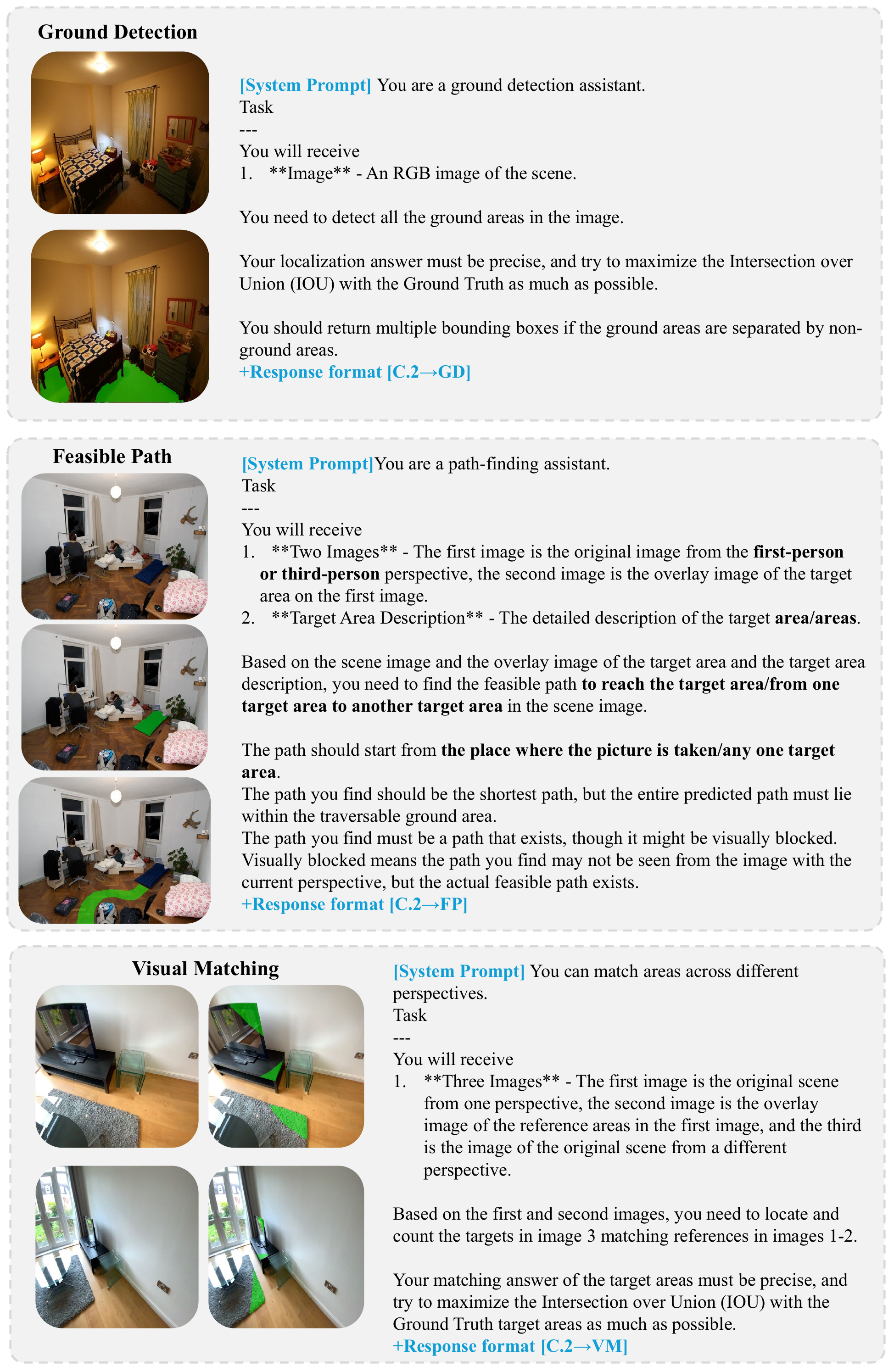}
    \caption{\textbf{Prompts for Navigation.} Navigation category includes three sub-tasks: Ground Detection (identifying traversable ground regions), Feasible Path (planning routes to target objects in egocentric view or between two targets in exocentric view), and Visual Matching (localizing the same reference area across different viewpoints).}
  \label{fig:nav}
\end{figure*}

\begin{figure*}[tb]
  \centering
  \includegraphics[width=0.8\textwidth]{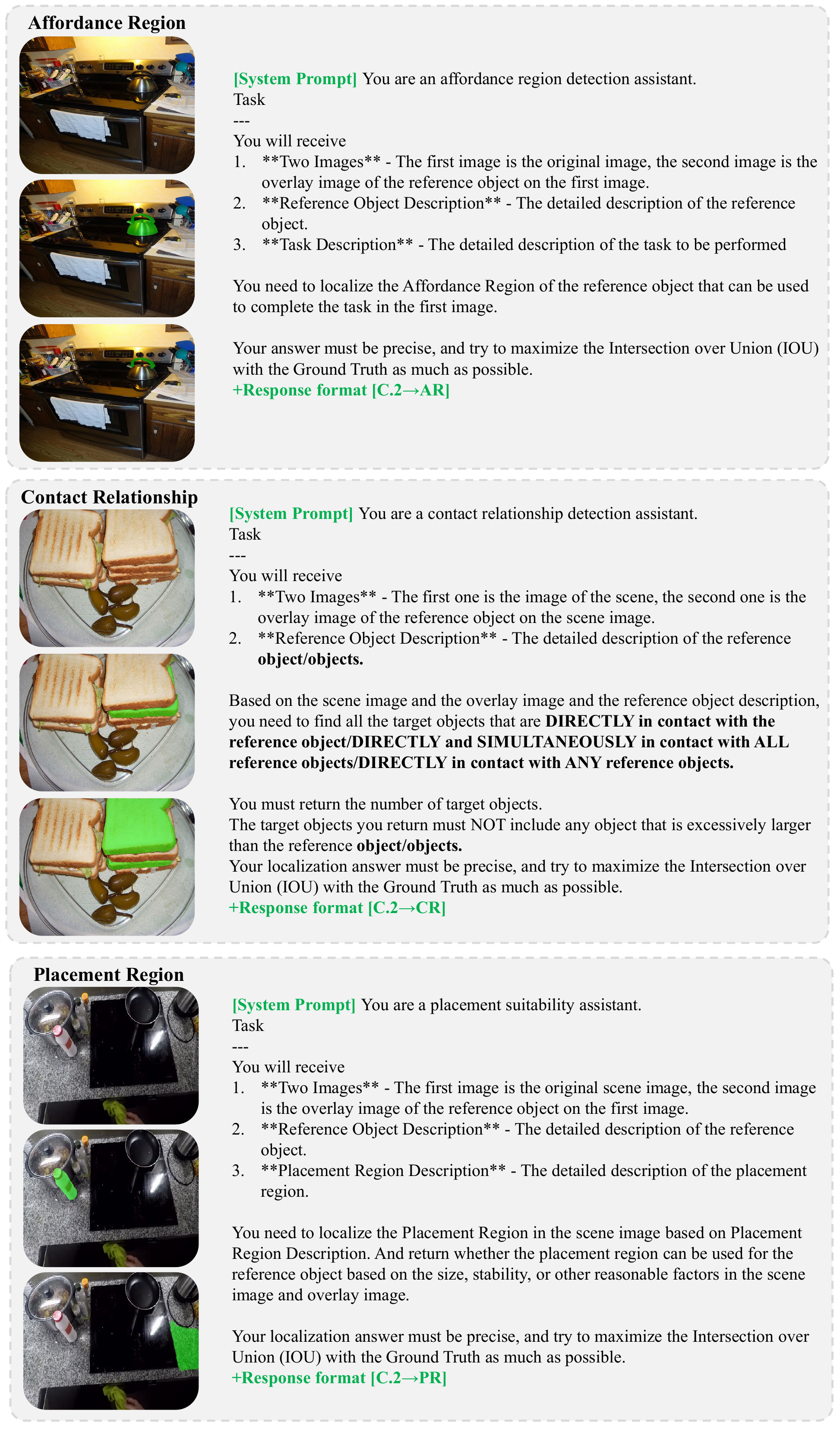}
    \caption{\textbf{Prompts for Manipulation.} Affordance Region (localizing operable regions of a reference object), Contact Relationship (identifying objects in contact with single/multiple reference objects under three contact conditions), and Placement Region (localizing placement areas and determining placement feasibility).}
  \label{fig:man}
\end{figure*}

\paragraph{Task Prompts.}
We provide task-specific prompt templates for Target Localization (Figure~\ref{fig:TL}), Navigation, including Ground Detection, Feasible Path, and Visual Matching (Figure~\ref{fig:nav}), and Manipulation, including Affordance Region, Contact Relationship, and Placement Region (Figure~\ref{fig:man}).

\begin{figure*}[htbp]
  \centering
  \includegraphics[width=\textwidth]{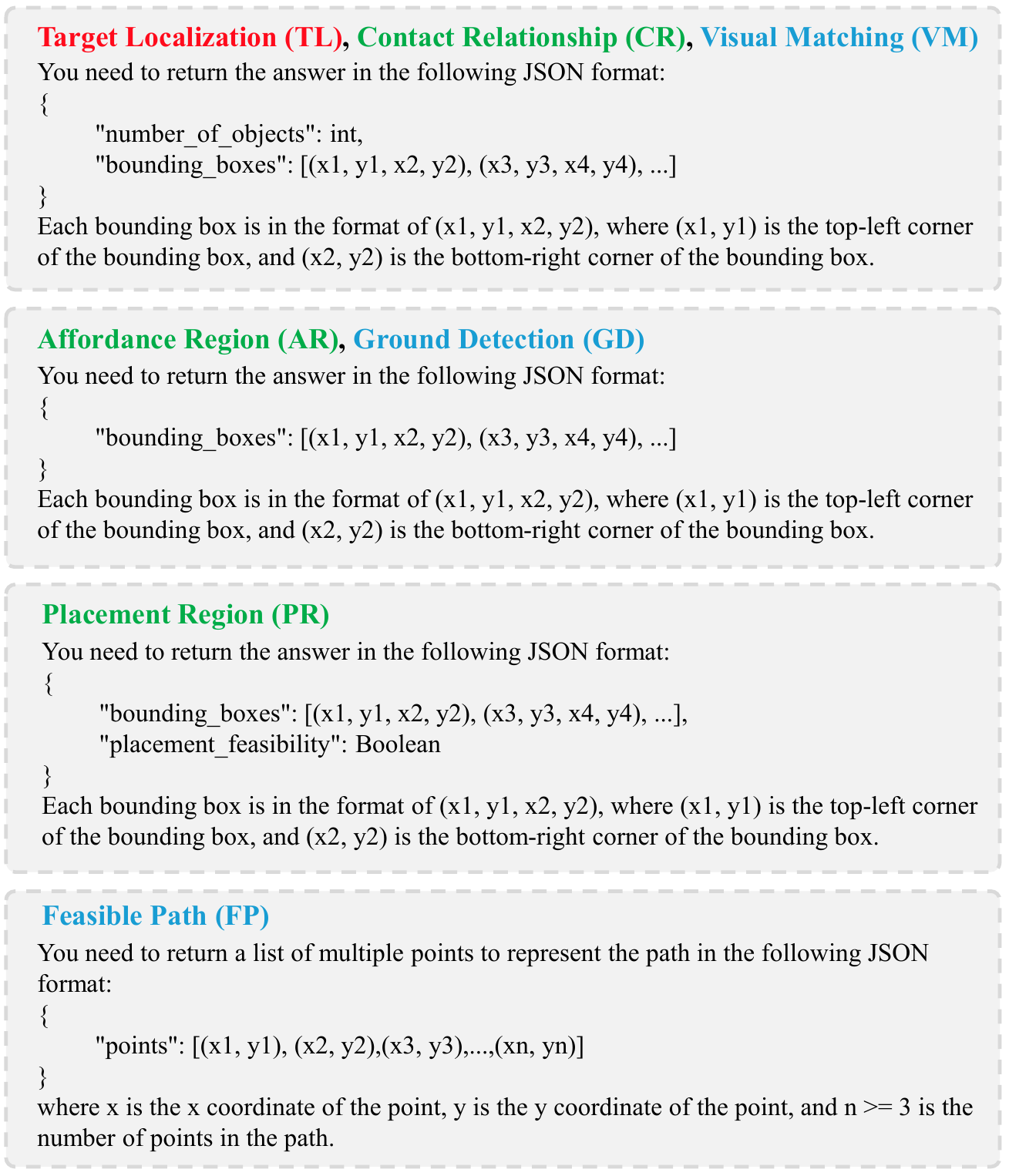}
\caption{\textbf{Response format for each sub-task.} For \textbf{Target Localization (TL)}, \textbf{Contact Relationship (CR)}, and \textbf{Visual Matching (VM)}, models are required to output bounding boxes and the count of target objects. For \textbf{Affordance Region (AR)} and \textbf{Ground Detection (GD)}, models are required to output the corresponding bounding boxes. For \textbf{Placement Region (PR)}, models are required to output the placement region bounding box along with a binary judgment on placement feasibility. For \textbf{Feasible Path (FP)}, models are required to output a sequence of points representing the path.}
\label{fig:response_format}
\end{figure*}

\paragraph{Response Format.}
As illustrated in Figure~\ref{fig:response_format}, we require models to return predictions in a structured JSON format for consistent evaluation. The grounding representation follows the format natively supported by each model: conventional VLMs typically output bounding-box coordinates, while models with pixel-level grounding capabilities may directly produce masks. We also implement a robust answer-extraction procedure that tolerates minor formatting errors, such as mismatched brackets or extra spaces. Our parser further handles different coordinate conventions, including absolute coordinates, coordinates normalized to $[0,1]$, and coordinates scaled to $[0,1000]$.

\subsubsection{Prompt Robustness}

We further evaluate the robustness of EPIC-Bench evaluation to prompt variations. We randomly sample 40 examples from each of the 23 subtasks, resulting in 920 examples, and evaluate 11 representative models under three prompting strategies: (1) a base prompt with the task description, (2) a simple CoT prompt additionally instructing the model to ``think step by step'', and (3) a complex CoT prompt with manually designed reasoning instructions.

\begin{table}[t]
\centering
\small
\resizebox{\columnwidth}{!}{
\begin{tabular}{lccc}
\toprule
\textbf{Model} & \textbf{Base} & \textbf{Simple CoT} & \textbf{Complex CoT} \\
\midrule
GPT-5.5                     & 54.10 & 53.67 & 54.11 \\
Gemini-3-Pro-Preview        & 53.44 & 56.12 & 56.16 \\
Doubao-Seed-1.8             & 50.38 & 50.54 & 50.49 \\
Qwen3-VL-30B-A3B-Thinking   & 45.49 & 45.63 & 46.34 \\
GLM-4.6V                    & 41.20 & 40.97 & 41.61 \\
InternVL3.5-241B-A28B       & 41.03 & 41.59 & 41.77 \\
RoboBrain2.0-32B            & 39.78 & 38.46 & 39.81 \\
Pelican1.0-VL-72B           & 35.75 & 24.41 & 34.71 \\
Gemma-3-27B-IT              & 28.52 & 28.23 & 29.63 \\
Cosmos-Reason1-7B           & 22.97 & 22.18 & 21.31 \\
LLaVA-NeXT-72B              & 22.55 & 22.24 & 21.95 \\
\midrule
\textbf{Average}            & \textbf{39.56} & \textbf{38.55} & \textbf{39.81} \\
\bottomrule
\end{tabular}
}
\caption{Performance under different prompting strategies.}
\label{tab:prompt_robustness}
\end{table}

As shown in Table~\ref{tab:prompt_robustness}, the average scores vary only slightly across the three prompting strategies, indicating that the evaluation results are generally robust to these prompt variations. EPIC-Bench does not require a fixed prompt template; users are encouraged to adapt the prompt format as needed, provided that the task semantics remain consistent with the benchmark definition.


\subsection{Post-Processing Strategies}
\label{app:post_process}

\paragraph{Main Evaluation.}
EPIC-Bench provides ground-truth annotations in mask format. Our main evaluation uses the grounding format natively supported by each model, as determined by its model design and training setup. For models that natively produce irregular mask outputs, we directly compute IoU between the predicted and ground-truth masks. For models whose grounding outputs are represented as bounding boxes, we convert each predicted box region into an equivalent mask before computing IoU, keeping the evaluation aligned with the spatial extent explicitly predicted by the model.

\paragraph{SAM3-based Post-Processing.}
To evaluate the effect of different post-processing strategies, we additionally consider a SAM3-based refinement of the original bounding-box predictions. Specifically, we use SAM3 to extract the mask of the most salient object or region within each predicted bounding box. We compare the overall score, per-sample win/loss/tie statistics, and IoU before and after post-processing. A sample is counted as a win or loss when the absolute score change exceeds 0.01; otherwise, it is counted as a tie.

\begin{table*}[tb]
\centering
\small
\resizebox{\textwidth}{!}{
\begin{tabular}{lccccccccc}
\toprule
\textbf{Model} &
\textbf{Score (BBox)} &
\textbf{Score (SAM3)} &
$\boldsymbol{\Delta}$\textbf{Score} &
\textbf{Win (\%)} &
\textbf{Loss (\%)} &
\textbf{Tie (\%)} &
\textbf{IoU (BBox)} &
\textbf{IoU (SAM3)} &
$\boldsymbol{\Delta}$\textbf{IoU} \\
\midrule
Gemini-3-Pro                     & 53.85 & 63.28 &  9.42 & 61.79 & 15.55 & 22.65 & 0.404 & 0.569 &  0.165 \\
Qwen3-VL-235B-A22B-Thinking     & 50.23 & 57.54 &  7.32 & 56.63 & 23.71 & 19.67 & 0.391 & 0.513 &  0.123 \\
GPT-5.5                          & 51.72 & 57.78 &  6.06 & 53.18 & 25.22 & 21.60 & 0.375 & 0.481 &  0.106 \\
GLM-4.6V                         & 42.07 & 46.69 &  4.61 & 42.34 & 24.67 & 33.00 & 0.285 & 0.360 &  0.075 \\
Doubao-Seed-1.8                  & 46.53 & 51.11 &  4.58 & 40.52 & 25.49 & 33.99 & 0.301 & 0.374 &  0.073 \\
InternVL3.5-241B-A28B            & 40.30 & 44.34 &  4.03 & 39.68 & 23.45 & 36.87 & 0.307 & 0.379 &  0.073 \\
RoboBrain2-32B                   & 39.40 & 40.90 &  1.49 & 35.52 & 32.66 & 31.82 & 0.307 & 0.336 &  0.028 \\
Cosmos-Reason1-7B                & 26.26 & 26.25 & -0.01 & 23.33 & 34.05 & 42.62 & 0.167 & 0.167 &  0.000 \\
LLaVA-NeXT-72B                   & 24.42 & 23.71 & -0.71 & 23.68 & 40.05 & 36.27 & 0.148 & 0.137 & -0.011 \\
Gemma-3-27B-IT                   & 26.68 & 25.08 & -1.61 &  6.64 & 36.27 & 57.09 & 0.058 & 0.030 & -0.028 \\
Pelican1.0-VL-72B                & 36.10 & 32.64 & -3.46 & 10.04 & 50.46 & 39.50 & 0.221 & 0.161 & -0.060 \\
\midrule
\textbf{Average}                 & \textbf{39.78} & \textbf{42.66} & \textbf{2.88} &
\textbf{35.76} & \textbf{30.14} & \textbf{34.10} &
\textbf{0.269} & \textbf{0.319} & \textbf{0.049} \\
\bottomrule
\end{tabular}
}
\caption{\textbf{Effect of SAM3-based post-processing on grounding evaluation.} BBox denotes evaluation using the original predicted bounding-box regions, while SAM3 denotes evaluation after refining these regions into salient-object masks using SAM3. The comparison shows that external segmentation-based refinement changes absolute scores for some models but largely preserves the relative performance trend across models.}
\label{tab:sam3_postprocess}
\end{table*}

\paragraph{Analysis.}
As shown in Table~\ref{tab:sam3_postprocess}, SAM3-based post-processing yields an average score increase of 2.88 and an average IoU increase of 0.049, while the average proportions of improved and degraded samples are comparable. However, the effect varies substantially across models. The improvement is strongly correlated with the original bbox-based performance: Spearman $\rho(\Delta\mathrm{Score}, \mathrm{Score}_{\mathrm{BBox}})=0.9000$ and $\rho(\Delta\mathrm{IoU}, \mathrm{IoU}_{\mathrm{BBox}})=0.8545$, with $p<0.001$ in both cases.

Stronger models generally benefit more from SAM3-based refinement, suggesting that their original predictions often already identify the correct target region and can therefore be further improved through finer boundary extraction. In contrast, weaker models frequently produce inaccurate or misaligned bounding boxes, limiting the benefit of subsequent segmentation. Although post-processing changes the absolute scores and enlarges the performance differences between some models, the overall model ranking remains largely unchanged.

These results suggest that our main evaluation protocol provides a stable basis for comparing the relative VLG capabilities of existing models while keeping the evaluation independent of external segmentation capabilities. Meanwhile, the mask-level annotations in EPIC-Bench enable direct fine-grained evaluation of models with native pixel-level outputs. We encourage further research on VLMs with native mask- or pixel-level grounding capabilities to make fuller use of such fine-grained supervision and evaluation.


\subsection{Path Score Definition}
\label{app:path_score}

The Path Score evaluates predicted navigation paths using five components: \textbf{Start--End Score}, \textbf{Traversable Path Ratio Score}, \textbf{Away-from-Start Score}, \textbf{Approaching-Goal Score}, and \textbf{Continuity Score}.

Given a sequence of predicted points, we first remove spatially redundant points using neighbor filtering. Starting from the first point, a subsequent point is retained only if its distance from the most recently retained point is at least $1/30$ of the image diagonal. The remaining points are then evaluated using the five components below.

\paragraph{Start--End Score.}
The Start--End Score measures whether the predicted path begins near the start region and terminates near the goal region. The definitions of the start and goal regions depend on the view setting. In the egocentric setting, they are determined by the predefined task geometry. In the exocentric setting, the two annotated endpoint regions are assigned adaptively: the region closer to the first predicted point is treated as the start region $S$, and the other as the goal region $G$.

For a point $p$, we define the normalized distance to a target region $R$ as
\begin{equation}
D_{\mathrm{norm}}(p,R)
=
\frac{d(p,R)}{D_{\mathrm{img}}},
\end{equation}
where $d(p,R)$ is the shortest distance from $p$ to region $R$, and $D_{\mathrm{img}}$ is the image diagonal length. The corresponding score is
\begin{equation}
E(p,R)
=
\max\left(0,1-kD_{\mathrm{norm}}(p,R)\right),
\end{equation}
where $k=3.0$. The first and last predicted points are evaluated against $S$ and $G$, respectively.

\paragraph{Traversable Path Ratio Score.}
For each segment $l_i$ connecting consecutive points $p_{i-1}$ and $p_i$, the Traversable Path Ratio Score measures the proportion of the segment lying within the ground-truth feasible-path region $M_{\mathrm{fp}}$:
\begin{equation}
T_i
=
\frac{\operatorname{Length}(l_i \cap M_{\mathrm{fp}})}
{\operatorname{Length}(l_i)}.
\end{equation}
In implementation, this ratio is approximated by uniformly sampling points along each segment and measuring the fraction that falls inside the feasible-path mask.

\paragraph{Away-from-Start and Approaching-Goal Scores.}
These two components measure directional progress along the predicted trajectory. For each point $p_i$ ($i\geq2$), we compare its distances to the start and goal regions with those of the preceding point:
\begin{equation}
A_i =
\begin{cases}
1, & d(p_i,S) > d(p_{i-1},S), \\
0, & \text{otherwise},
\end{cases}
\end{equation}
and
\begin{equation}
B_i =
\begin{cases}
1, & d(p_i,G) < d(p_{i-1},G), \\
0, & \text{otherwise}.
\end{cases}
\end{equation}

\paragraph{Continuity Score.}
The Continuity Score penalizes excessively large jumps between consecutive path points. Let $D_{SG}$ denote the distance between the start and goal regions. For each segment, we define
\begin{equation}
C_i =
\begin{cases}
1, & d(p_i,p_{i-1}) \leq \frac{2}{3}D_{SG}, \\
0, & \text{otherwise}.
\end{cases}
\end{equation}
Together with the neighbor-filtering step, this encourages the predicted trajectory to maintain a reasonable spatial progression.

After filtering, a valid path must contain at least three points. Each point is assigned a weighted combination of the applicable components according to its position in the sequence, as shown in Table~\ref{tab:path_score_weight}. The final Path Score is the average of the resulting point-wise scores.

\begin{table}[t]
\centering
\small
\resizebox{\columnwidth}{!}{
\begin{tabular}{lccccc}
\toprule
\textbf{Point Type} &
\textbf{Start--End} &
\makecell{\textbf{Path} \\ \textbf{Ratio}} &
\makecell{\textbf{Away} \\ \textbf{Start}} &
\makecell{\textbf{Approach} \\ \textbf{Goal}} &
\textbf{Continuity} \\
\midrule
Start        & 1.0 & 0   & 0   & 0   & 0   \\
End          & 0.3 & 0.4 & 0.1 & 0.1 & 0.1 \\
Intermediate & 0   & 0.4 & 0.2 & 0.2 & 0.2 \\
\bottomrule
\end{tabular}
}
\caption{Weights of the Path Score components for different point types.}
\label{tab:path_score_weight}
\end{table}

%% file: sections/Appendix/Section_E.tex
\section{Full Evaluation Results}
\label{app:full_results}

Table~\ref{tab:table-performance-full} reports the complete EPIC-Bench results for over 80 VLM variants. We group the evaluated models into four categories.

\paragraph{Proprietary Models.}
This category includes representative closed-source systems from Gemini, Claude, GPT, Doubao-Seed, and HunYuan.

\paragraph{Open-source Models.}
This category covers widely used open-source VLM families, including Qwen-VL~\cite{qwen3vl}, InternVL~\cite{internvl3,internvl35}, Gemma 3~\cite{gemma3}, MiMo-VL~\cite{MiMo-VL}, GLM-4.6V~\cite{hong2025glm}, Step3-VL~\cite{step3vl}, and LLaVA-Next~\cite{llavanext}.

\paragraph{Segmentation-capable Models.}
This category includes VLMs such as GLaMM~\cite{rasheed2024glamm}, LISA++~\cite{yang2023lisa++}, and UFO~\cite{tang2026ufo} that natively support pixel-level or mask-form grounding outputs.

\paragraph{Embodied Foundation Models.}
This category includes models specifically developed for embodied perception or reasoning, such as RoboBrain~\cite{RoboBrain2.5}, MiMo-Embodied~\cite{MiMo-Embodied}, Pelican-VL~\cite{Pelican-VL}, RynnBrain~\cite{RynnBrain}, VeBrain~\cite{VeBrain}, and Cosmos-Reason~\cite{Cosmos-Reason1}.

Overall, the evaluation spans diverse model families, parameter scales, training paradigms, and grounding output formats, providing a comprehensive comparison of current VLMs on embodied vision-language grounding.

\input{sections/appendix_tab}

%% file: sections/appendix_tab.tex
\setlength{\overfullrule}{0pt}

\onecolumn
\setlength{\tabcolsep}{4.5pt}
\begin{longtable}{lc ccc ccc ccc}
\captionsetup{width=\textwidth}
\toprule
    \multirow{2}{*}{\textbf{Model}}
    & \multirow{2}{*}{\textbf{Average}}
    & \multicolumn{3}{c}{\textbf{Target Local.}}
    & \multicolumn{3}{c}{\textbf{Navigation}}
    & \multicolumn{3}{c}{\textbf{Manipulation}} \\
\cmidrule(r){3-5} \cmidrule(r){6-8} \cmidrule(r){9-11}
&  & BA & SRA & ECA & GD & FP & VM & AR & CR & PR \\
\midrule
\endfirsthead 
\toprule
\multirow{2}{*}{\textbf{Model}}
    & \multirow{2}{*}{\textbf{Average}}
    & \multicolumn{3}{c}{\textbf{Target Local.}}
    & \multicolumn{3}{c}{\textbf{Navigation}}
    & \multicolumn{3}{c}{\textbf{Manipulation}} \\
\cmidrule(r){3-5} \cmidrule(r){6-8} \cmidrule(r){9-11}
&  & BA & SRA & ECA & GD & FP & VM & AR & CR & PR \\
\midrule
\endhead
\bottomrule
\multicolumn{11}{r}{\footnotesize Continued on next page} \\ 
\endfoot
\endlastfoot

\rowcolor{blue!8}
\multicolumn{11}{c}{\textbf{Proprietary Models}} \\
Gemini-2.5-Flash-Lite & 34.77 & 39.06 & 39.53 & 31.51 & 35.97 & 40.21 & 33.02 & 2.58 & 30.17 & 32.38 \\
Gemini-2.5-Flash & 36.33 & 38.46 & 39.16 & 34.05 & 27.28 & 51.98 & 40.79 & 2.88 & 37.15 & 36.28 \\
Gemini-2.5-Pro$^{\dagger}$ & 37.63 & 42.86 & 41.62 & 36.43 & 9.63 & 46.29 & 42.70 & 8.72 & 33.94 & 42.78 \\
Gemini-3-Flash-Pre & 37.04 & 46.40 & 31.48 & 32.29 & 48.98 & 22.89 & 47.25 & 11.91 & 42.29 & 48.09 \\
Gemini-3-Flash-Pre$^{\dagger}$ & 44.69 & 50.26 & 41.70 & 37.79 & \textbf{51.11} & 52.63 & 49.92 & 22.62 & 46.51 & 50.89 \\
Gemini-3-pro$^{\dagger}$ & \cellcolor{green!10}\textbf{54.81} & \underline{56.81} & 54.50 & \cellcolor{green!10}\textbf{53.90} & 48.82 & \cellcolor{green!10}\textbf{70.82} & \cellcolor{green!10}\textbf{60.08} & \cellcolor{green!10}\textbf{30.40} & \underline{51.78} & \underline{53.42} \\
Gemini-3.1-Pro$^{\dagger}$ & \underline{54.72} & 55.38 & \cellcolor{green!10}\textbf{56.65} & \underline{53.04} & 47.22 & \underline{70.32} & \underline{59.06} & \underline{28.89} & \cellcolor{green!10}\textbf{53.70} & \cellcolor{green!10}\textbf{55.25} \\
Claude-Sonnet-4.6$^{\dagger}$ & 43.24 & 45.43 & 45.49 & 36.47 & 40.16 & 68.46 & 42.77 & 9.79 & 46.21 & 43.81 \\
Claude-Sonnet-4.6 & 31.34 & 43.84 & 32.16 & 21.94 & 5.25 & 39.56 & 40.90 & 0.54 & 35.90 & 35.77 \\
Claude-Haiku-4.5$^{\dagger}$ & 22.03 & 25.55 & 24.15 & 17.59 & 3.92 & 33.82 & 36.34 & 0.40 & 9.81 & 30.37 \\
Claude-Haiku-4.5 & 30.94 & 31.96 & 34.50 & 28.78 & 24.76 & 40.91 & 35.88 & 1.89 & 29.88 & 33.18 \\
Claude-Opus-4 & 25.71 & 26.60 & 31.22 & 22.47 & 12.44 & 27.30 & 32.57 & 1.42 & 27.27 & 34.81 \\
o3$^{\dagger}$ & 36.05 & 39.37 & 40.96 & 32.86 & 36.24 & 30.64 & 40.73 & 5.49 & 32.43 & 43.64 \\
o4-mini$^{\dagger}$ & 32.83 & 36.75 & 38.02 & 32.73 & 18.27 & 20.49 & 45.63 & 3.54 & 33.58 & 38.75 \\
GPT-4o & 29.52 & 27.42 & 35.99 & 26.69 & 26.95 & 32.45 & 36.36 & 2.18 & 33.22 & 34.50 \\
GPT-5-mini$^{\dagger}$ & 35.21 & 38.70 & 39.43 & 30.24 & 24.36 & 38.53 & 45.97 & 3.86 & 41.32 & 38.78 \\
GPT-5.1 & 34.81 & 35.31 & 39.66 & 32.05 & 36.62 & 38.62 & 44.69 & 4.34 & 26.44 & 38.63 \\
GPT-5.2 & 36.82 & 38.26 & 41.53 & 32.02 & 33.06 & 48.72 & 44.10 & 5.01 & 35.25 & 37.26 \\
GPT-5.4 & 38.95 & 43.78 & 43.90 & 33.84 & 32.24 & 43.12 & 44.35 & 8.67 & 39.55 & 37.30 \\
GPT-5.5 & 50.16 & \textbf{57.78} & \underline{55.37} & 45.19 & \underline{50.97} & 38.57 & 51.65 & 24.62 & 49.67 & 48.70 \\
Doubao-Seed-1.8$^{\dagger}$ & 46.32 & 55.29 & 49.68 & 46.10 & 46.02 & 53.43 & 45.90 & 10.94 & 14.36 & 40.79 \\
Doubao-Seed-1.6-Vision$^{\dagger}$ & 46.80 & 54.64 & 49.92 & 42.27 & 41.64 & 58.74 & 42.66 & 9.52 & 47.48 & 41.62 \\
HunYuan-T1-Vision$^{\dagger}$ & 43.00 & 51.24 & 47.60 & 39.67 & 42.63 & 50.08 & 33.54 & 6.63 & 38.95 & 34.49 \\
HunYuan-Vision-1.5 & 32.55 & 38.37 & 39.60 & 29.67 & 31.72 & 17.07 & 35.71 & 6.38 & 32.88 & 29.25 \\
HunYuan-Turbos-Vision & 24.61 & 24.72 & 33.33 & 25.58 & 4.49 & 17.88 & 36.14 & 0.67 & 22.02 & 26.69 \\

\midrule
\rowcolor{blue!8}[0pt][0pt]
\multicolumn{11}{@{}c@{}}{\textbf{Open-source Models}} \\
Qwen3.6-35B-A3B & 38.52 & 46.40 & 46.19 & 40.01 & 3.90 & 39.16 & 39.42 & 5.41 & 34.19 & 34.52 \\
Qwen3.6-27B & 45.38 & 51.04 & 48.02 & 41.46 & 49.99 & 59.15 & 43.45 & 10.61 & 38.69 & 37.35 \\
Qwen3.5-397B-A17B$^{\dagger}$ & \underline{47.47} & \underline{55.45} & 51.82 & 42.10 & 51.10 & 60.11 & 45.35 & 12.40 & 26.33 & \underline{42.51} \\
Qwen3.5-397B-A17B & 45.16 & 50.56 & 50.06 & \underline{43.45} & \textbf{51.94} & 50.61 & 41.46 & 8.31 & 34.51 & 37.08 \\
Qwen3.5-122B-A10B$^{\dagger}$ & 47.24 & 53.61 & 49.13 & 41.16 & \underline{51.42} & \textbf{63.47} & 45.19 & 12.31 & 46.04 & 39.77 \\
Qwen3.5-122B-A10B & 44.54 & 50.77 & 49.28 & 42.31 & 49.65 & 50.50 & 42.62 & 8.30 & 36.91 & 32.16 \\
Qwen3.5-35B-A3B$^{\dagger}$ & 45.58 & 53.51 & 49.18 & 39.03 & 43.09 & \underline{61.17} & 41.18 & 10.75 & 41.70 & 38.05 \\
Qwen3.5-35B-A3B & 41.32 & 46.76 & 45.82 & 38.19 & 48.00 & 45.47 & 37.01 & 10.01 & 36.94 & 32.36 \\
Qwen3.5-27B$^{\dagger}$ & 45.67 & 54.15 & 50.03 & 40.61 & 46.96 & 39.43 & \underline{46.77} & 13.15 & \textbf{49.56} & 39.84 \\
Qwen3.5-27B & 44.71 & 49.10 & 48.55 & 42.23 & 49.74 & 53.50 & 44.10 & 11.93 & 37.22 & 36.42 \\
Qwen3.5-9B$^{\dagger}$ & 38.86 & 44.60 & 43.38 & 33.87 & 31.26 & 49.06 & 40.22 & 6.07 & 37.64 & 34.92 \\
Qwen3.5-9B & 35.45 & 40.49 & 42.04 & 31.90 & 38.07 & 36.81 & 33.93 & 6.09 & 26.70 & 28.25 \\
Qwen3-VL-235B-A22B$^{\dagger}$ & \textbf{50.93} & \cellcolor{green!10}\textbf{58.00} & \textbf{55.67} & \textbf{48.12} & 48.18 & 51.19 & \textbf{49.01} & \underline{16.96} & \underline{48.04} & \textbf{46.49} \\
Qwen3-VL-235B-A22B & 42.64 & 50.12 & 45.86 & 39.34 & 50.96 & 43.14 & 38.69 & 12.00 & 34.55 & 35.71 \\
Qwen3-VL-30B-A3B$^{\dagger}$ & 44.02 & 52.69 & 49.37 & 40.87 & 42.34 & 41.65 & 36.20 & 10.62 & 38.78 & 40.72 \\
Qwen3-VL-30B-A3B & 30.60 & 28.31 & 34.27 & 25.04 & 49.37 & 37.09 & 30.12 & 9.33 & 31.49 & 34.07 \\
Qwen3-VL-32B$^{\dagger}$ & 37.21 & 39.94 & 41.26 & 36.05 & 31.03 & 42.04 & 39.07 & 4.84 & 35.23 & 40.03 \\
Qwen3-VL-32B & 35.21 & 37.49 & 39.82 & 33.21 & 44.38 & 36.74 & 36.57 & 4.72 & 30.62 & 30.49 \\
Qwen3-VL-8B$^{\dagger}$ & 35.55 & 39.92 & 41.66 & 33.89 & 21.68 & 38.51 & 34.94 & 5.22 & 32.05 & 37.01 \\
Qwen3-VL-8B & 34.07 & 39.48 & 42.54 & 32.10 & 38.11 & 26.67 & 31.84 & 5.19 & 14.63 & 30.91 \\
Qwen3-VL-4B$^{\dagger}$ & 32.83 & 38.42 & 41.30 & 31.97 & 18.29 & 29.47 & 30.95 & 3.91 & 22.15 & 31.83 \\
Qwen3-VL-4B & 33.24 & 39.95 & 41.55 & 33.78 & 40.39 & 18.45 & 30.17 & 3.20 & 11.30 & 27.00 \\
Qwen3-VL-2B$^{\dagger}$ & 31.82 & 39.67 & 40.75 & 30.08 & 15.36 & 24.02 & 22.59 & 6.07 & 28.24 & 28.97 \\
Qwen3-VL-2B & 32.53 & 38.43 & 42.87 & 29.08 & 37.55 & 9.78 & 28.55 & 7.99 & 29.56 & 29.59 \\
Qwen2.5-VL-72B & 42.51 & 47.54 & 49.79 & 40.92 & 49.45 & 42.49 & 37.08 & 14.77 & 21.21 & 35.90 \\
InternVL3.5-241B-A28B & 40.75 & 44.65 & 50.59 & 37.61 & 35.94 & 35.47 & 33.73 & 16.85 & 36.41 & 37.60 \\
InternVL3.5-38B & 42.54 & 48.69 & \underline{52.60} & 39.95 & 43.21 & 35.01 & 28.07 & \textbf{17.22} & 34.05 & 35.83 \\
InternVL3.5-30B-A3B & 29.71 & 30.96 & 39.11 & 27.03 & 45.43 & 24.11 & 17.24 & 8.41 & 18.75 & 25.30 \\
InternVL3.5-14B & 38.92 & 43.95 & 49.96 & 37.13 & 46.11 & 31.82 & 24.57 & 11.23 & 25.98 & 28.66 \\
InternVL3.5-8B & 33.58 & 40.67 & 45.40 & 32.05 & 32.15 & 22.77 & 15.94 & 10.79 & 17.42 & 25.57 \\
InternVL3-78B & 36.04 & 40.68 & 44.56 & 32.78 & 35.98 & 30.55 & 34.93 & 6.90 & 29.07 & 31.69 \\
InternVL3-38B & 28.50 & 34.88 & 28.85 & 23.22 & 36.69 & 22.31 & 34.88 & 3.99 & 30.43 & 30.44 \\
InternVL3-14B & 30.17 & 31.98 & 38.44 & 25.41 & 32.29 & 29.53 & 27.64 & 3.25 & 27.49 & 31.19 \\
InternVL3-8B & 30.23 & 35.51 & 40.72 & 26.69 & 33.38 & 21.53 & 23.16 & 3.83 & 16.08 & 26.07 \\
Gemma-3-27B-IT & 27.17 & 26.93 & 33.66 & 26.68 & 11.62 & 32.17 & 26.33 & 1.65 & 31.26 & 32.18 \\
Gemma-3-12B-IT & 26.21 & 25.55 & 32.44 & 24.59 & 13.71 & 36.27 & 23.39 & 1.30 & 27.48 & 31.43 \\
Gemma-3-4B-IT & 23.50 & 21.79 & 31.48 & 22.31 & 8.04 & 27.53 & 22.04 & 0.97 & 28.80 & 29.59 \\
MiMo-VL-7B-SFT-2508 & 31.18 & 33.01 & 37.17 & 32.98 & 17.59 & 29.14 & 36.89 & 4.91 & 23.20 & 32.82 \\
MiMo-VL-7B-RL-2508 & 34.65 & 36.65 & 42.31 & 34.92 & 18.31 & 36.90 & 37.45 & 6.51 & 30.37 & 32.86 \\
GLM-4.6V$^{\dagger}$ & 42.84 & 50.19 & 47.71 & 37.89 & 42.46 & 51.11 & 36.55 & 8.32 & 34.14 & 39.79 \\
Step3-VL-10B & 32.40 & 42.22 & 40.10 & 31.88 & 17.96 & 24.03 & 17.77 & 5.94 & 27.22 & 27.56 \\
LLaVA-NeXT-72B & 20.40 & 26.54 & 29.42 & 19.93 & 14.18 & 6.34 & 6.19 & 2.99 & 12.75 & 18.63 \\
\midrule
\rowcolor{blue!8}[0pt][0pt]
    \multicolumn{11}{@{}c@{}}{\textbf{Segmentation-capable Models}} \\
    GLaMM-7B  & 28.03 & 37.64 & 27.26 & 18.92 & \cellcolor{green!10}\textbf{84.23} & \underline{37.88} & 0.01 & 1.27 & 3.40 & 17.67 \\

    UFO-8B    & \textbf{42.39} & \textbf{50.29} & \textbf{51.21} & \textbf{34.31} & \underline{78.51}  & \textbf{45.45} & \underline{3.52} & \textbf{8.00} & \textbf{35.97} & \underline{30.21} \\

    LISA++-7B & \underline{37.17} & \underline{37.96} & \underline{50.22} & \underline{28.94} & 76.34 & 23.56 & \textbf{26.51} & \underline{6.70} & \underline{26.73} & \textbf{31.37} \\

\midrule
\rowcolor{blue!8}[0pt][0pt]
\multicolumn{11}{@{}c@{}}{\textbf{Embodied Foundation Models}} \\
RoboBrain2-32B & \textbf{39.32} & \textbf{50.66} & \textbf{49.67} & \textbf{38.83} & 38.85 & 0.61 & 37.63 & 0.08 & \textbf{36.22} & \textbf{39.61} \\
RoboBrain2-7B & 30.09 & 35.93 & 36.01 & 30.52 & \underline{42.92} & 0.00 & 24.31 & 2.29 & \underline{31.58} & 32.95 \\
RoboBrain2.5-8B-NV & 23.81 & 24.81 & 32.88 & 24.97 & 22.57 & 1.66 & 27.98 & 1.13 & 24.98 & 24.03 \\
RoboBrain2.5-8B-MT & 29.83 & 31.20 & 37.41 & 26.76 & 33.52 & \underline{30.89} & 27.84 & 2.65 & 25.45 & 24.34 \\
MiMo-Embodied-7B & \underline{37.16} & 41.22 & \underline{44.17} & \underline{38.74} & 24.67 & \textbf{32.22} & \textbf{40.23} & \textbf{8.71} & 28.23 & 33.88 \\
Pelican1.0-VL-72B & 35.29 & \underline{44.08} & 42.11 & 34.79 & \textbf{44.91} & 0.00 & \underline{38.13} & \underline{7.32} & 22.68 & \underline{37.96} \\
RynnBrain-8B & 22.34 & 28.31 & 33.28 & 24.81 & 7.98 & 5.95 & 15.04 & 0.13 & 13.84 & 11.26 \\
RynnBrain-CoP-8B & 16.83 & 12.48 & 21.78 & 16.08 & 21.51 & 22.51 & 13.58 & 0.28 & 19.62 & 18.09 \\
RynnBrain-Plan-8B & 23.59 & 23.53 & 32.53 & 26.08 & 2.12 & 19.57 & 21.41 & 0.11 & 18.79 & 31.02 \\
RynnBrain-2B & 12.05 & 12.84 & 16.49 & 17.07 & 0.00 & 7.10 & 13.35 & 0.01 & 8.73 & 12.45 \\
VeBrain & 29.27 & 33.63 & 35.98 & 27.69 & 29.56 & 20.04 & 27.58 & 2.85 & 19.25 & 32.30 \\
Cosmos-Reason1-7B & 23.25 & 29.74 & 26.64 & 19.67 & 24.68 & 3.03 & 25.44 & 5.37 & 22.12 & 32.75 \\

\bottomrule

\caption{The full evaluation results for all \textbf{82 VLM variants} on EPIC-Bench.
\colorbox{green!10}{\textbf{Green bold}} indicates the overall best result across all models.
\textbf{Bold} and \underline{underline} indicate the best and second-best results within each model category, respectively. \textbf{$^{\dagger}$} denotes models evaluated with thinking mode enabled. BA, SRA, and ECA denote Basic Attributes, Spatial-Related Attributes, and Embodied Compositional Attributes; GD, FP, and VM denote Ground Detection, Feasible Path, and Visual Matching; AR, CR, and PR denote Affordance Region, Contact Relationship, and Placement Region.}

\label{tab:table-performance-full}
\end{longtable}